%% file: main.tex
\documentclass{article} % For LaTeX2e
\usepackage{iclr2027_conference,times}

\input{math_commands.tex}

\usepackage[hidelinks]{hyperref}
\usepackage{url}
\usepackage{booktabs}
\usepackage{longtable}
\usepackage{graphicx}
\usepackage{amsmath}
\usepackage{placeins}
\usepackage{enumitem}
\makeatletter\newcommand{\captionof}[2]{\def\@captype{#1}\caption{#2}}\makeatother
\usepackage{wrapfig}
\title{Semantic Capability Acquisition and \\ Specialization During Vision-Language \\ Model Fine-Tuning}

\author{Suguru Onda$^1$ \quad Matthew Bailey$^2$ \quad Ryan Farrell$^1$ \\
$^1$Brigham Young University \quad $^2$Carnegie Mellon University}

\iclrfinalcopy
\begin{document}

\maketitle
\lhead{Preprint}

\begin{abstract}
Fine-tuning vision-language models (VLMs) is typically evaluated at a single downstream checkpoint, obscuring whether a semantic capability was never acquired or emerged earlier and later declined during specialization. We ask how semantic capabilities are acquired, when they peak, how well they transfer, and what remains at deployment. We study these dynamics as a \emph{semantic capability trajectory}, tracking identity- and attribute-based capabilities over training.

We formulate a trajectory-based framework that separates capability acquisition, capability-specific optima, and later specialization, and introduce Structured Semantic Routing (SSR) to study how the representation of supervision shapes what is acquired. Across six pretrained backbones spanning DFN, MetaCLIP, and OpenAI CLIP, we show that fine-tuning can acquire semantic capability beyond the pretrained state, including gains observed on held-out evaluations. Unstructured name-and-attribute supervision produces strong name-and-attribute retrieval with comparatively weak name-free attribute-profile retrieval, whereas SSR yields substantially stronger name-free attribute-profile retrieval and is further strengthened by stochastic name-branch dropout. Different capabilities can peak at different stages, so a checkpoint selected by target class-name retrieval need not coincide with a transferable semantic optimum. Continued optimization can therefore preserve strong target class-name retrieval while reducing previously acquired transferable semantic capability. In a representative diagnostic study, this late specialization is consistent with reduced cross-modal semantic accessibility while substantial image-only class structure remains available.
\end{abstract}

\input{sections/01_introduction.tex}
\input{sections/02_related_work.tex}

\input{sections/03_method_and_evaluation.tex}

\input{sections/04_trajectory.tex}
\input{sections/08_discussion.tex}

\subsection*{AI use statement}

In this work, we used generative AI tools for tasks requiring disclosure,
including providing feedback on research methodology and experimental design,
assisting with software and analysis code, assisting in the interpretation and
organization of experimental results, and processing dataset text annotations.
For CUB/NABirds and FineView, source descriptions were collected from public
web resources and generative AI was used to normalize/shorten the descriptions
and extract individual observable visual attributes. For ImageNet, class-level
descriptions and visual attributes were generated with generative AI.

Additionally, we used generative AI tools to summarize and synthesize
author-provided literature, brainstorm and organize the manuscript, suggest
manuscript structure, draft and edit manuscript text for clarity, assist with
scientific figure planning, verify and format bibliographic information, and
propose titles and keywords.

The literature included in the manuscript was curated, reviewed, and verified
by the authors. We reviewed all AI-assisted
work. Quantitative results were obtained from our experimental runs and checked
against the corresponding experiment outputs and provenance records.
AI-processed dataset annotations were reviewed as part of the dataset
construction and experimental pipeline. Literature-based claims and citations
were checked against the source papers, and AI-assisted code and analyses were
reviewed and validated by the authors. We take responsibility for the final
content of this work, including text, claims, dataset annotations, code,
analyses, and artifacts produced with the aid of generative AI.

\subsection*{Reproducibility statement}

Section~\ref{sec:capability_measurement} defines the supervision conditions,
semantic capability probes, checkpoint-selection protocol, and trajectory
quantities used throughout the paper. The supplementary material provides
additional training and evaluation details, exact probe definitions,
checkpoint schedules, per-backbone results, evaluation populations, and
uncertainty analyses where applicable. We report the evaluated checkpoints
and pretrained baselines separately for each evaluation distribution so that
the acquisition, deployment, and post-peak quantities can be reconstructed
from the underlying results.

\subsection*{Acknowledgments}

For the CUB-200-2011 and NABirds species descriptions, we acknowledge the
Cornell Lab of Ornithology's \emph{All About Birds} resource. For the
FineView Butterfly species-description data, we acknowledge Butterflies and
Moths of North America (BAMONA): ``Data were provided by Metalmark Web and
Data, LLC and the many participants who contribute to its Butterflies and
Moths of North America project.''

\bibliography{references}
\bibliographystyle{iclr2027_conference}

\appendix
\input{supplement/supplementary.tex}

\end{document}

%% file: math_commands.tex
\usepackage{amsmath,amsfonts,bm}

\def\eqref#1{equation~\ref{#1}}
\def\1{\bm{1}}

\DeclareMathAlphabet{\mathsfit}{\encodingdefault}{\sfdefault}{m}{sl}
\SetMathAlphabet{\mathsfit}{bold}{\encodingdefault}{\sfdefault}{bx}{n}

%% file: sections/01_introduction.tex
\section{Introduction}
\label{sec:introduction}

Vision--language models (VLMs) are commonly adapted by fine-tuning a pretrained model and evaluating a single resulting checkpoint. This endpoint view cannot distinguish a semantic capability that was never acquired from one that emerged during training and later declined, or remained represented but became less accessible through the cross-modal interface. Prior work has shown that robustness can change substantially over fine-tuning and depends on the adaptation strategy \citep{andreassen2021evolution,wortsman2022wiseft,goyal2023flyp}. We ask a capability-specific question: \emph{what semantic capabilities are acquired during fine-tuning, when do they peak, how well do they transfer, and what remains at deployment?}

This question is especially important when class names provide strong identity anchors. Richer language, fine-grained captions, and structured alignment can improve VLMs \citep{menon2023visualdescription,xie2025fgclip,xie2026fgclip2}, yet strong name-and-attribute retrieval does not necessarily indicate strong use of attributes. WaffleCLIP shows that performance gains from adding descriptive text can persist even when the added text is only weakly semantic \citep{roth2023waffleclip}, while controlled tests that change one attribute or relation at a time show that VLMs may rely less on those details than standard retrieval scores suggest \citep{yuksekgonul2023bags,bader2025sub}. A zero-shot CUB analysis likewise shows that simply adding descriptive text to a class-name prompt can reduce retrieval (Appendix~\ref{app:B_zeroshot}), further motivating the broader question: \emph{which semantic capability is actually acquired?}

We study fine-tuning as a \emph{semantic capability trajectory}. We track four distinct probes: class-name retrieval (P1), name-and-attribute retrieval (P2), name-free attribute-profile retrieval (P3), and Counterfactual Attribute-Profile Preference (CAPP). For each capability, we distinguish the pretrained state, acquisition beyond that state, its capability-specific optimum, and what remains at the checkpoint selected independently by target-domain class-name retrieval. This separates capability acquisition from subsequent specialization and allows different semantic capabilities to reach their optima at different stages of the same fine-tuning run.

To study how supervision affects these trajectories, we introduce \emph{Structured Semantic Routing} (SSR), which represents class identity and descriptive attributes through separate semantic routes during fine-tuning. Across six pretrained backbones spanning DFN~\citep{fang2024dfn}, MetaCLIP~\citep{xu2024metaclip}, and OpenAI CLIP~\citep{radford2021clip}, we find that fine-tuning can acquire semantic capability beyond the pretrained state, including gains observed on held-out evaluations. However, what is acquired depends strongly on how supervision is represented: unstructured name-and-attribute supervision can produce strong name-and-attribute retrieval despite comparatively weak name-free attribute-profile retrieval, whereas SSR produces substantially stronger name-free capability, further strengthened by stochastic name-branch dropout. Different capabilities can peak at different stages, so continued optimization may preserve strong target class-name retrieval while reducing transferable semantic capability acquired earlier.

Finally, we examine what changes between an earlier transferable optimum and a later specialized state. Across a broad trajectory grid, greater peak-to-deployment backbone movement is associated with larger held-out performance decreases in both class-name retrieval (P1) and name-free attribute-profile retrieval (P3), indicating that this association is not specific to the name-free semantic probe. In a representative P3 transition, similarity to both the correct and competing semantic anchors increases, but more strongly for the competitor, reducing the correct-minus-wrong margin while substantial image-only class structure remains accessible. We interpret this representative pattern as reduced cross-modal semantic accessibility.

Our contributions are:
\begin{itemize}[leftmargin=1.35em, labelsep=0.45em]
    \item We formulate VLM fine-tuning as a semantic capability trajectory and measure capability acquisition, capability-specific optima, deployment-time capability, and later specialization.
    \item We introduce Structured Semantic Routing (SSR) and show across six diverse pretrained backbones that supervision content and semantic routing can produce substantially different acquired capability profiles.
    \item We show that transferable semantic capability can peak before a target-selected deployment checkpoint, and we provide representative diagnostic evidence that late degradation can reflect reduced cross-modal semantic accessibility while substantial visual semantic structure remains accessible.
\end{itemize}

%% file: sections/02_related_work.tex
\section{Related Work}
\label{sec:related_work}

\paragraph{Fine-tuning trajectories and preservation.}
Fine-tuning can improve target performance while changing robustness outside the adaptation distribution.
\citet{andreassen2021evolution} explicitly track effective OOD robustness throughout fine-tuning, showing that inherited robustness can be transient; WiSE-FT~\citep{wortsman2022wiseft} recovers robustness by interpolating pretrained and fine-tuned weights, FLYP~\citep{goyal2023flyp} changes downstream behavior by preserving the contrastive pretraining objective, and LiT~\citep{zhai2022lit} learns cross-modal alignment while one modality remains fixed.
These works establish that optimization stage and adaptation strategy matter.
Our focus is capability-specific: we ask whether a semantic capability is inherited or newly acquired, when it peaks, and whether it remains transferable at deployment.

\paragraph{Descriptive supervision and semantic diagnostics.}
Classification-by-description methods enrich class names with visual descriptors~\citep{menon2023visualdescription}, while WaffleCLIP shows that such prompt gains need not isolate useful descriptive semantics~\citep{roth2023waffleclip}.
Name-free descriptive recognition has also been studied directly~\citep{baron2024realclassification}, including with large VLMs~\citep{atabuzzaman2025zeroshotfinegrained}; FG-CLIP and FG-CLIP~2 add long captions, regional alignment, hard negatives, and stronger fine-grained textual discrimination during training~\citep{xie2025fgclip,xie2026fgclip2}.
Other diagnostics expose failures hidden by aggregate accuracy:
ARO/NegCLIP and VL-CheckList test compositional or attribute sensitivity~\citep{yuksekgonul2023bags,zhao2022vlchecklist}, FG-OVD uses attribute-modified alternatives~\citep{bianchi2024fgovd}, and SUB shows that concept predictions can follow class-correlated expectations rather than modified visual evidence~\citep{bader2025sub}.
Our novelty is not name removal or counterfactual testing itself, but using distinct probes longitudinally to measure acquisition, peak timing, and retention.

\paragraph{Cross-modal accessibility and grounding.}
Poor multimodal output need not imply absent visual information.
Hidden in Plain Sight finds information readable from visual features that the full VLM fails to exploit~\citep{fu2025hidden}, while VLMs Need Words documents strong reliance on nameable semantic anchors~\citep{shahgir2026vlmsneedwords}.
Modality-gap work further cautions that global image/text geometry is not itself evidence of semantic loss~\citep{liang2022mindgap}.
Conversely, RegionCLIP and GLIP learn explicit region--text or phrase--region correspondence~\citep{zhong2022regionclip,li2022glip}, a stronger spatial claim than our probes support.
We therefore use \emph{cross-modal semantic accessibility} narrowly: P3 and CAPP measure semantic discrimination in the shared embedding space, not spatial grounding or localization.

%% file: sections/03_method_and_evaluation.tex
\section{Structured Semantic Fine-Tuning and Capability Measurement}
\label{sec:capability_measurement}

We evaluate each fine-tuning run at multiple checkpoints, separating the supervision intervention from the capability probe: the same training condition can score very differently depending on whether evaluation retains a class-name anchor, removes it, or tests a controlled semantic counterfactual.

\begin{figure}[t]
    \centering
    \includegraphics[width=\linewidth]{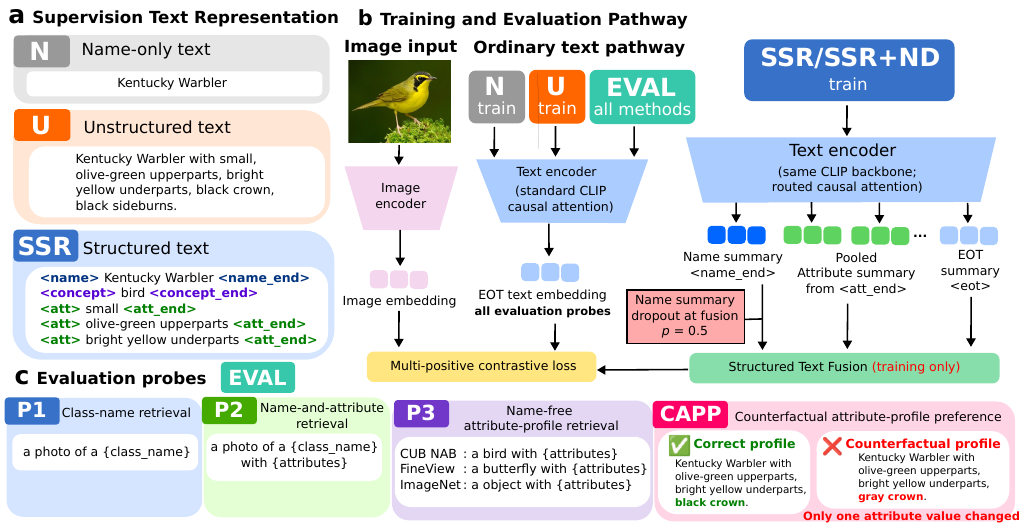}
    \vspace{-2pt}
    \caption{\textbf{Supervision representation, training pathway, and capability probes.} \textbf{(a)} N uses name-only supervision, U encodes the class name and attributes in one ordinary text stream, and SSR/SSR+ND use structured NAME and ATTRIBUTE segments, with a shared CONCEPT segment for CUB/NAB. \textbf{(b)} N/U training and all evaluations use the standard CLIP causal text pathway; SSR/SSR+ND use routed causal attention and a training-only fusion of NAME, pooled ATTRIBUTE, and EOT summaries, with NAME-summary dropout for SSR+ND. \textbf{(c)} Exact P1--P3 prompts and the CAPP counterfactual comparison. Photograph: CUB-200-2011.}
    \label{fig:method}
    \vspace{-2pt}
\end{figure}

\paragraph{Supervision interventions.}
We compare four conditions (Figure~\ref{fig:method}). \textbf{Name-only (N)} uses class-name-centered contrastive supervision. \textbf{Unstructured (U)} encodes the class name and attributes together in one ordinary text stream. \textbf{Structured Semantic Routing (SSR; Appendix~\ref{app:ssr_architecture})} uses the same pretrained VLM backbone and descriptive supervision, fine-tuned in every condition, but separates NAME and ATTRIBUTE segments with routed causal attention during training. ATTRIBUTE routes have no direct NAME access; for CUB/NAB they may read a shared coarse CONCEPT such as ``bird,'' which is context only and is not fused (ImageNet uses no CONCEPT segment). A training-only Structured Text Fusion combines the NAME summary, a pooled ATTRIBUTE summary, and the EOT summary. \textbf{SSR+ND} is identical to SSR except that the NAME-summary contribution is dropped with probability $p{=}0.5$ at training-time fusion, with the remaining fusion weights renormalized.

All methods are evaluated identically. P1, P2, P3, and CAPP use the standard CLIP causal text encoder and the ordinary EOT representation; structured tokens, routed attention, attribute pooling, and Structured Text Fusion are used only during SSR/SSR+ND training. Canonical SSR routing masks are shown in Figure~\ref{fig:S1_routing}; the FULL/PARTIAL/NO name-access masks used in the ablation are shown in Figure~\ref{fig:S2_nameaccess}.

\paragraph{Capability probes.}
\textbf{P1} (class-name retrieval) uses \texttt{a photo of a \{class\_name\}} (Recall@1) and measures class-identity accessibility.
\textbf{P2} (name-and-attribute retrieval) uses \texttt{a photo of a \{class\_name\} with \{attributes\}}, encoded through the ordinary EOT pathway (not the fusion output); because both identity and descriptive information are present, P2 is not attribute-isolating.
\textbf{P3} (name-free attribute-profile retrieval) removes the class-name anchor and measures class-level attribute accessibility through the ordinary EOT pathway. The exact templates are \texttt{a bird with \{attributes\}} (CUB/NAB), \texttt{a butterfly with \{attributes\}} (FineView), and the frozen canonical ImageNet template \texttt{a object with \{attributes\}}, used exactly as evaluated; P3 does not establish localization, spatial grounding, or part binding.
\textbf{CAPP} (Counterfactual Attribute-Profile Preference) contrasts the correct full attribute profile with a same-image counterfactual differing in one valid semantic slot/value. Pairwise CAPP uses one alternative; CAPP-Multi ranks the correct value against all valid alternatives for the same semantic slot, with accuracy averaged equally across class--slot pairs. It measures controlled attribute-value discrimination and is complementary to, not interchangeable with, P3.
Exact prompts and the CAPP construction are shown in Figure~\ref{fig:method}c.

\paragraph{Acquisition, deployment, and post-peak change.}
For evaluation distribution $d$, PT is the pretrained checkpoint evaluated on $d$; $P3_{\mathrm{peak}}(d)$ is the trajectory maximum and $P3_{\mathrm{deploy}}(d)$ its value at the P1-selected deployment checkpoint:
\begin{equation}
\begin{aligned}
\mathrm{Gain}_{\mathrm{peak}}
    &= P3_{\mathrm{peak}}-P3_{\mathrm{PT}}, &
\mathrm{Gain}_{\mathrm{deploy}}
    &= P3_{\mathrm{deploy}}-P3_{\mathrm{PT}},\\
\mathrm{Forget}_{\mathrm{postpeak}}
    &= P3_{\mathrm{peak}}-P3_{\mathrm{deploy}} .
\end{aligned}
\end{equation}
The deployment checkpoint is the predefined evaluated checkpoint with maximum \emph{target-domain} P1 for that run; it is never selected using P3, CAPP, or held-out performance, so a semantic optimum may fall before, at, or after the deployment checkpoint.
We call post-peak decline a reduction of \emph{acquired} capability only when the trajectory first rose above its pretrained baseline.

\paragraph{Backbones, datasets, and text sources.}
We evaluate six pretrained dual encoders: DFN5B ViT-H/14-378 and DFN ViT-L/14~\citep{fang2024dfn}, MetaCLIP ViT-H/14 and ViT-L/14~\citep{xu2024metaclip}, and OpenAI CLIP ViT-L/14 and ViT-B/32~\citep{radford2021clip}, on ImageNet~\citep{deng2009imagenet}, CUB-200-2011~\citep{wah2011cub}, NABirds~\citep{vanhorn2015nabirds}, FineView Butterfly~\citep{onda2025fineview}, and the ImageNet-V2/-R/-Sketch shifts~\citep{recht2019imagenetv2,hendrycks2021imagenetr,wang2019imagenetsketch}.
CUB/NABirds/FineView descriptions come from public species resources, LLM-processed to normalize length and extract attributes; ImageNet descriptions and attributes are LLM-generated.
For reciprocal bird transfer, CUB-trained models are evaluated on the dataset-class-disjoint NAB-348 subset, and NAB-trained models on the dataset-class-disjoint CUB-59 subset; results on the corresponding full populations are reported separately in the supplement.
Full protocol details are in Appendix~\ref{app:shared_protocol}.

%% file: sections/04_trajectory.tex
\section{Semantic Capability Trajectories}
\label{sec:trajectory}

Figure~\ref{fig:acquisition}a illustrates the trajectory view in a representative CUB-trained model. P1 and P3 do not evolve together: held-out P3 can rise above the pretrained state, peak early, and then decline while target-domain P1 continues toward its own maximum. Thus, the checkpoint selected by class-name retrieval can occur after the best transferable semantic checkpoint. This is clearest for SSR+ND on held-out NAB-348, where held-out P3 peaks around epoch 20 while the P1-selected deployment checkpoint is epoch 60; other methods and domains can peak before, at, or after their deployment checkpoint, so the general point is that capability-specific optima need not coincide with the P1-selected checkpoint. Target and held-out semantic capabilities can also peak at different stages.

\subsection{Capability Acquisition}
\label{sec:acquisition}

Can fine-tuning acquire semantic capability beyond the pretrained state? We define acquisition as $\mathrm{Gain}_{\mathrm{peak}} = P3_{\mathrm{peak}} - P3_{\mathrm{PT}}$ (\S\ref{sec:capability_measurement}), a retrospective measure of whether a capability ever emerged, not a deployment rule.

\begin{figure}[t]
    \centering
    \includegraphics[width=\linewidth]{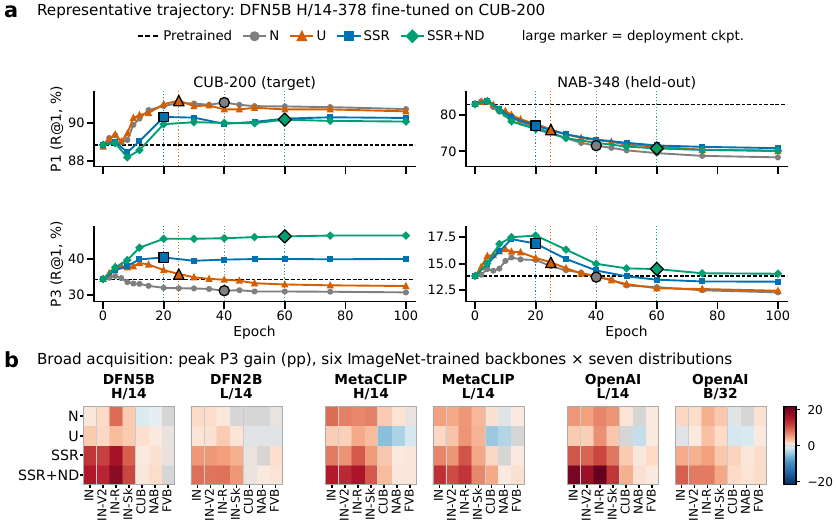}
    \vspace{-4pt}
    \caption{\textbf{Capability acquisition and trajectory structure.} \textbf{(a)} DFN5B H/14-378 CUB$\to$NAB trajectories; large markers = P1-selected deployment checkpoints. \textbf{(b)} Peak P3 gain over pretrained, six ImageNet-trained backbones $\times$ seven evaluation distributions.}
    \label{fig:acquisition}
    \vspace{-2pt}
\end{figure}

Figure~\ref{fig:acquisition}b shows that fine-tuning can acquire P3 beyond the pretrained state across six heterogeneous ImageNet-trained backbones, which we treat as distinct models rather than stochastic replicates, and seven evaluation distributions (ImageNet val, three same-ontology ImageNet shifts, and three cross-ontology fine-grained datasets). The largest gains occur on the target and same-ontology ImageNet evaluations, where SSR and especially SSR+ND generally produce much larger gains than N or U; U's mean gain is much smaller and positive in most, but not all, backbones. Cross-ontology gains are substantially smaller and more variable, although structured supervision acquires positive P3 more consistently than N/U overall; the ordering of SSR and SSR+ND is not uniform on every held-out dataset, and no condition is superior in every cell (Table~\ref{tab:S4_summary}).

Held-out acquisition is not restricted to structured supervision: in Figure~\ref{fig:acquisition}a, even name-only fine-tuning transiently raises held-out P3 above the pretrained baseline (peaks of roughly 15.6--17.7 versus about 13.9), without implying direct attribute supervision. Timing mismatches are analyzed systematically in \S\ref{sec:specialization}.

A weak deployed score therefore does not establish that a capability was never learned; the next question is what determines which capability is acquired.

\subsection{Supervision Shapes the Acquired Capability Profile}
\label{sec:capability_differences}

\begin{figure}[t]
\centering
\begin{minipage}[t]{0.475\linewidth}
\centering\footnotesize\setlength{\tabcolsep}{3pt}
\begin{tabular}{lccccc}
\toprule
& PT & N & U & SSR & SSR+ND \\
\midrule
P1 & 79.7 & \textbf{85.8} & \textbf{85.8} & 84.9 & 84.7 \\
P2 & 74.1 & 81.1 & \textbf{86.0} & 82.0 & 82.4 \\
P3 & 29.9 & 32.2 & 28.1 & 35.6 & \textbf{44.6} \\
\bottomrule
\end{tabular}
\par\vspace{2pt}
\captionof{table}{\textbf{ImageNet deployment profile.} Six-backbone median R@1 (\%) on ImageNet val at each run's P1-selected checkpoint. Per-backbone values: Table~\ref{tab:S8_deployment}.}\label{tab:deploy_profile}
\end{minipage}\hfill
\begin{minipage}[t]{0.505\linewidth}
\centering\footnotesize\setlength{\tabcolsep}{2.6pt}
\begin{tabular}{lccccc}
\toprule
& PT & N & U & SSR & SSR+ND \\
\midrule
CUB Pairwise & 70.8 & 70.9 & 69.9 & 72.3 & \textbf{74.7} \\
CUB CAPP-Multi & 45.0 & 44.6 & 42.4 & 46.8 & \textbf{49.9} \\
NAB Pairwise & 63.4 & 60.6 & 63.3 & 64.7 & \textbf{65.8} \\
NAB CAPP-Multi & 31.6 & 28.4 & 30.2 & 32.5 & \textbf{33.9} \\
\bottomrule
\end{tabular}
\par\vspace{2pt}
\captionof{table}{\textbf{Target CAPP summary.} Six-backbone median (\%) at each run's P1-selected deployment checkpoint. Per-backbone values: Tables~\ref{tab:S11_target_pairwise}--\ref{tab:S11_multi}.}\label{tab:capp_target_summary}
\end{minipage}
\end{figure}

We next examine whether adding descriptive information is sufficient, or whether its representation and routing determine what is learned. Table~\ref{tab:deploy_profile} shows that the four interventions reach different deployed capability profiles: P1 remains broadly strong across conditions, U reaches the highest P2, and P3 diverges sharply, with SSR and especially SSR+ND reaching much higher values than PT/N/U.

\begin{figure}[t]
    \centering
    \includegraphics[width=\linewidth]{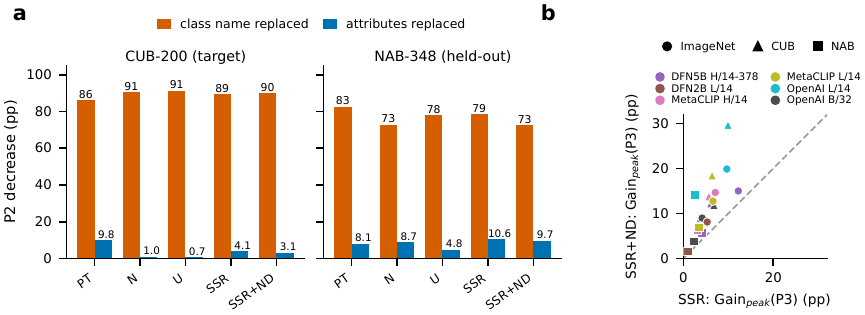}
    \vspace{-2pt}
    \includegraphics[width=\linewidth]{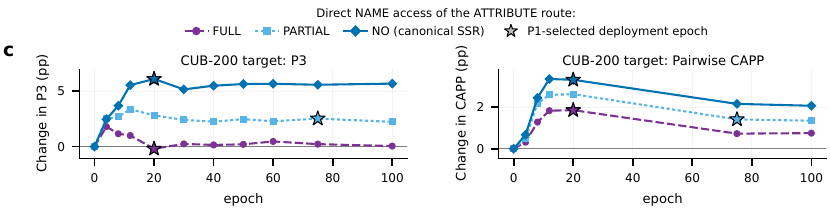}
    \vspace{-4pt}
    \caption{\textbf{Why supervision changes what is acquired.} \textbf{(a)} P2 decrease when only the class name or only the attributes are replaced; DFN5B H/14-378, CUB-trained (CIs: Table~\ref{tab:S9_name_dependency}). \textbf{(b)} SSR+ND vs.\ SSR peak target P3 gain, 18 backbone$\times$domain pairs (Table~\ref{tab:S10_s_vs_snd_full}). \textbf{(c)} Direct name-access ablation on the CUB target (DFN5B/CUB; one run per variant). Restricting ATTRIBUTE-route NAME access increases target P3 and Pairwise CAPP; NO is canonical SSR. Stars mark P1-selected deployment; PARTIAL deploys later, so this is a trajectory rather than matched-deployment comparison. Full trajectories: Figure~\ref{fig:S_trackA_traj}.}
    \label{fig:supervision}
    \vspace{-2pt}
\end{figure}

High P2 does not imply strong attribute use. Changing only the class name reduces P2 by 73--91 percentage points across all five conditions on CUB-200 and strict held-out NAB-348, whereas changing only the attributes reduces it by roughly 1--11 points (Figure~\ref{fig:supervision}a; DFN5B H/14-378 CUB-trained run; 95\% bootstrap CIs over classes, 1,000 resamples, in Table~\ref{tab:S9_name_dependency}). Thus U's high P2 largely reflects class-name reliance, consistent with its comparatively weak deployed P3. SSR and SSR+ND are more attribute-sensitive than U but remain strongly name-dependent, showing increased attribute use alongside, rather than instead of, identity. This is a single-run result, not a six-backbone claim.

How the identity contribution is handled during training -- not just the text content -- determines how much capability is acquired. SSR and SSR+ND receive identical structured text and use identical routing; SSR+ND additionally drops the NAME channel at fusion with probability 0.5 during training (\S\ref{sec:capability_measurement}). Across six backbones and three training domains, SSR+ND reaches a larger peak target P3 gain than SSR in all 18 tested backbone$\times$domain pairs (Figure~\ref{fig:supervision}b; full values in Table~\ref{tab:S10_s_vs_snd_full}). Making the identity contribution less consistently available at training-time fusion changes how much name-free capability is acquired, not only how it is later retained.

CAPP provides complementary evidence by testing controlled attribute discrimination rather than profile retrieval. At each run's own P1-selected deployment checkpoint, SSR+ND has the highest six-backbone median among the fine-tuned conditions in all four target-CAPP rows (Table~\ref{tab:capp_target_summary}). On the CUB target ontology, SSR improves Pairwise CAPP over U in all six backbones and SSR+ND improves it further in all six, and CAPP-Multi gives the same broad conclusion (Tables~\ref{tab:S11_target_pairwise}--\ref{tab:S11_multi}); the NAB target is less uniform across backbones, and cross-ontology CAPP is more variable still (Table~\ref{tab:S12_heldout_capp}). Thus P3 and CAPP support the same broad conclusion while measuring different capabilities; they are not interchangeable.

In a single DFN5B/CUB run, isolating the ATTRIBUTE route from direct class-name access (Figure~\ref{fig:S2_nameaccess}) improves target P3 and target CAPP (Figure~\ref{fig:supervision}c); full target/held-out results and additional diagnostics are reported in Tables~\ref{tab:E_trackA_p3}--\ref{tab:E_trackA_rsa} and Figure~\ref{fig:S_trackA_traj}. However, SSR does not eliminate identity reliance: P2 remains strongly dependent on the class name, as shown by the class-name replacement test.

Together, these results show that adding attribute text and changing how that information is routed are different interventions. U sees descriptive attributes but remains strongly name-dominated in P2. SSR produces stronger name-free P3, while target CAPP improves most consistently on CUB; making the NAME contribution unreliable during training with SSR+ND further increases target P3 acquisition across all 18 backbone$\times$domain comparisons. Which capability is acquired is only part of the trajectory; its own optimum need not coincide with the checkpoint selected by target P1.

\subsection{Continued Optimization Can Produce Semantic Specialization}
\label{sec:specialization}

After a semantic capability is acquired, we distinguish acquisition ($\mathrm{Gain}_{\mathrm{peak}}$), what remains at deployment ($\mathrm{Gain}_{\mathrm{deploy}}$), and their difference ($\mathrm{Forget}_{\mathrm{postpeak}} = P3_{\mathrm{peak}} - P3_{\mathrm{deploy}}$, \S\ref{sec:capability_measurement}).

\begin{figure}[t]
    \centering
    \includegraphics[width=\linewidth]{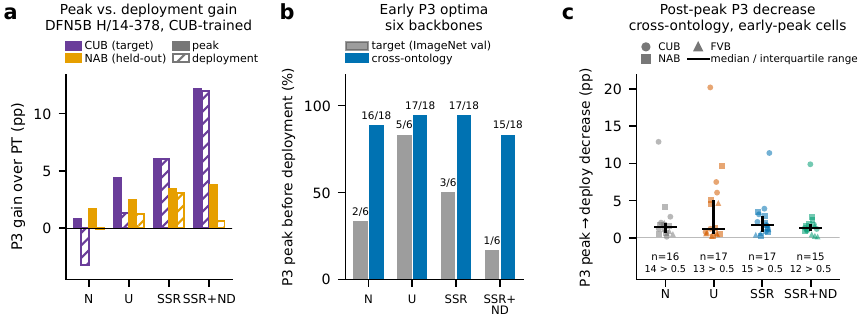}
    \vspace{-4pt}
    \caption{\textbf{Capability-specific timing and specialization.} \textbf{(a)} Representative peak versus deployment P3 gain for DFN5B H/14-378 fine-tuned on CUB-200. \textbf{(b)} Fraction of target (ImageNet val) and cross-ontology evaluations whose P3 optimum precedes deployment, by method across six backbones. \textbf{(c)} Peak-to-deployment P3 decrease for cross-ontology trajectories whose P3 optimum precedes deployment; $n$ varies because cells peaking at deployment have no post-peak interval.}
    \label{fig:specialization}
    \vspace{-2pt}
\end{figure}

Figure~\ref{fig:specialization}a illustrates why peak and deployment must be separated: a capability can be acquired strongly and still be reduced before the P1-selected checkpoint. The panel shows the representative DFN5B H/14-378 CUB-trained run. N acquires target and held-out P3 above pretrained at its peak, but by deployment its held-out gain has essentially disappeared and target P3 is below the pretrained level. SSR retains target P3 almost completely, and SSR+ND acquires the largest target P3 gain of any condition; its held-out NAB P3, however, peaks around epoch 20, before the epoch-60 deployment checkpoint, while its target CUB P3 peaks later, around epoch 75. Target and held-out capabilities can therefore specialize differently. In this run, the P3 and Pairwise-CAPP peaks also need not coincide with each other or with deployment (Appendix Figure~\ref{fig:S_capp_timing}).

Across the six backbones and four methods, cross-ontology P3 peaks before deployment in 65 of 72 cells, compared with 40 of 72 same-ontology shifts and 11 of 24 target-domain cells (Figure~\ref{fig:specialization}b plots the target and cross-ontology fractions; same-ontology counts are in Appendix~\ref{app:timing}). This pattern is consistent across methods, with 15--17 of 18 cross-ontology cells peaking early for each method. Among these 65 early-peaking trajectories, 54 lose more than 0.5 percentage points of P3 by deployment; the 0.5-point cutoff is a descriptive robustness threshold, not a significance test. The median decline is similar across methods, ranging from 1.21 to 1.71 points (Figure~\ref{fig:specialization}c). Thus, transferable P3 often reaches its best value before the target-P1-selected deployment checkpoint and then declines with continued optimization, while target-domain P3 more often peaks at the deployment checkpoint. Full timing results are provided in Appendix~\ref{app:timing}.

\subsection{Understanding Late Specialization}
\label{sec:mechanism}

To characterize the transition from a capability's peak to a later specialized state, we first test whether parameter movement tracks acquisition and the post-peak decrease across the broader trajectory grid, then examine the geometry of the held-out decrease in a representative transition.

\begin{figure}[t]
    \centering
    \includegraphics[width=\linewidth]{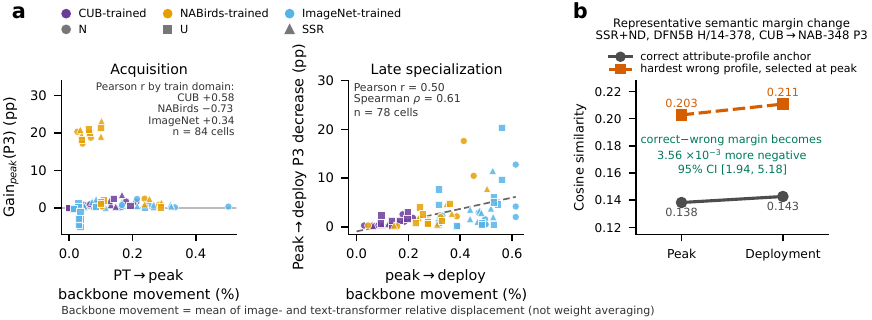}
    \vspace{-4pt}
    \caption{\textbf{Late semantic specialization.} \textbf{(a)} Joint backbone movement before the P3 peak versus acquisition, and from peak to deployment versus the post-peak P3 decrease, across held-out N/U/SSR trajectories; $r$ = Pearson, $\rho$ = Spearman, $n$ = trajectory cells. \textbf{(b)} Representative fixed-competitor analysis for SSR+ND, DFN5B H/14-378, CUB$\to$NAB-348. The peak-selected hardest wrong class is held fixed at deployment; the correct-minus-wrong margin decreases by 0.00356 (95\% class-bootstrap CI $[0.00194,0.00518]$). Because the competitor is the hardest wrong profile selected at the peak, its mean similarity can already exceed that of the correct profile; the analysis tests the subsequent change in their fixed-competitor margin.}
    \label{fig:mechanism}
    \vspace{-2pt}
\end{figure}

\textbf{Broad association.} We next ask whether parameter movement accompanies acquisition or later decline. In a separate movement-analysis grid covering four backbones and N/U/SSR (Appendix~\ref{app:G_move}; SSR+ND is not included), we measure movement as the arithmetic mean of matched relative displacement in the pretrained image and text transformers. Pretrained-to-peak movement has no consistent relationship with P3 acquisition: the association changes sign across training domains (Figure~\ref{fig:mechanism}a, left). In contrast, among held-out trajectories that peak before deployment, greater peak-to-deployment movement is associated with a larger P3 decrease ($r=0.50$, $\rho=0.61$, $n=78$; Figure~\ref{fig:mechanism}a, right). Image- and text-encoder movement are strongly correlated ($r=0.85$), and which encoder appears more strongly associated with the post-peak decrease depends on how movement is scaled across backbones; we therefore do not attribute the effect to either encoder individually. These are descriptive associations across heterogeneous backbones, not causal estimates or stochastic replicates. Accordingly, this association should not be interpreted as evidence that the same movement relationship holds for SSR+ND. The relationship is not specific to P3: over the same intervals, movement is even more strongly associated with held-out P1 decline ($r=0.73$, $\rho=0.79$; Appendix~\ref{app:G_p1move}).

\textbf{Representative geometry.} Figure~\ref{fig:mechanism}b examines the representative SSR+ND CUB$\to$NAB-348 transition. From the P3 peak to deployment, the peak-selected wrong attribute profile moves closer to the image more than the correct profile does, reducing the correct-versus-wrong margin on the held-out data (Appendix~\ref{app:G_enc}). Supplementary analyses support and bound this interpretation. The effect extends across the wrong-class similarity spectrum rather than being specific to the hardest competitor (Appendix~\ref{app:G_hardness}), but the exact signature is not universal: an independent text-frozen U control does not exhibit the same resolved geometric signature (Appendices~\ref{app:G_enc}--\ref{app:G_tf}). Image-only class structure remains substantially preserved across pretrained, peak, and deployment states (Appendix~\ref{app:G_loo}), and coarse geometry hypotheses do not fully explain the effect (Appendix~\ref{app:G_geom}). Together, these results suggest that the representative late-stage decline reflects reduced cross-modal semantic accessibility rather than broad erasure of visual class information, while also showing that late-stage semantic degradation does not always arise through the same embedding-geometry change.

%% file: sections/08_discussion.tex
\section{Discussion and Limitations}
\label{sec:discussion}

Viewing fine-tuning as a capability trajectory rather than a single pretrain-versus-deployment comparison changes what a weak deployed score means: the capability may never have been acquired, or it may have emerged earlier and later become less accessible. Our results show both cases. Unstructured descriptive supervision can acquire little transferable name-free capability, whereas structured supervision can acquire substantial capability whose optimum precedes deployment.

Two additional controls help bound the late-specialization interpretation (Appendix~\ref{app:controls}). In the representative DFN5B/CUB setting, WiSE interpolation partially recovers held-out P3 for N/U while preserving strong target P1/P3, showing that the decline need not be irreversible. We do not apply this interpolation to SSR/SSR+ND because their structured components lack direct pretrained counterparts. Separately, a target-side hard-negative objective improves target P3 without improving held-out P3 at matched later checkpoints, showing that stronger in-domain semantic discrimination does not necessarily translate into stronger transferable capability.

\paragraph{Limitations.} The \S\ref{sec:mechanism} geometric diagnostics are representative single-run analyses, and the broader movement analysis is descriptive rather than causal; neither supports encoder-specific attribution. Main trajectory cells are also single runs, so variation across the six backbones reflects model heterogeneity rather than training stochasticity. Our comparisons are controlled supervision/routing interventions rather than a benchmark against task-specific fine-grained methods. Deployment is selected by target P1, and P2/P3/CAPP have the scope limitations defined in \S\ref{sec:capability_measurement}. Dataset-text sources differ across domains, so residual source, template, or generation effects cannot be excluded. Additional limitations are detailed in Appendix~\ref{app:limitations}.

\section{Conclusion}
\label{sec:conclusion}

We studied VLM fine-tuning as a semantic capability trajectory rather than a single pretrain-to-deployment endpoint. Across six pretrained backbones, fine-tuning can acquire capabilities beyond the pretrained state, and Structured Semantic Routing, particularly with NAME-summary dropout, strengthens name-free attribute-profile capability in most tested settings, showing that how descriptive supervision is represented and routed changes what is acquired. Different capabilities reach their optima at different stages: transferable name-free attribute capability often peaks before the checkpoint selected by target class-name retrieval. In a representative diagnostic study, the subsequent decline was consistent with reduced cross-modal semantic accessibility while substantial image-only class structure remained. Tracking capability trajectories therefore distinguishes failure to acquire from post-acquisition decline.

%% file: supplement/supplementary.tex
% Supplementary tables/figures use an "S" prefix and their own counters, independent of the
% main-paper numbering.
\setcounter{table}{0}
\renewcommand{\thetable}{S\arabic{table}}
\setcounter{figure}{0}
\renewcommand{\thefigure}{S\arabic{figure}}
% hyperref anchor names must stay unique across main-paper and supplement counters.
\renewcommand{\theHtable}{S.\arabic{table}}
\renewcommand{\theHfigure}{S.\arabic{figure}}

\section{SSR Architecture and Objective}
\label{app:ssr_architecture}

This appendix provides detailed method, protocol, and numerical documentation for the paper. Sections~\ref{app:ssr_architecture}--\ref{app:training} describe the method, data, probes, and training recipes; Sections~\ref{app:acquisition}--\ref{app:controls} give the full numerical support for every main-paper figure and claim; Section~\ref{app:provenance} maps each main-paper item to its supplement table, population, checkpoint rule, and source artifact. Six-backbone results are descriptive across heterogeneous pretrained models and are never treated as stochastic replicates. Sections marked ``single run'' report one training run. The formulation below is consistent with the current implementation and the archived parameter records for the reported runs.

\subsection{Structured semantic representation and routed attention}
\label{app:ssr_arch_overview}

Let $\mathcal{D}=\{(I_i,y_i)\}$ be a training set. Each class $c$ has a class name $n_c$ and attributes $\mathcal{A}_c=\{a_{c,1},\dots,a_{c,K_c}\}$. For the CUB/NAB canonical SSR, with shared coarse concept $q_c$ (e.g., ``bird''), the structured sequence is
\[
\begin{aligned}
x_c=\big[\,&\texttt{<sot>},\ \texttt{<name>}\,n_c\,\texttt{<name\_end>},\ \texttt{<concept>}\,q_c\,\texttt{<concept\_end>},\\
&\{\texttt{<att>}\,a_{c,k}\,\texttt{<att\_end>}\}_{k=1}^{K_c},\ \texttt{<eot>}\,\big],
\end{aligned}
\]
for example \texttt{<name> Kentucky Warbler <name\_end> <concept> bird <concept\_end> <att> small <att\_end> <att> olive-green upperparts <att\_end> $\ldots$}. The concept is a coarse semantic context shared across the fine-grained classes of the dataset; it is not the class identity. For ImageNet canonical SSR there is no concept segment:
\[
x_c=[\texttt{<sot>},\ \texttt{<name>}\,n_c\,\texttt{<name\_end>},\ \{\texttt{<att>}\,a_{c,k}\,\texttt{<att\_end>}\}_{k=1}^{K_c},\ \texttt{<eot>}].
\]
ImageNet training uses no concept token. The word ``object'' occurs only in the P3 \emph{evaluation} template (``a object with \{attributes\}'') and never as a training-time concept.

\paragraph{Routed attention.} Every position attends to \texttt{<sot>} and itself, and attention inside a segment is causal. Let $r^{\mathrm{name}}$, $r^{\mathrm{att}}_k$, and $r^{\mathrm{eot}}$ denote the positions of \texttt{<name\_end>}, the $k$-th \texttt{<att\_end>}, and \texttt{<eot>}.
\begin{itemize}
\item \textbf{CUB/NAB (shared-concept routing).} The NAME route reads only the NAME segment. CONCEPT tokens read only the CONCEPT segment. The $k$-th ATTRIBUTE route reads its own segment \emph{and the shared CONCEPT content}, but not the NAME segment and not other attribute segments. \texttt{<eot>} reads only the summary endpoints $\{r^{\mathrm{name}}\}\cup\{r^{\mathrm{att}}_k\}_k$.
\item \textbf{ImageNet (no-concept routing).} As above without a CONCEPT segment: the $k$-th ATTRIBUTE route reads only its own segment, there is no NAME-to-ATTRIBUTE access, and \texttt{<eot>} reads only $\{r^{\mathrm{name}}\}\cup\{r^{\mathrm{att}}_k\}_k$.
\end{itemize}
\texttt{<eot>} does not attend to every raw token and does not read the concept boundary token. The exact masks are given in \S\ref{app:ssr_masks}.

\paragraph{Relation to earlier SSR routing variants.} Earlier SSR routing variants allowed each attribute route to access the class-name segment directly and used no concept segment. In the current canonical CUB/NAB formulation, we instead provide the attribute routes with a shared coarse concept (e.g., ``bird'') while removing their direct access to the fine-grained class name. This provides coarse semantic context without giving the attribute route direct access to the fine-grained class identity; it does not imply complete disentanglement, and class-name information still reaches the fused representation through the NAME summary (and ordinary text-encoder P2 remains strongly name-dominated; \S\ref{app:supervision}). ImageNet uses the corresponding no-concept routing. Some historical runs (\S\ref{app:multiseed}) used the earlier routing and are never pooled with the canonical one.

\subsection{Attribute pooling, structured fusion, and name-branch dropout}
\label{app:ssr_fusion}

Passing $x_c$ through the routed text encoder gives final-layer hidden states $H_c$, from which three kinds of features are read at the boundary tokens:
\[
h^{\mathrm{name}}_c=H_c[\texttt{<name\_end>}],\qquad h^{\mathrm{att}}_{c,k}=H_c[\texttt{<att\_end>}_k],\qquad h^{\mathrm{eot}}_c=H_c[\texttt{<eot>}].
\]
No concept feature is extracted. The concept affects the attribute features only through the routed attention.

\paragraph{Attribute Importance Pooling (AIP).} The ATT\_END features are pooled into one attribute summary with a learned importance score (GELU $\sigma$; padded attributes masked before the softmax):
\[
\beta_{c,k}=\frac{\exp\!\big(w_2^\top\sigma(W_1h^{\mathrm{att}}_{c,k})\big)}{\sum_{j=1}^{K_c}\exp\!\big(w_2^\top\sigma(W_1h^{\mathrm{att}}_{c,j})\big)},\qquad h^{\mathrm{attr}}_c=\sum_{k=1}^{K_c}\beta_{c,k}h^{\mathrm{att}}_{c,k}.
\]
AIP is applied to the final-layer hidden states before the text projection.

\paragraph{StructuredTextFusion (training only).} The three channels (NAME, pooled ATTRIBUTE, EOT) are passed through the text projection $P$ and normalized, $\tilde h^{\mathrm{ch}}_c=\mathrm{norm}(P\,h^{\mathrm{ch}}_c)$ for $\mathrm{ch}\in\{\mathrm{eot},\mathrm{name},\mathrm{attr}\}$. A gate over the concatenated channels predicts fusion weights, and the structured training text embedding is the normalized weighted sum:
\[
\alpha_c=\mathrm{softmax}\Big(W_2\,\sigma\big(W_1[\tilde h^{\mathrm{eot}}_c;\tilde h^{\mathrm{name}}_c;\tilde h^{\mathrm{attr}}_c]\big)\Big)\in\mathbb{R}^3,\qquad
t_c=\mathrm{norm}\Big(\alpha^{\mathrm{eot}}_c\tilde h^{\mathrm{eot}}_c+\alpha^{\mathrm{name}}_c\tilde h^{\mathrm{name}}_c+\alpha^{\mathrm{attr}}_c\tilde h^{\mathrm{attr}}_c\Big).
\]
The fusion inputs are only the NAME, ATTRIBUTE, and EOT summaries; the concept is \emph{not} a fusion input. The learned gate is a two-layer network ($3D\to D\to 3$).

\paragraph{SSR+ND.} SSR+ND uses the same structured sequence, routed encoder, and routing mask as SSR. During training, for each sample independently with probability $p=0.5$, the NAME channel's gate logit is set to $-\infty$ before the softmax, so $\alpha^{\mathrm{name}}_c=0$ and the EOT and ATTRIBUTE weights are renormalized by the softmax. The NAME tokens remain in the encoded sequence, the routing mask is unchanged, and the NAME feature still enters the gate network's input; only its fusion weight is removed. Dropout is applied in training only.

\paragraph{Inference and evaluation.} At inference and for every capability probe (P1, P2, P3, CAPP), plain prompts are encoded by the ordinary causal text encoder and the feature at the EOT position is projected (the ``ordinary EOT text embedding''); no structured tokens, routing mask, AIP, or fusion are used. The structured fusion is a training-time construct only, and the earlier formulation's fused-representation inference and attribute-patch reranking are not part of the evaluation reported here. N and U use the ordinary text pathway in both training and evaluation.

\subsection{Multi-positive CLIP-style contrastive objective}
\label{app:ssr_loss}

For a training batch of image embeddings $Z\in\mathbb{R}^{B\times d}$ with labels $y_i$ and the corresponding (structured, for SSR/SSR+ND) text embeddings $T\in\mathbb{R}^{B\times d}$, let $S=\gamma ZT^\top$ with the learnable CLIP logit scale $\gamma$. Because several images in a batch share a class, the one-hot CLIP target is replaced by the multi-positive target
\[
Q^{I\to T}_{ij}=\frac{\mathbf{1}[y_i=y_j]}{\sum_{k=1}^{B}\mathbf{1}[y_i=y_k]},\qquad
\mathcal{L}_{I\to T}=-\frac1B\sum_{i=1}^{B}\sum_{j=1}^{B}Q^{I\to T}_{ij}\log\frac{\exp(S_{ij})}{\sum_{k=1}^{B}\exp(S_{ik})},
\]
with $\mathcal{L}_{T\to I}$ defined symmetrically, and
\[
\mathcal{L}_{\mathrm{mp}}=\tfrac12\big(\mathcal{L}_{I\to T}+\mathcal{L}_{T\to I}\big).
\]
This is a multi-positive objective, not single-positive InfoNCE. In the implementation the targets are tiled $1/(\text{positives per class})$ entries over class-balanced effective batches (200 classes $\times$ 10 images for CUB, 555 $\times$ 4 for NAB), which equals the definition above; the contrastive loss spans the full effective batch across gradient-accumulation steps. N, U, SSR, and SSR+ND all use this same loss. The remaining recipe settings are in \S\ref{app:C_recipes}.

\subsection{Canonical semantic routing masks}
\label{app:ssr_masks}

\paragraph{Routing sets.} Let $S^{\mathrm{name}}$, $S^{\mathrm{con}}$, and $S^{\mathrm{att}}_k$ be the token positions of the NAME, CONCEPT, and $k$-th ATTRIBUTE segments. Every position attends to \texttt{<sot>} and to itself; within a segment attention is causal. The additional access of each route is:
\begin{itemize}
\item \textbf{Shared-concept routing (CUB/NAB).} NAME tokens (including $r^{\mathrm{name}}$) read only earlier NAME tokens. CONCEPT tokens read only earlier CONCEPT tokens. ATTRIBUTE$_k$ tokens (including $r^{\mathrm{att}}_k$) read earlier tokens of their own segment \emph{and the CONCEPT content tokens} (excluding the concept boundary token), but neither the NAME segment nor other attribute segments. \texttt{<eot>} reads only $\{r^{\mathrm{name}}\}\cup\{r^{\mathrm{att}}_k\}_k$; it does not read the concept boundary token or content tokens directly.
\item \textbf{No-concept routing (ImageNet).} As above with no CONCEPT segment: ATTRIBUTE$_k$ tokens read only their own segment (the attribute-to-name link is off), and \texttt{<eot>} reads only $\{r^{\mathrm{name}}\}\cup\{r^{\mathrm{att}}_k\}_k$.
\end{itemize}
Consequently the ATTRIBUTE routes have no direct access to the class name, while name information reaches the fused representation through $r^{\mathrm{name}}$ (via \texttt{<eot>} and the NAME channel).

\paragraph{The concept segment.} In the shared-concept routing, a single coarse category word (e.g., ``bird'' for CUB and NAB) is placed between the NAME and ATTRIBUTE segments. Every ATTRIBUTE route reads it, so that each attribute phrase (``black crown'') is interpreted in the same coarse category context without exposing the class name; the concept is shared across all classes of the dataset and carries no class identity. It is context for the attribute routes and is not a fusion input; \texttt{<eot>} does not read it directly, so it influences the class embedding only through the attribute summaries. ImageNet has no single coarse category, so its routing has no concept segment.

\begin{figure}[t]
    \centering
    \includegraphics[width=\linewidth]{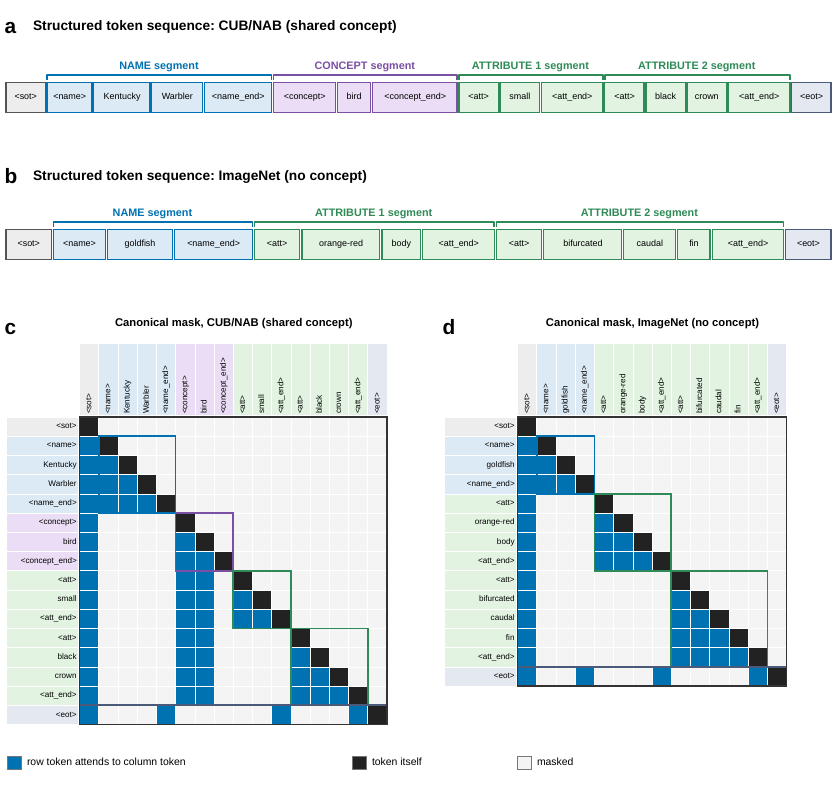}
    \caption{Structured token sequences and canonical routed-attention masks. \textbf{(a)} CUB/NAB sequence with the shared concept, showing every start/end boundary token (\texttt{<sot>}, \texttt{<name>}/\texttt{<name\_end>}, \texttt{<concept>}/\texttt{<concept\_end>}, \texttt{<att>}/\texttt{<att\_end>}, \texttt{<eot>}); the example attributes are ``small'' and ``black crown''. \textbf{(b)} ImageNet sequence without a concept (class ``goldfish'', attributes ``orange-red body'' and ``bifurcated caudal fin''). \textbf{(c, d)} Token-level masks built from the mask definitions used in training: a blue cell means the row token attends to the column token, black is the token itself, light grey is masked. Every token attends \texttt{<sot>}; attention within a segment is causal. In (c), ATTRIBUTE tokens also read the CONCEPT tokens (\texttt{<concept>} and its content, but not \texttt{<concept\_end>}) and never the NAME segment or other attribute segments; \texttt{<eot>} reads only \texttt{<name\_end>} and the \texttt{<att\_end>} tokens. In (d) there is no concept and ATTRIBUTE tokens read only their own segment. ``object'' is a P3 evaluation-template word only, never a training token.}
    \label{fig:S1_routing}
\end{figure}

Figure~\ref{fig:S1_routing} shows the token-level access pattern. SSR and SSR+ND use the identical mask; SSR+ND's name dropout acts at the fusion input, not in the mask. An implementation summary of these routings is in Table~\ref{tab:C_rawflags}.

\subsection{Controlled name-access variants}
\label{app:ssr_name_access}

\begin{figure}[t]
    \centering
    \includegraphics[width=\linewidth]{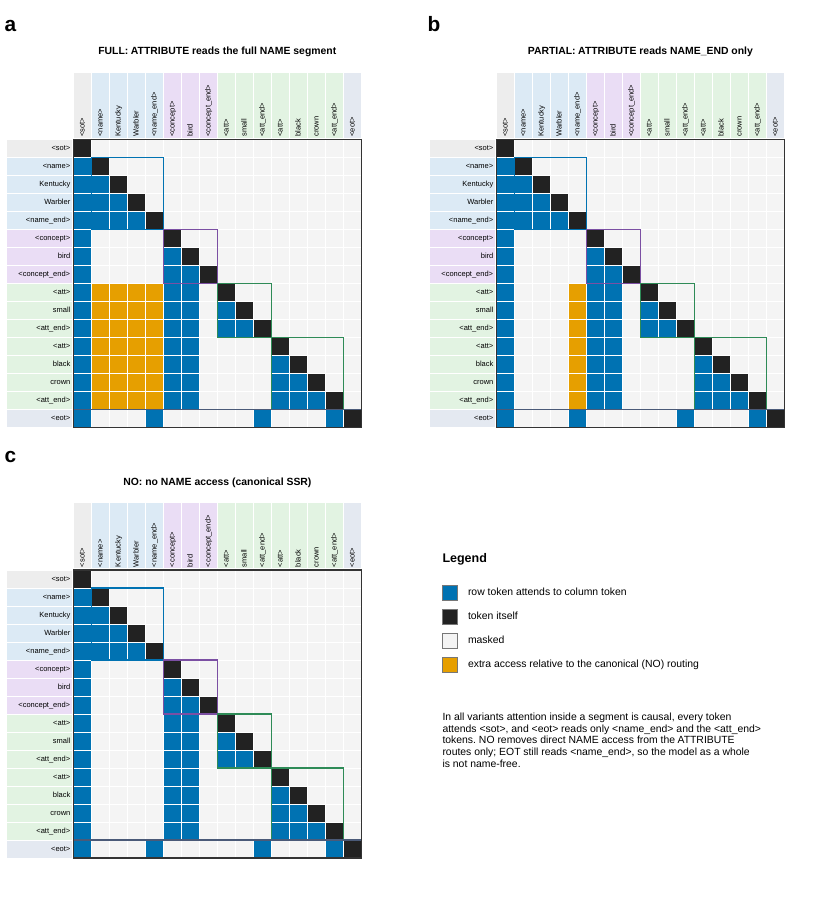}
    \caption{Token-level masks for the controlled name-access variants (direct name-access ablation; CUB/NAB shared-concept setup; same conventions as Figure~\ref{fig:S1_routing}). Orange cells mark access beyond the canonical routing. \textbf{(a) FULL:} ATTRIBUTE tokens also read all NAME tokens (\texttt{<name>}, content, \texttt{<name\_end>}). \textbf{(b) PARTIAL:} ATTRIBUTE tokens also read only the \texttt{<name\_end>} boundary token, not the class-name content. \textbf{(c) NO:} no NAME access; identical to the canonical SSR routing. In all variants \texttt{<eot>} still reads \texttt{<name\_end>}, so NO removes direct NAME access from the ATTRIBUTE route only and the model as a whole is not name-free.}
    \label{fig:S2_nameaccess}
\end{figure}

Figure~\ref{fig:S2_nameaccess} shows the three variants. They are controlled leakage/name-access analysis variants, not standard SSR recipes, and are used only for the analysis in \S\ref{app:trackA} (Tables~\ref{tab:E_trackA_p3}--\ref{tab:E_trackA_rsa}). In all cases EOT remains name-accessible through NAME\_END.

\FloatBarrier

\section{Datasets, Text Construction, and Evaluation Protocol}
\label{app:shared_protocol}

\subsection{Datasets and source descriptions}
\label{app:datasets}

We use ImageNet-1k~\citep{deng2009imagenet} and its same-ontology shifts ImageNet-V2~\citep{recht2019imagenetv2}, ImageNet-R~\citep{hendrycks2021imagenetr}, and ImageNet-Sketch~\citep{wang2019imagenetsketch}; CUB-200-2011~\citep{wah2011cub} (200 classes); NABirds~\citep{vanhorn2015nabirds} (555 classes as used here); and FineView Butterfly~\citep{onda2025fineview} (130 classes). ImageNet-trained models are evaluated on ImageNet val (target), the three shifts (same ontology), and CUB-200, NAB-348, and FineView Butterfly (cross-ontology). CUB- and NAB-trained models are evaluated on their own test sets (target) and reciprocally on the other bird dataset (\S\ref{app:B_pop}).

\subsection{LLM processing and attribute extraction}
\label{app:text_construction}
\label{app:attribute_sources}

\paragraph{Sources.} For CUB-200-2011 and NABirds we obtain field-guide-style species descriptions from Cornell Lab's \emph{All About Birds}~\citep{cornell_allaboutbirds}. For FineView Butterfly we use species descriptions from Butterflies and Moths of North America (BAMONA; \citealp{lotts2026bamona}). Because these descriptions vary in length, style, and visual detail, an LLM was used to normalize or shorten them and to extract individual observable visual attributes; these processed descriptions and attributes form the text inputs of the unstructured and structured conditions and of the semantic probes. For ImageNet, class-level descriptions and visual attributes are generated with an LLM (there is no raw web source). Where applicable, U and SSR/SSR+ND derive their semantic information from the same underlying descriptions.

\paragraph{Verified examples.} Table~\ref{tab:B_examples} shows one class per dataset with the source identity (where one exists), the extracted attributes actually used, and the resulting probe strings.

\begin{table}[t]
\centering
\scriptsize
\setlength{\tabcolsep}{3pt}\begin{tabular}{p{1.5cm}p{2.7cm}p{3.9cm}p{4.0cm}}
\toprule
Dataset / class & Source & Extracted attributes used & P1 / P3 strings \\
\midrule
CUB: Kentucky Warbler & Source: Cornell Lab, All About Birds species account~\citep{cornell_allaboutbirds} & small; olive-green upperparts; bright yellow underparts; black crown; black sideburns; yellow spectacles; no wing bars & P1: ``a photo of a Kentucky Warbler''; P3: ``a bird with \{attributes\}'' \\
NAB: female or juvenile Lesser Goldfinch & Source: Cornell Lab, All About Birds species account~\citep{cornell_allaboutbirds} & stub bill; long pointed wings; short notched tail; olive back; dull yellow underparts; black wings; whitish wingbars & P3: ``a bird with \{attributes\}'' \\
FineView: Monarch & Source: BAMONA species account~\citep{lotts2026bamona} & orange upperside; wide black borders; black veins; blurred black veins; white border spots; white wing-tip spots & P3: ``a butterfly with \{attributes\}'' \\
ImageNet: Goldfish & (LLM-generated; no raw source) & orange-red body; bright yellow body; torpedo-shaped body; large reflective scales; bifurcated caudal fin; stiff dorsal fin; smooth operculum & P3: ``a object with \{attributes\}'' \\
\bottomrule
\end{tabular}
\caption{Verified examples of source resource, extracted attributes, and probe strings. NAB classes are plumage forms (e.g., female or juvenile), split from the per-species source text by the LLM. CUB attribute list shown is the CUB-59 strict-set list.}
\label{tab:B_examples}
\end{table}

\subsection{Exact capability-probe definitions}
\label{app:B_probes}

All probes retrieve text prompts from images (image-to-text Recall@1 unless stated) through the \emph{ordinary CLIP/EOT text pathway}; none uses structured fusion.
\begin{itemize}
\item \textbf{P1} (class-name identity): \texttt{a photo of a \{class\_name\}}.
\item \textbf{P2} (joint name+attribute compatibility): \texttt{a photo of a \{class\_name\} with \{attributes\}}. P2 is class-name dominated and is not an attribute-isolating metric.
\item \textbf{P3} (name-free class-level attribute-profile retrieval): CUB/NAB \texttt{a bird with \{attributes\}}; FineView \texttt{a butterfly with \{attributes\}}; ImageNet family \texttt{a object with \{attributes\}}. The ungrammatical ``a object'' is the exact string used. P3 is not localization, spatial grounding, or part binding.
\end{itemize}
The name-swap diagnostic replaces only the class name in the P2 prompt with another class's name; the attribute-swap diagnostic replaces only the attributes with another class's attributes.

\subsection{CAPP construction and aggregation}
\label{app:B_capp}

CAPP (Counterfactual Attribute-Profile Preference) presents the \emph{same image} with a full correct attribute profile (rendered, e.g., ``a bird with black crown, bright yellow underparts, olive-green upperparts'') and a full counterfactual profile that differs in exactly one valid semantic slot/value; all other profile content is identical. The counterfactual value is the value of that slot for another class (donor). Pairwise CAPP uses one counterfactual; CAPP-Multi ranks the correct value against all valid same-slot alternatives (top-1 reported). Accuracies are macro-averaged over (class, slot) cells (class-slot-macro). A profile/slot is valid only if the slot is well-defined for the class under the strict-contrast rules used to construct the author-reviewed full-class CAPP profiles. CAPP measures controlled attribute-value discrimination; it is not spatial grounding or localization and is complementary to, not interchangeable with, P3.

\begin{quote}\small\ttfamily
for each class c, slot s with valid value v\_c[s]:\\
\hspace*{1em} correct = render(profile(c))\\
\hspace*{1em} for each donor value w != v\_c[s] (pairwise: one w; multi: all valid w):\\
\hspace*{2em} counter = render(profile(c) with slot s := w)\\
\hspace*{2em} score = [sim(img, correct) > sim(img, counter)]   \# multi: rank of correct\\
\hspace*{1em} cell(c,s) = mean over images and donors\\
CAPP = mean over cells (c,s)
\end{quote}

\subsection{Evaluation populations}
\label{app:B_pop}

CUB-trained models are evaluated cross-ontology on NAB-555 (full) and NAB-348 (strict, dataset-class-disjoint from CUB). NAB-trained models are evaluated on CUB-200 (full) and CUB-59 (strict, CUB classes that do not overlap NABirds under the frozen dataset-class mapping). Full and strict populations are reported separately and never pooled; CUB-to-NAB and NAB-to-CUB are never pooled.

\paragraph{Provenance of the strict subsets.} We use frozen strict-overlap subsets containing 348 NABirds classes and 59 CUB classes. The archived class lists and downstream evaluation artifacts are internally consistent, but the original scripts used to derive these subsets were not preserved; we therefore treat the released class lists as the authoritative evaluation artifacts.

\paragraph{Taxonomy note.} Because CUB and NABirds use historical dataset taxonomies, some legacy dataset classes may correspond to taxa that have subsequently been renamed, split, or merged in external taxonomic resources. We preserve the original dataset class identities for evaluation; attribute descriptions may therefore be shared when the external source treats those historical classes jointly. The strict split is defined operationally by the original dataset identities and the frozen mapping used in the experiments, and is not claimed to be disjoint under current taxonomy.

\subsection{Checkpoint and deployment conventions}
\label{app:B_deploy}

Deployment is each run's own checkpoint with the highest target-domain P1; it is never selected using P3, CAPP, or held-out performance. CUB/NAB trajectory locations are epochs. ImageNet checkpoints are 11 cumulative-data milestones (Table~\ref{tab:C_recipe_imagenet}), not epochs and not proportional to gradient updates; the deployment milestone of every ImageNet-trained run, for all six backbones, is the milestone with the highest ImageNet-val P1 measured on the full 50{,}000-image validation set during training, which is the final milestone (11) for all 24 runs. For OpenAI N/U/SSR the training records of the earliest milestones (1--5 for L/14, 1--6 for B/32) are not archived; these milestones are excluded from the selection, which does not affect it because their P1 on the smaller 5{,}000-image probe subset is at least 4 points below that of the final milestone. Per-milestone P1 used in the trajectory tables comes from that probe subset, whose argmax differs from the full-validation choice by one milestone for three runs (MetaCLIP H/14 SSR+ND, MetaCLIP L/14 U, OpenAI B/32 SSR; probe margins at most 0.3\,pp), so the full-validation rule is used for deployment selection throughout. PT is the pretrained checkpoint evaluated directly on the dataset named in each table block. $\mathrm{Gain}_{\mathrm{peak}}=P3_{\mathrm{peak}}-P3_{\mathrm{PT}}$, $\mathrm{Gain}_{\mathrm{deploy}}=P3_{\mathrm{deploy}}-P3_{\mathrm{PT}}$, $\mathrm{Forget}_{\mathrm{postpeak}}=P3_{\mathrm{peak}}-P3_{\mathrm{deploy}}$; a post-peak decline is called reduction of an acquired capability only when the trajectory first rose above PT.

\subsection{Bootstrap, duplicate-evaluation, and uncertainty conventions}
\label{app:B_bootstrap}

\begin{itemize}
\item \textbf{P2 name-/attribute-swap intervals (Table~\ref{tab:S9_name_dependency}):} class-level bootstrap, 1000 replicates (classes resampled with replacement, all images of sampled classes kept), 95\% percentile intervals, seed 20260917.
\item \textbf{CAPP paired comparisons and name-access-ablation CAPP contrasts:} paired class-level bootstrap with 10{,}000 resamples, same classes for both methods. No CAPP intervals appear in the main tables.
\item \textbf{Geometry correlations (Table~\ref{tab:G_geom}):} class-level Spearman correlations with 10{,}000 class resamples for 95\% intervals.
\item \textbf{ImageNet duplicate-evaluation variability:} repeated evaluation of the same checkpoint differs by about 0.18\,pp per run (repeated-evaluation checks); differences below roughly 0.5\,pp are interpreted cautiously.
\item \textbf{No inferential tests across backbones or methods are reported.} The six backbones are heterogeneous, not replicates; bootstrap intervals quantify evaluation-set uncertainty, and multi-seed runs (\S\ref{app:multiseed}) quantify training stochasticity only for the configurations actually repeated.
\end{itemize}

\subsection{Zero-shot motivation: description dilution}
\label{app:B_zeroshot}

This subsection documents the pretrained, zero-shot observation cited in the Introduction. It is a motivational analysis of frozen pretrained encoders and is not part of the fine-tuning trajectory experiments; it should not be extended to fine-tuned models. Table~\ref{tab:B_zeroshot} reports zero-shot CUB accuracy for two pretrained backbones under three text inputs: the class name alone, a concept-plus-class-name prompt, and the concept-plus-class-name prompt followed by a field-guide-style description of about 77 tokens. For both backbones, accuracy is highest with the class name alone and declines as more text is added. Descriptions longer than the standard 77-token context window can only be processed without truncation by a long-context variant of each backbone (TULIP~\citep{tulip2024}); both variants decline further as description length grows, more steeply for the DFN-5B-based variant. Providing more semantic text does not by itself make that information more accessible through the resulting representation. The descriptions are unprocessed field-guide text rather than the extracted attribute lists used for P2/P3, and this result does not compare against shuffled or wrong-class text.

\begin{table}[!htbp]
\centering
\small
\setlength{\tabcolsep}{6pt}
\begin{tabular}{lcc}
\toprule
Text input & DFN5B ViT-H/14-378 & OpenAI ViT-L/14 \\
\midrule
Class name only & \textbf{88.80} & \textbf{57.18} \\
Concept + class name & 87.88 & 55.93 \\
Concept + class name + description ($\sim$77 tokens) & 84.77 & 50.07 \\
\midrule
\multicolumn{3}{l}{\textit{Long-context TULIP variant of the same backbone}} \\
Concept + class name + description (200--300 tokens) & 36.02 & 41.50 \\
Concept + class name + description (100--500 tokens) & 28.55 & 39.30 \\
\bottomrule
\end{tabular}
\caption{Zero-shot CUB retrieval accuracy (\%) for two pretrained backbones as text is added to the class-name prompt (frozen encoders, evaluation only). The lower block uses the long-context TULIP variant~\citep{tulip2024} of each backbone, which can process descriptions beyond the 77-token window without truncation.}
\label{tab:B_zeroshot}
\end{table}

\FloatBarrier

\section{Training and Implementation Details}
\label{app:training}

\subsection{Backbone initialization and pretrained sources}
\label{app:C_backbones}

\begin{table}[t]
\centering
\small
\begin{tabular}{lll}
\toprule
Backbone & open\_clip model & Pretrained tag \\
\midrule
DFN5B H/14-378 & ViT-H-14-378-quickgelu & dfn5b \\
DFN2B L/14 & ViT-L-14 & dfn2b\_s39b \\
MetaCLIP H/14 & ViT-H-14-quickgelu & metaclip\_fullcc \\
MetaCLIP L/14 & ViT-L-14-quickgelu & metaclip\_fullcc \\
OpenAI L/14 & ViT-L-14-quickgelu & openai \\
OpenAI B/32 & ViT-B-32-quickgelu & openai \\
\bottomrule
\end{tabular}
\caption{Pretrained initializations (open\_clip model names and pretrained tags read from the run parameter files). Backbone descriptions and parameter counts are in Table~\ref{tab:S6_panel}.}
\label{tab:C_backbones}
\end{table}

\subsection{Training recipes by dataset/domain}
\label{app:C_recipes}

All runs use the open\_clip trainer with a multi-positive contrastive loss (temperature 1, mixing coefficient 0.5), bf16 mixed precision, per-GPU batch 20 with gradient accumulation, seed 0, no additional augmentation, text context extended from 77 to 600 tokens (dynamic NTK, scale factor 8), and no frozen image encoder. Tables~\ref{tab:C_recipe_cubnab} and~\ref{tab:C_recipe_imagenet} give the settings read from the run parameter files; they are parent-specific and \emph{not} normalized across runs.

\input{supplement/tables/audit/C_recipe_imagenet.tex}
\input{supplement/tables/audit/C_recipe_cubnab.tex}

\subsection{Backbone-specific hyperparameters}
\label{app:C_bbspecific}

The following deviations from a single recipe were verified in the parameter files and are preserved in the tables: (i) OpenAI L/14 and B/32 CUB runs use weight decay 0.2 and warmup 10000 (DFN/MetaCLIP CUB: 0.1 and 1000; OpenAI NAB: 0.1 and 1000); (ii) DFN5B CUB SSR and SSR+ND use a structured-layer learning rate of $5\times10^{-6}$, whereas all other CUB/NAB structured runs use $10^{-5}$ (ImageNet: $5\times10^{-6}$ throughout); (iii) the canonical DFN5B CUB SSR+ND run differs from the SSR run only by the added name-drop (weight decay 0.1); an earlier historical NAME-drop calibration used weight decay 0.2 and is reported separately in \S\ref{app:C_pcal}, but is not used in the canonical main results; (iv) the ImageNet OpenAI N/U/SSR parameter files were located only for steps 6--11 (L/14) and 7--11 (B/32).

\subsection{Structured fusion and routing implementation}
\label{app:C_fusion}

\begin{table}[t]
\centering
\scriptsize
\setlength{\tabcolsep}{3pt}\begin{tabular}{p{2.0cm}p{5.0cm}p{5.0cm}}
\toprule
Method & CUB/NAB & ImageNet \\
\midrule
N & Ordinary causal text pathway; class-name-only text & same \\
U & Ordinary causal text pathway; unstructured descriptive text (class name and attributes in one stream) & same \\
SSR & Routed causal attention over NAME and ATTRIBUTE segments; shared CONCEPT segment (context only, not fused); ATTRIBUTE summary pooled from the \texttt{<att\_end>} states; learned-gate Structured Text Fusion of NAME, ATTRIBUTE and EOT summaries (training only); the six added start/end marker embeddings (NAME, ATTRIBUTE, CONCEPT) are randomly initialized and trained & Same routing and fusion; no CONCEPT segment; the four added start/end marker embeddings (NAME, ATTRIBUTE) are randomly initialized and trained \\
SSR+ND & SSR with NAME-summary dropout $p=0.5$ at training-time fusion & same \\
Name-access ablation & FULL: the ATTRIBUTE route reads the full NAME segment; PARTIAL: it reads only the NAME\_END token; NO: no direct NAME access (identical to canonical SSR) & -- \\
\bottomrule
\end{tabular}
\caption{Implementation summary of the training-time routings (reproducibility mapping; all conditions are evaluated with the ordinary text encoder and EOT embedding).}
\label{tab:C_rawflags}
\end{table}

The gate and AIP parameters are learned during training only; evaluation never invokes fusion (\S\ref{app:ssr_fusion}). For the text-frozen U control used in \S\ref{app:mech}, the entire text encoder including layer norms is locked and only the image encoder and logit scale are trained.

\subsection{Checkpoint schedules}
\label{app:C_sched}

ImageNet: 11 chained milestones at cumulative data fractions 0.25, 0.5, 1, 2, 5, 10, 20, 40, 60, 80, 100\% (Table~\ref{tab:C_recipe_imagenet}). CUB/NAB: 100-epoch runs evaluated at dense epochs; the name-access ablation and hard-negative runs were evaluated at epochs 0, 4, 8, 12, 20, 30, 40, 50, 60, 75, 100. The canonical DFN5B/CUB SSR+ND run was evaluated at 8 checkpoints (epochs 0, 4, 8, 12, 20, 60, 75, 100), so its peak-epoch resolution is coarser than for the other methods.

\subsection{Multi-seed robustness checks}
\label{app:multiseed}

We report available multi-seed runs as robustness checks rather than treating the six pretrained backbones as stochastic replicates. Some multi-seed runs were conducted under earlier SSR routing variants; where the structured configuration differs from the canonical SSR routing used in the main paper, we report those runs separately and do not pool them with the current trajectory results.

An audit of all archived multi-seed groups found five seeds (10/20/30/40/50) for N, U, and an earlier SSR routing variant, on DFN5B H/14-378 only, for CUB-200, NABirds, and FineView. \textbf{No archived group matches the current canonical SSR recipe} (current: shared-concept routing, LR $5\times10^{-6}$), and there are no multi-seed runs for SSR+ND, ImageNet, or any other backbone; hence there are no ``current-recipe replications''. All groups are labeled \emph{historical matched robustness checks}. The metric is in-domain validation R@1 at the best checkpoint; P1/P2/P3/CAPP were not evaluated for these runs, so they say nothing about the variance of the current P3/CAPP trajectories. Ordinary in-domain retrieval differences between methods are small relative to, or comparable to, seed-to-seed variation. NABirds runs were walltime-limited and are not converged, and are not an exact stochastic-uncertainty estimate for any result in the paper. Obtaining current-recipe multi-seed numbers would require new runs with the canonical configuration.

\input{supplement/tables/audit/C_multiseed.tex}
\input{supplement/tables/audit/C_multiseed_perseed.tex}

\subsection{Compute and software notes}
\label{app:C_compute}

Each training run used one GPU (one process), 8 CPU cores, and 100\,GB memory. Slurm accounting records identify NVIDIA A100 or H100 GPUs for most runs; a minority have untyped GPU allocations from a partition currently comprising B200 nodes, so GPU type varied across runs. Runs used Python 3.10 and a local, untracked open\_clip fork reporting version 2.24.0. The current environment contains PyTorch 2.8.0+cu128 with CUDA 12.8; package versions were not logged by individual runs. No code revision identifier is available because the experimental fork was not under version control.

\FloatBarrier

\subsection{Historical NAME-drop probability calibration}
\label{app:C_pcal}

This subsection reports an earlier \emph{historical, single-backbone} calibration of the NAME-summary dropout probability $p$ (DFN5B H/14-378 fine-tuned on CUB-200 or on NABirds-555; $p\in\{0,0.25,0.5,0.75,1.0\}$; one run per value), evaluated with the P1/P2/P3 definitions of \S\ref{sec:capability_measurement}. It addresses why $p{=}0.5$ was used and whether the effect of NAME-drop depends on that one value. It is not a main result: it is not pooled with the six-backbone counts, timing statistics, Table~\ref{tab:deploy_profile} medians or canonical trajectories, and, being one run per value, it does not estimate stochastic variance. The runs are matched historical calibration runs that vary $p$, with the other training settings recorded in their parameter files identical except as follows. In the CUB sweep the $p{=}0$ point is the baseline SSR run (weight decay 0.1, warmup 1000), whereas the four $p>0$ runs used weight decay 0.2 (warmup 10000, which is inert under fine-tuning from a pretrained checkpoint; 4 rather than 8 data-loader workers; an additional epoch-90 checkpoint), so the $p{=}0$ endpoint is not weight-decay matched. In the NABirds sweep all runs used weight decay 0.1; the $p{=}0$ run was evaluated only through epoch 40 whereas the other four were evaluated to epoch 100.

\paragraph{Target-domain P3 increases with $p$.} Table~\ref{tab:C_pcal_peak} gives the maximum observed P3 by $p$. The maximum observed target-domain P3 over the saved checkpoints increases monotonically with $p$ in both sweeps: CUB-200 from 40.42 to 50.35 (+9.93\,pp) and NAB-555 from 15.29 to 17.93 (+2.64\,pp). The CUB trend is also monotone among the four weight-decay-matched $p>0$ runs alone (44.60 to 50.35). P1 and P2 remain near their ceiling in this calibration (CUB-trained peak P1 90.37 to 90.01, peak P2 90.16 to 89.97).

\paragraph{Held-out effects are smaller and mixed.} Held-out peak P3 changes far less than the in-domain values: CUB-trained to NAB-348 rises from 17.35 to 18.07, CUB-trained to FineView is flat to slightly lower (9.90 to 9.52), NABirds-trained to CUB-59 rises from 53.80 to 56.10, and NABirds-trained to FineView is flat (10.94 to 10.58). We therefore do not read these data as a monotonic transfer benefit of larger $p$.

\paragraph{Held-out retention.} For the CUB-trained sweep, Table~\ref{tab:C_pcal_ret} gives the relative decrease of held-out P3 from its peak to the P1-selected checkpoint. Moderate dropout ($p{=}0.25$--$0.5$) has the smallest decrease on both held-out sets (NAB-348 18.8--19.5\% versus 23.1\% at $p{=}0$ and 20.9\% at $p{=}1$; FineView 7.9--9.5\% versus 12.4\% and 15.4\%), a U-shaped pattern in post-peak decrease (equivalently, an inverted-U retention pattern) in this single run per value. The NABirds-trained retention sweep is not reported because its $p{=}0$ run has shorter evaluation coverage (epoch 40 versus 100), which makes the comparison unreliable.

\paragraph{Reading of the calibration.} The historical single-backbone calibration shows that increasing NAME-summary dropout strengthens target-domain P3, while held-out retention is best at moderate dropout rather than increasing monotonically. We therefore treat $p{=}0.5$ as a reasonable operating point rather than a globally optimized value.

\paragraph{Relation to the canonical $p{=}0.5$ run.} The historical CUB-trained $p{=}0.5$ calibration run used the earlier weight-decay-0.2 recipe, whereas the canonical DFN5B/CUB SSR+ND run uses weight decay 0.1 (it is a one-flag clone of the baseline SSR run). At $p{=}0.5$ the historical and canonical runs show closely matching trajectories on the compared checkpoints (epochs 0, 4, 8, 12, 20, 75 and 100): CUB P1 differs by at most 0.2\,pp and CUB P3 by at most 0.27\,pp (for example 45.56 versus 45.63 at epoch 20 and 46.82 versus 46.55 at epoch 100), CUB CAPP by at most 0.10\,pp (pairwise) and 0.14\,pp (multi-way), and NAB-348 pairwise CAPP by at most 0.1\,pp; P2 and FineView were not compared. The two runs also differ in their inert warmup setting, data-loader workers and one extra checkpoint epoch. This comparison is a robustness bridge indicating that the $p{=}0.5$ behavior does not appear to be peculiar to the older weight-decay-0.2 recipe. The two runs are not stochastic replicates, the comparison is not a controlled weight-decay ablation, and it does not show that the full $p$ sweep would be unchanged under weight decay 0.1.

\input{supplement/tables/audit/C_pcal_peak.tex}
\input{supplement/tables/audit/C_pcal_ret.tex}

\section{Capability Acquisition Across Backbones}
\label{app:acquisition}

This section gives the numerical support for Figure~\ref{fig:acquisition}a--b and \S\ref{sec:acquisition}. PT means the pretrained model evaluated directly on the evaluation dataset of each table block, never carried over from the training ontology. ImageNet P3 uses the exact evaluation template ``a object with \{attributes\}''. Numbers in parentheses are ImageNet milestone indices (1--11), not epochs.

\input{supplement/tables/generated_supplement_tables.tex}

\FloatBarrier

\section{Supervision Determines the Acquired Capability}
\label{app:supervision}

This section supports Table~\ref{tab:deploy_profile}, Figure~\ref{fig:supervision}, and \S\ref{sec:capability_differences}. The CAPP tables report all six backbones (the DFN/MetaCLIP backbones plus the additional OpenAI L/14 and B/32 backbones, marked with a divider) under the same matched protocol; the six are not pooled as replicates. CUB target structured improvements are consistent, whereas NAB-target and held-out effects are backbone-dependent, and held-out CAPP gains are not universal.

\input{supplement/tables_sec5/generated_section5_supplement.tex}
\begin{figure}[!htbp]
\centering
\includegraphics[width=\linewidth]{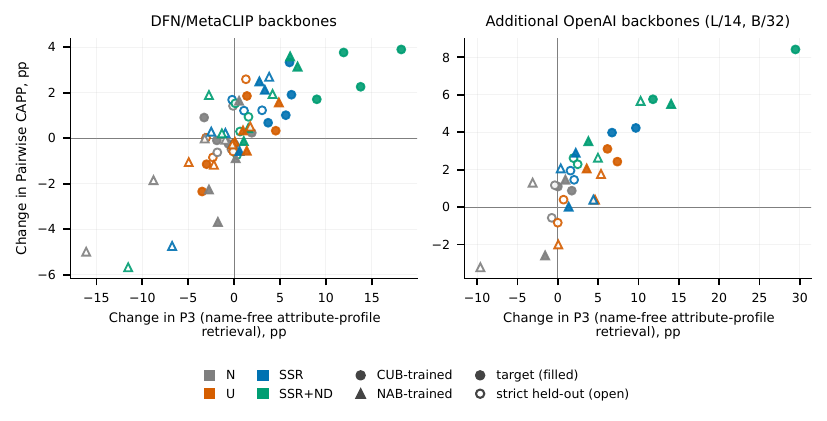}
\caption{\textbf{Name-free attribute-profile retrieval (P3) change versus Pairwise CAPP change} across intervention conditions, each relative to the pretrained model, at every run's own P1-selected deployment checkpoint. Left: DFN/MetaCLIP backbones; right: additional OpenAI L/14 and B/32 backbones (never pooled). Colors: N, U, SSR, SSR+ND; circles CUB-trained and triangles NABirds-trained; filled markers are the training-domain target and open markers the strict dataset-class-disjoint held-out set (NAB-348 for CUB-trained, CUB-59 for NABirds-trained; full held-out sets are not plotted). Across many intervention conditions, gains in name-free attribute-profile retrieval and local counterfactual discrimination tend to move in the same direction, although the probes measure distinct capabilities. No correlation coefficient is computed. Delta values use the pretrained result from the same evaluation job as each point and may therefore differ slightly from differences reconstructed from separately tabulated evaluations.}
\label{fig:S_p3_capp_scatter}
\end{figure}

\paragraph{Direct name-access ablation results.} Tables~\ref{tab:E_trackA_p3}--\ref{tab:E_trackA_rsa} give the FULL/PARTIAL/NO results (single runs, DFN5B H/14-378, CUB-trained). Restricting direct NAME access from the ATTRIBUTE route increases target P3 and CAPP and makes the attribute route more similar to the pretrained name-free profile geometry and less similar to class-name geometry, while ordinary P2 still collapses under name-swap in every variant. NO is not globally name-free: EOT remains globally name-accessible. The name-access effect of this ablation and the late-specialization analyses in \S\ref{app:mech} are mechanistically distinct and should not be conflated. Caveats: PARTIAL deploys at epoch 75 versus epoch 20 for the others; NAB differences of about 1\,pp should not be over-read.

\label{app:trackA}
\input{supplement/tables/audit/E_trackA_p3.tex}
\input{supplement/tables/audit/E_trackA_capp.tex}
\input{supplement/tables/audit/E_trackA_p2.tex}
\input{supplement/tables/audit/E_trackA_rsa.tex}
\begin{figure}[!htbp]
\centering
\includegraphics[width=0.96\linewidth]{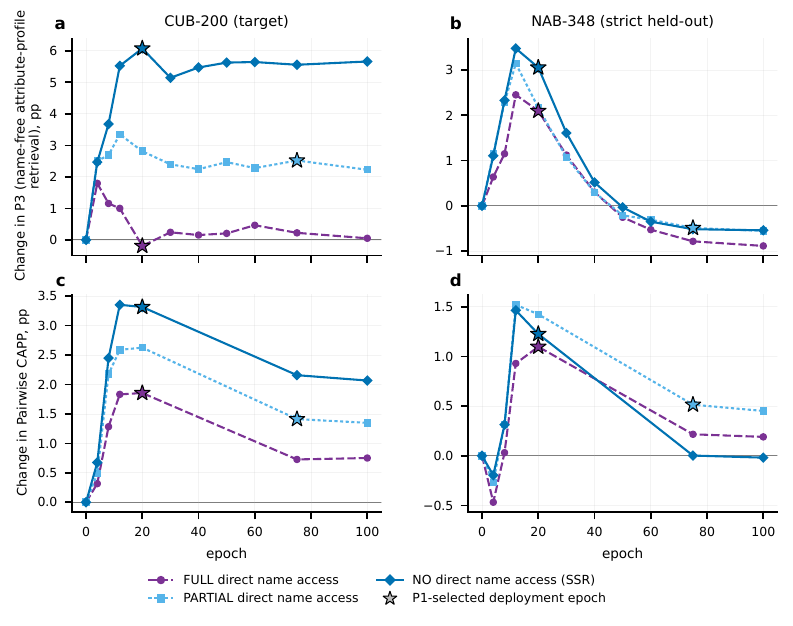}
\caption{\textbf{Direct name-access ablation trajectories,} DFN5B H/14-378 fine-tuned on CUB-200 (one run per variant, single seed). Rows: P3 (name-free attribute-profile retrieval) and Pairwise CAPP; columns: CUB-200 (target) and strict held-out NAB-348. FULL, PARTIAL and NO direct name access denote how much of the NAME segment the ATTRIBUTE routes may read (Appendix~\ref{app:ssr_name_access}); NO is the canonical SSR run. Values are the change from each run's own pretrained epoch-0 evaluation; stars mark each run's P1-selected deployment epoch (FULL 20, PARTIAL 75, NO 20). P3 is evaluated at 11 epochs and CAPP at 7 (0, 4, 8, 12, 20, 75, 100), so the CAPP segment between epochs 20 and 75 is a straight line across unmeasured epochs. Restricting direct name access from the ATTRIBUTE route changes P3 and CAPP acquisition and trajectory behavior, supporting a role for representation and routing rather than content alone; NO is not name-free globally, because the EOT summary still reads NAME\_END, and PARTIAL deploys later, so deployment-level comparisons are timing-confounded.}
\label{fig:S_trackA_traj}
\end{figure}

\FloatBarrier

\section{Capability-Specific Timing and Specialization}
\label{app:timing}

\subsection{P1 and P3 trajectory atlas}
\label{app:F_atlas}
\begin{figure}[!htbp]
\centering
\includegraphics[width=\linewidth]{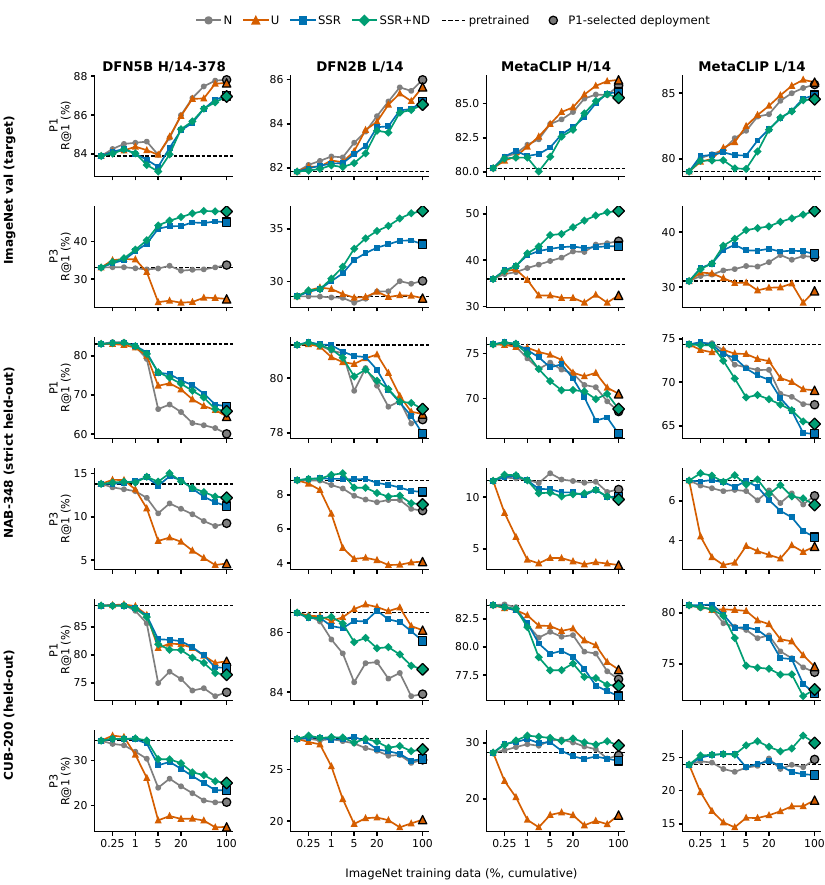}
\caption{\textbf{ImageNet-trained P1/P3 trajectories, DFN/MetaCLIP backbones} (columns: DFN5B H/14-378, DFN2B L/14, MetaCLIP H/14, MetaCLIP L/14). For each evaluation population, the top row is P1 and the bottom row P3: ImageNet val (target), NAB-348 (strict dataset-class-disjoint held-out) and CUB-200 (held-out). The horizontal axis is the cumulative ImageNet training-data fraction at the 11 geometric milestones (not epochs). Deployment markers show the ImageNet-val-P1-selected checkpoint, which is shared by every population of a run; P3 uses the templates in Appendix~\ref{app:B_probes} (ImageNet: \texttt{a object with \{attributes\}}; bird evaluations: \texttt{a bird with \{attributes\}}).}
\label{fig:S_atlas_in_core}
\end{figure}

\begin{figure}[!htbp]
\centering
\includegraphics[width=0.72\linewidth]{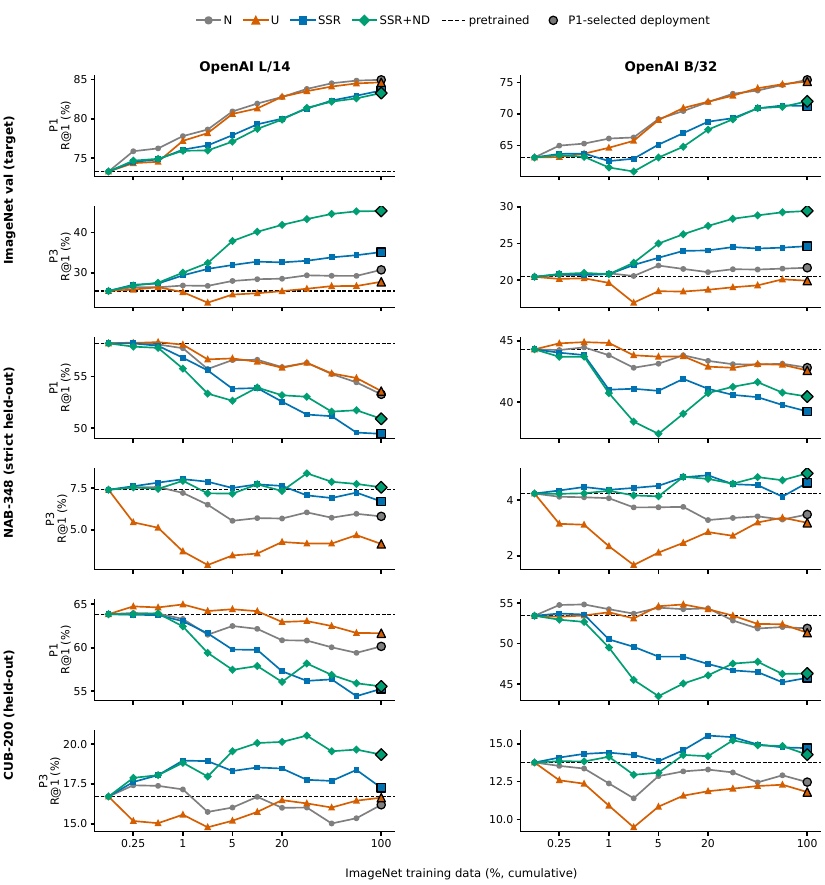}
\caption{\textbf{ImageNet-trained P1/P3 trajectories, additional OpenAI backbones} (OpenAI L/14 and B/32; the two families are never pooled). Layout, populations and definitions as in Figure~\ref{fig:S_atlas_in_core}. The OpenAI deployment checkpoint is the ImageNet-val-P1 argmax over the 11 milestones; for L/14 it is the final milestone.}
\label{fig:S_atlas_in_openai}
\end{figure}

\begin{figure}[!htbp]
\centering
\includegraphics[width=\linewidth]{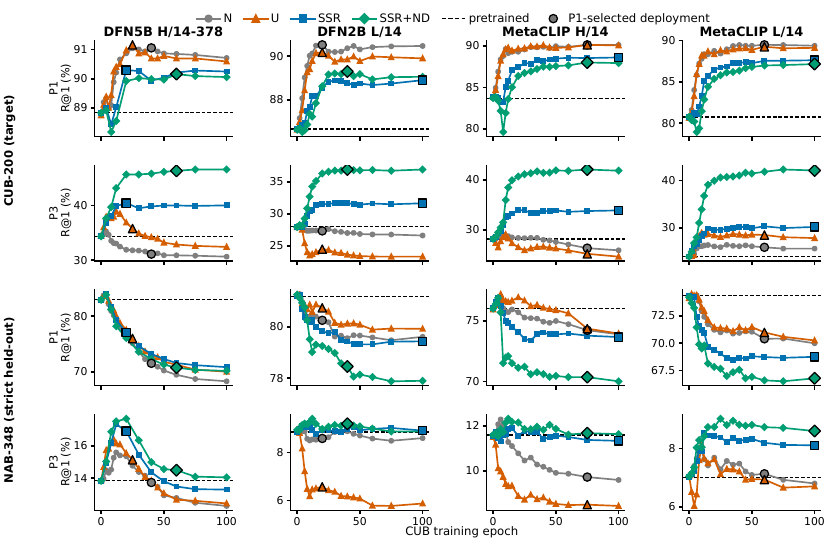}
\caption{\textbf{CUB-trained P1/P3 trajectories, DFN/MetaCLIP backbones.} Rows: CUB-200 (target) and NAB-348 (strict dataset-class-disjoint held-out), each with P1 above P3. Horizontal axis: CUB training epoch. Deployment markers show each run's CUB-P1-selected checkpoint. The DFN5B SSR+ND curve is the canonical SSR+ND run; dense fine-grained trajectories for the DFN/MetaCLIP backbones come from the per-epoch trajectory evaluations rather than a consolidated table, and the full NAB-555 held-out population has no P3 trajectory for the DFN/MetaCLIP backbones (it is available for OpenAI in Figure~\ref{fig:S_atlas_fg_openai}).}
\label{fig:S_atlas_fg_cub}
\end{figure}

\begin{figure}[!htbp]
\centering
\includegraphics[width=\linewidth]{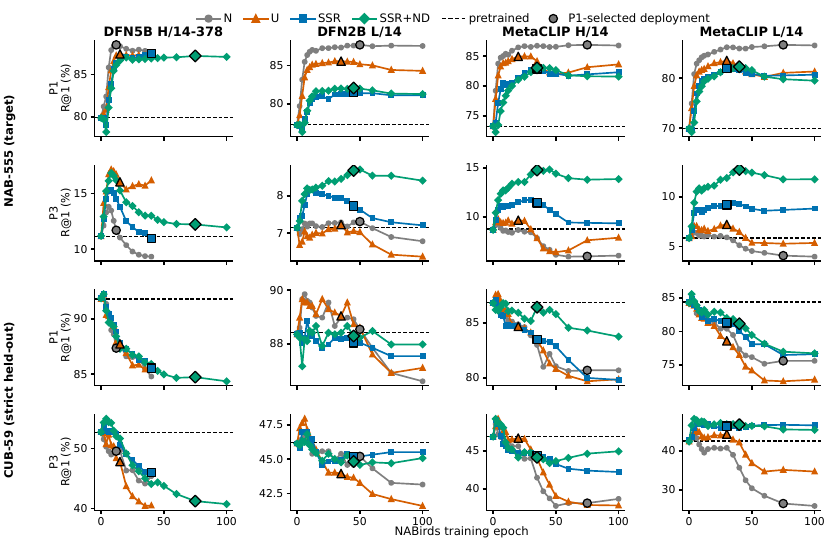}
\caption{\textbf{NABirds-trained P1/P3 trajectories, DFN/MetaCLIP backbones.} Rows: NAB-555 (target) and CUB-59 (strict dataset-class-disjoint held-out), each with P1 above P3. Horizontal axis: NABirds training epoch (DFN5B N/U/SSR were evaluated to epoch 40). Deployment markers show each run's NAB-P1-selected checkpoint. Definitions and single-run caveats as in Figure~\ref{fig:S_atlas_fg_cub}.}
\label{fig:S_atlas_fg_nab}
\end{figure}

\begin{figure}[!htbp]
\centering
\includegraphics[width=\linewidth]{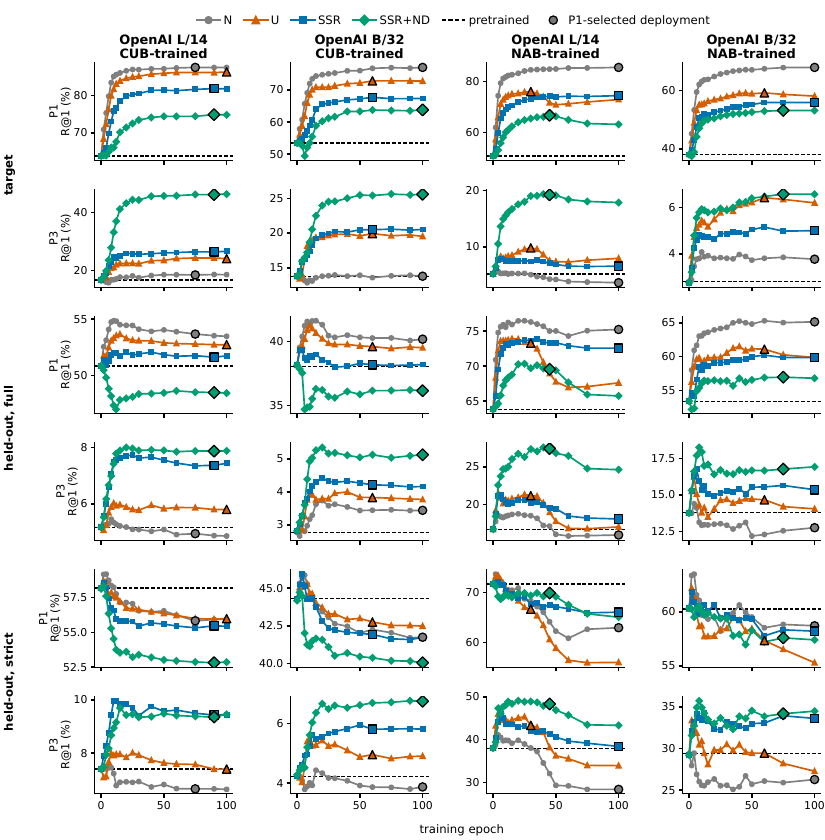}
\caption{\textbf{OpenAI fine-grained P1/P3 trajectories} (columns: OpenAI L/14 and B/32, CUB-trained and NABirds-trained). Rows: the training-domain target (CUB-200 or NAB-555), the full held-out population (NAB-555 or CUB-200; not dataset-class-disjoint) and the strict held-out population (NAB-348 or CUB-59; dataset-class-disjoint), each with P1 above P3; full and strict populations are never pooled. Deployment markers show each run's target-domain-P1-selected epoch (argmax over the saved epochs after epoch 0). All four methods are single runs; OpenAI CUB-trained runs use weight decay 0.2 for all four methods.}
\label{fig:S_atlas_fg_openai}
\end{figure}

Figures~\ref{fig:S_atlas_in_core}--\ref{fig:S_atlas_fg_openai} show raw P1 (class-name retrieval) and P3 (name-free attribute-profile retrieval) trajectories in matched rows, so that acquisition, capability-specific peak timing, post-peak decline, the divergence between P1 and P3, and the position of the deployment checkpoint can be read directly. All values are absolute image$\to$text R@1 (\%) at the saved checkpoints only, connected by straight lines without interpolation; the dashed line is the pretrained model; the large outlined marker on each curve is that run's existing P1-selected deployment checkpoint (target-domain P1; no separate P1 optimum is selected, and curves are not normalized). Every panel is a single run per method and backbone, and the backbones are heterogeneous models, not replicates. P2 trajectories are omitted here because P2 has a different role in the paper. Not every trajectory declines: for OpenAI L/14 trained on ImageNet the deployment checkpoint is the final evaluated milestone and coincides with the P3 maximum for all four conditions on the target and same-ontology evaluations (\S\ref{app:F_openai}; those populations are not plotted here).

\subsection{Representative P3 / CAPP / deployment timing}
\label{app:F_repr}

\begin{figure}[!htbp]
\centering
\includegraphics[width=0.62\linewidth]{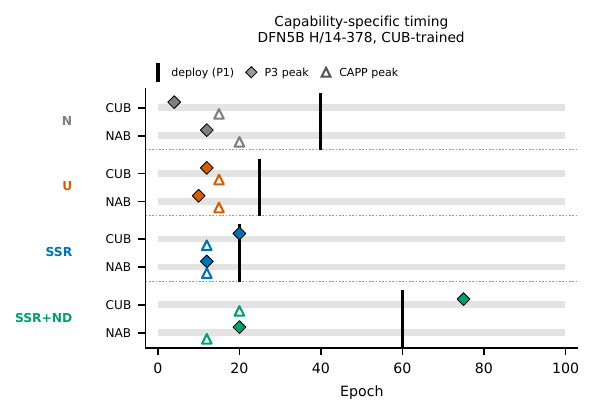}
\caption{P3 peak (diamond), Pairwise-CAPP peak (open triangle), and P1-selected deployment epoch (tick) for each method and evaluation set, DFN5B H/14-378 fine-tuned on CUB-200 (single run per method). The three need not coincide.}
\label{fig:S_capp_timing}
\end{figure}

Table~\ref{tab:S_capp_timing} gives the P3 peak epoch, Pairwise-CAPP peak epoch, and P1-selected deployment epoch for the DFN5B H/14-378 CUB-trained runs (plotted in Figure~\ref{fig:S_capp_timing}) visualizes these trajectories. This CAPP-timing observation is a representative single-run result; CAPP peak timing was not evaluated across six backbones. Values are consistent with the reported trajectory results. The SSR+ND run has 8 dense checkpoints versus 11 for the other methods, so its peak-epoch resolution is coarser.

\begin{table}[t]
\centering
\small
\begin{tabular}{llrrrr}
\toprule
Method & Eval. & P3 peak (ep.) & P3 deploy (ep.) & CAPP peak (ep.) & CAPP deploy (ep.) \\
\midrule
N & CUB-200 (target) & 4 & 40 & 15 & 40 \\
N & NAB-348 (held-out) & 12 & 40 & 20 & 40 \\
U & CUB-200 (target) & 12 & 25 & 15 & 25 \\
U & NAB-348 (held-out) & 10 & 25 & 15 & 25 \\
SSR & CUB-200 (target) & 20 & 20 & 12 & 20 \\
SSR & NAB-348 (held-out) & 12 & 20 & 12 & 20 \\
SSR+ND & CUB-200 (target) & 75 & 60 & 20 & 60 \\
SSR+ND & NAB-348 (held-out) & 20 & 60 & 12 & 60 \\
\bottomrule
\end{tabular}
\caption{Dense P3 and Pairwise-CAPP peak/deployment epochs, DFN5B H/14-378 fine-tuned on CUB-200, evaluated on CUB-200 (target) and NAB-348 (strict held-out); underlies Figure~\ref{fig:S_capp_timing}. Deployment epoch is each method's own P1-selected checkpoint, shared by construction between the P3 and CAPP columns. CAPP is Pairwise class-slot-macro accuracy. Single run per method.}
\label{tab:S_capp_timing}
\end{table}

\subsection{Six-backbone before/at/after deployment counts}
\label{app:F_counts}

Peak and deployment milestones are compared \emph{categorically} (before / at / after) within each run's own schedule; absolute epochs or milestone indices are never pooled across incompatible schedules. Summed over all six backbones and four methods, the counts (before/at/after) are 11/13/0 for the ImageNet target (24 cells), 40/32/0 for the same-ontology shifts (72 cells), and 65/7/0 for the cross-ontology evaluations (72 cells). Target and same-ontology optima occur at or after deployment more often, whereas cross-ontology peaks occur before deployment in the large majority of cells.

\subsection{Prevalence and magnitude of early P3 optima}
\label{app:F_prevmag}
Table~\ref{tab:F_prev} gives the full before/at/after timing counts by population and method (target 11/13/0 of 24, same-ontology 40/32/0 of 72, cross-ontology 65/7/0 of 72), which underlie Figure~\ref{fig:specialization}b; only target and cross-ontology are plotted there. Table~\ref{tab:F_mag} gives the peak-to-deployment P3 decrease among cells whose peak precedes deployment (Figure~\ref{fig:specialization}c shows the cross-ontology cells), including the $>0.5$\,pp counts, the target summaries that are too small to plot, and cross-ontology medians by dataset (CUB-200 1.78\,pp, NAB-348 1.76\,pp, FineView Butterfly 0.58\,pp).
\input{supplement/tables/audit/F_prev.tex}
\input{supplement/tables/audit/F_mag.tex}

\subsection{Per-backbone timing reconstruction}
\label{app:F_perbb}

Table~\ref{tab:F_counts} gives the per-backbone counts whose column sums reproduce the population totals (Table~\ref{tab:F_prev}) behind Figure~\ref{fig:specialization}b. Per-cell peak and deployment milestones are in Tables~\ref{tab:S1_full}--\ref{tab:S2_full}.

\subsection{OpenAI L/14 evaluated-trajectory boundary case}
\label{app:F_openai}

For OpenAI L/14 the P1-selected deployment checkpoint is the final evaluated ImageNet milestone (11), which also coincides with the P3 maximum in all 16 target and same-ontology cells (Table~\ref{tab:F_openai}); hence no post-peak decline was observed within the evaluated trajectory. This describes the audited schedule and does not establish that specialization cannot occur for this backbone.

\input{supplement/tables/generated_timing_tables.tex}

\FloatBarrier

\section{Analyses of Late Specialization}
\label{app:mech}

Except for \S\ref{app:G_move} (broad trajectory grid) and \S\ref{app:G_swapext} (additional swaps on other runs), all results in this section are single-run observations for the representative DFN5B H/14-378 SSR+ND model fine-tuned on CUB-200 and evaluated on NAB-348 (and CUB-200 where stated), plus an independent text-frozen U run where stated. They are not claimed to generalize across backbones or training domains, and no replication is implied. The preferred interpretation is reduced cross-modal semantic accessibility; the analyses do not show that semantic information is erased.

\subsection{Broad direct-movement analysis (N/U/SSR grid)}
\label{app:G_move}
This subsection supports Figure~\ref{fig:mechanism}a and, unlike the other subsections, covers the broad N/U/SSR trajectory grid (4 backbones $\times$ 3 training domains $\times$ 3 methods; 120 train/eval/backbone/method cells, 84 held-out; SSR+ND is not included). For each evaluation distribution the P3-peak checkpoint is that distribution's own P3 maximum (epoch 0 excluded) and the deployment checkpoint is the run's P1-selected checkpoint; no checkpoint was substituted. Table~\ref{tab:G_move_order} gives the temporal order of the two checkpoints; the primary late-specialization analysis uses only $t_{P3}<t_{\mathrm{deploy}}$ cells.

\paragraph{Movement definitions.} For each transformer block $l$ of a encoder, $D_{\mathrm{pre}}(l)=\|W_{\mathrm{peak}}(l)-W_{\mathrm{PT}}(l)\|_F/\|W_{\mathrm{PT}}(l)\|_F$ and $D_{\mathrm{post}}(l)=\|W_{\mathrm{deploy}}(l)-W_{\mathrm{peak}}(l)\|_F/\|W_{\mathrm{PT}}(l)\|_F$, averaged over blocks; both use the pretrained norm so the two phases are directly comparable, and both are net checkpoint displacements, not SGD path lengths. Image movement uses \texttt{visual.transformer.resblocks.*} (32 blocks, 629.7M parameters for DFN5B and MetaCLIP H/14; 24 blocks, 302.3M for DFN2B and MetaCLIP L/14). Text movement uses the shared pretrained text transformer \texttt{transformer.resblocks.*} (24 blocks, 302.3M parameters; 12 blocks, 85.1M). Excluded from the text metric because they have no pretrained counterpart or are not comparable across N/U/SSR: AIP and StructuredTextFusion parameters, the 4--6 newly appended SSR token-embedding rows, positional embeddings, and the logit scale. The main-text quantity is the joint pretrained-backbone movement $D_{\mathrm{joint}}=(D_{\mathrm{image}}+D_{\mathrm{text}})/2$ (not total model movement).

\paragraph{Results.} Pre-peak movement has no consistent relationship with acquisition for either encoder: the sign differs by training domain (image: pooled $r=-0.17$; CUB $+0.49$, NABirds $-0.72$, ImageNet $+0.33$; text: pooled $-0.14$; CUB $+0.63$, NABirds $-0.73$, ImageNet $+0.35$; joint pooled $-0.16$). Post-peak movement is moderately and positively associated with subsequent P3 loss (image $r=0.43$, $\rho=0.51$; text $r=0.54$, $\rho=0.64$; joint $r=0.50$, $\rho=0.61$; $n=78$; Table~\ref{tab:G_move_tower}). Image and text movement are strongly coupled (post-peak $r=0.85$, $\rho=0.89$; within-backbone standardized $r=0.95$; pre-peak $r=0.95$), and their apparent relative importance depends on normalization (Table~\ref{tab:G_move_joint}): pooled and domain-standardized regressions favor text, while within-backbone standardization gives near-equal image and text associations with variance-inflation factors up to about 11. Broad encoder-specific attribution is therefore not identifiable, and image-side localization is supported only by the representative encoder-state intervention below. Robustness: the text transformer plus projection and final LayerNorm gives $r=0.54$ and the pretrained token-embedding rows alone $r=0.41$.

An earlier radial quantity (distance-to-PT at deployment minus distance-to-PT at the peak, image encoder) correlates with the direct image post-peak quantity at $r=0.95$ over the 107 peak-before-deployment cells, with the direct value a median 1.11$\times$ larger, and gave a somewhat stronger held-out association with the post-peak decrease ($r\approx0.56$ versus $0.43$); the direct displacement is used throughout. Normalizing by the peak norm instead of the pretrained norm gives the same correlation to three decimals, because checkpoint changes are below 1\% of the parameter norm. The ImageNet-trained image-encoder association is near zero when pooled partly because of between-backbone scale structure (within individual ImageNet backbones $r$ is about $+0.26$ to $+0.49$). All statistics are descriptive associations across heterogeneous conditions, not causal or replicate-based estimates.
\input{supplement/tables/audit/G_move_order.tex}
\input{supplement/tables/audit/G_move_tower.tex}
\input{supplement/tables/audit/G_move_joint.tex}
\input{supplement/tables/audit/G_move_corr.tex}

\subsection{Held-out P1 over the same P3-peak$\to$deployment interval}
\label{app:G_p1move}
\begin{figure}[!htbp]
\centering
\includegraphics[width=0.62\linewidth]{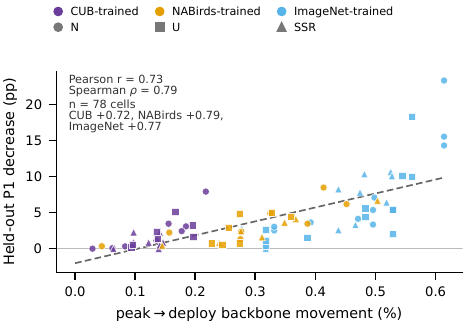}
\caption{Backbone movement versus held-out P1 decrease, N/U/SSR grid, same 78 held-out cells and same joint image--text backbone movement as the right panel of Figure~\ref{fig:mechanism}a. P1 is evaluated at each trajectory's P3-peak checkpoint and at its deployment checkpoint; no separate P1-optimal checkpoint is selected, and P1 decrease $=$ P1 (P3-peak checkpoint) $-$ P1 (deployment). This is a robustness analysis; held-out P1 is not treated as a name-free semantic-capability measure.}
\label{fig:S_P1_movement}
\end{figure}

As a robustness check on Figure~\ref{fig:mechanism}a, we asked whether the same joint backbone movement also tracks held-out P1 over the identical interval. For each of the 78 held-out trajectory cells with $t_{P3}<t_{\mathrm{deploy}}$, P1 is evaluated at that cell's P3-peak checkpoint and at its P1-selected deployment checkpoint, and \emph{P1 decrease} is the P1 value at the P3-peak checkpoint minus the value at deployment (positive = held-out P1 got worse); no separate P1-optimal checkpoint is selected. The joint movement is the same quantity as in Figure~\ref{fig:mechanism}a. Movement is more strongly associated with the P1 decrease than with the P3 decrease over the same cells (Figure~\ref{fig:S_P1_movement}: Pearson $r=0.73$, Spearman $\rho=0.79$, versus $r=0.50$, $\rho=0.61$ for P3; CUB-trained $+0.72$, NABirds-trained $+0.79$, ImageNet-trained $+0.77$; N $+0.76$, U $+0.65$, SSR $+0.82$), and the P1 and P3 decreases themselves correlate at $r=0.54$ across cells. Controlling for the peak-to-deployment time gap leaves the P1 association strong within each training domain (partial $r$ 0.74--0.83) and weaker for P3 (0.33--0.62). Held-out P1 is class-name retrieval on held-out species and is not treated as a name-free semantic-capability measure; the association is descriptive and does not imply a shared mechanism for the P1 and P3 changes.

\subsection{Wrong-class dependence of the P3 margin decrease}
\label{app:G_hardness}
The Figure~\ref{fig:mechanism}b analysis follows each image's peak-selected hardest wrong class. The correct anchor is the ground-truth class, and means are class-macro over the 348 classes. Because the competitor is each image's hardest wrong class at the peak and P3 R@1 is low, its mean similarity can exceed that of the ground-truth anchor; the relevant quantity is the paired peak-to-deployment margin change. Mean cosine similarity rises for both the correct attribute-profile anchor and the peak-selected wrong class, but rises more for the competitor. To clarify its scope, we repeated it for wrong classes at predetermined peak similarity ranks (Table~\ref{tab:G_hardness}; hardest = rank 1 of the 347 wrong classes, hard = rank 35, medium = rank 174, easy = rank 312; class identity fixed at the peak and reevaluated at deployment; class-macro, 10,000 class-bootstrap resamples). On P3 the correct-minus-wrong margin decrease is $0.00356$ [$0.00194$, $0.00518$] for the hardest, $0.00393$ [$0.00227$, $0.00553$] for the hard, $0.00760$ [$0.00569$, $0.00941$] for the median, and $0.01238$ [$0.01027$, $0.01439$] for the easy wrong class; across 15 rank levels the Spearman correlation between rank and encroachment is $+0.94$ (class-bootstrap 95\% [0.89, 0.99]; descriptive, because ranks are nested within images). A split-half selection check (ranking from half of each class's images, evaluating on the other half) gives $+0.0034$, $+0.0047$, $+0.0072$ and $+0.0116$, so the hardest-competitor effect remains positive ($+0.0034$, 95\% CI about $[+0.0014,+0.0053]$). The decrease in correct-minus-wrong P3 margin is therefore not concentrated among the wrong classes that are hardest at the peak; it extends across the wrong-class similarity spectrum and is larger for initially easier wrong classes. At the class level the hardest-competitor margin change is heterogeneous (mean $-0.00356$, 95\% CI $[-0.00518,-0.00194]$; median $-0.0034$; IQR $[-0.0141,+0.0073]$; 10th--90th percentile $[-0.0245,+0.0170]$; 59.5\% of classes decrease and 40.5\% increase). P1 does not reproduce the monotonic P3 pattern: its encroachment is $+0.0022$ [$+0.0006$, $+0.0038$] at rank 1, $-0.0033$ [$-0.0054$, $-0.0010$] at rank 35, $+0.0029$ [$+0.0007$, $+0.0051$] at rank 174 and $+0.0083$ [$+0.0060$, $+0.0107$] at rank 312 (non-monotone; the negative rank-35 value persists under split-half selection), so P1 and P3 do not share a single competitor-rank dependence.
\input{supplement/tables/audit/G_hardness.tex}

\subsection{Encoder-state swap}
\label{app:G_tower}
Table~\ref{tab:G_tower} gives the full swap results for the representative run. Held-out NAB-348 P3 is 17.63 at peak and 14.50 at deployment; combining the deployment image encoder with the peak text/routing encoder gives 15.15, and the reverse gives 17.09.
\input{supplement/tables/audit/G_tower.tex}
Table~\ref{tab:G_towerboot} gives the paired class-bootstrap decomposition (NAB-348 P3: image-late $-2.53$\,pp [$-3.53$, $-1.56$], text/SSR-late $-0.66$\,pp [$-1.19$, $-0.15$], difference $-1.88$\,pp [$-2.84$, $-0.90$], interaction approximately zero). This localization is specific to this representative transition.
\input{supplement/tables/audit/G_towerboot.tex}

\subsection{Additional encoder-state swaps on other SSR+ND runs}
\label{app:G_swapext}
To test whether the representative localization above recurs, we ran the same evaluation-only four-state encoder swap (same encoder-state substitution procedure and encoder partition: image backbone; text backbone plus SSR modules, with the logit scale and buffers taken from the deployment base checkpoint) on three additional SSR+ND runs, with no new training. The selection rule was fixed before the encoder-swap analysis. Eligible runs were identified from the NAB-trained SSR+ND runs audited on the dataset-class-disjoint CUB-59 subset (all 18 saved checkpoints; own CUB-59 P3 peak, epochs 4/8/2/4 for DFN5B H/14-378, DFN2B L/14, MetaCLIP H/14, MetaCLIP L/14; all four had a peak before the P1-selected deployment checkpoint and a decrease above 0.5\,pp), and candidates were then taken in a fixed order: MetaCLIP H/14 CUB$\to$NAB-348 (peak 12, deployment 75; only a 0.64\,pp decline, retained because it is the only different-backbone SSR+ND case of the same strict transfer), DFN2B L/14 NABirds$\to$CUB-59 (8, 45) and MetaCLIP L/14 NABirds$\to$CUB-59 (4, 40). DFN5B NABirds$\to$CUB-59 (which had the largest decline, 13.76\,pp) was not selected because it was not among the candidates under this rule; runs were selected by the fixed rule, not by decline magnitude or any swap result. Results are in Table~\ref{tab:G_swapext}.

The outcome is heterogeneous and largely unresolved. For MetaCLIP H/14 (CUB$\to$NAB-348), the image-late effect on held-out P3 is negative ($-0.73$\,pp, CI including zero) and the text/SSR-late effect is near zero, the same direction as the representative run but not resolved; on held-out P1 the image-late effect ($-2.44$\,pp) excludes zero and the image$-$text contrast excludes zero. For both NABirds$\to$CUB-59 runs, held-out P3 losses are split about equally between the encoders (image-late $-0.56$ and $-0.69$\,pp, text/SSR-late $-0.63$ and $-0.73$\,pp; image$-$text $+0.07$ and $+0.04$\,pp) with intervals that span zero over only 59 evaluation classes, so no encoder-specific localization is supported there. Target-domain P1 and P3 are not stable across the cross states in these runs, because their P3 peaks are early relative to deployment and target performance is still changing. We therefore keep the encoder localization explicitly representative of the single DFN5B/CUB$\to$NAB-348 run and do not generalize it. These are independent existing runs, not stochastic replicates, and the intervals reflect evaluation-class uncertainty only.
\input{supplement/tables/audit/G_swapext.tex}

\subsection{Cross-state P1/P3 margin recombination}
\label{app:G_cross}
Table~\ref{tab:G_crossstate} gives raw correct similarity, hardest-competitor similarity, and margin for the four states; these results are plotted in Figure~\ref{fig:S_mech_old}b. They agree with the encoder swap: substituting only the late image state reproduces nearly all of the held-out margin loss for P3 and P1.
\input{supplement/tables/audit/G_crossstate.tex}

\subsection{Correct-anchor versus hardest-competitor similarity}
\label{app:G_anchor}
\begin{figure}[!htbp]
\centering
\includegraphics[width=0.95\linewidth]{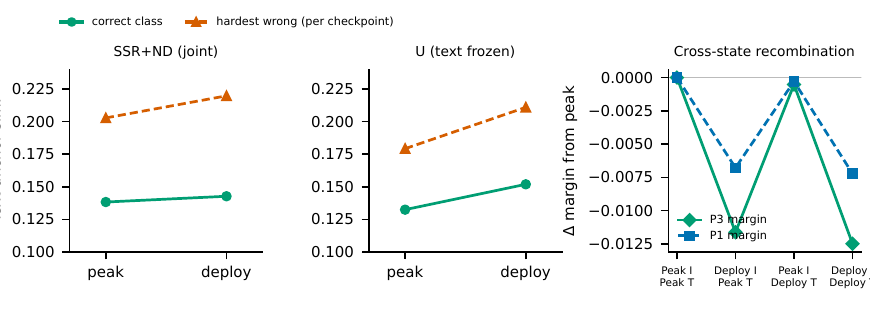}
\caption{Supplementary late-specialization diagnostics. \textbf{(a)} Correct-class versus hardest-wrong text-anchor similarity, peak to deployment, where the hardest wrong class is re-selected at each checkpoint (see Table~\ref{tab:G_enc} for why this statistic can overstate encroachment). \textbf{(b)} Change in NAB-348 correct-minus-competitor margin from peak under cross-state recombination (I = image encoder, T = text/routing encoder).}
\label{fig:S_mech_old}
\end{figure}

Table~\ref{tab:G_anchor} (plotted in Figure~\ref{fig:S_mech_old}a) reports the per-checkpoint hardest-wrong version: from peak to deployment the correct-class similarity is roughly flat to slightly rising while the hardest competitor's similarity rises more, for both SSR+ND and the text-frozen U run.
\input{supplement/tables/audit/G_anchor.tex}

\subsection{Fixed-competitor encroachment}
\label{app:G_enc}
Table~\ref{tab:G_enc} gives the peak$\to$deployment competitor-encroachment analysis summarized in \S\ref{sec:mechanism} and plotted per class in Figure~\ref{fig:mechanism}b, with per-checkpoint (dynamic) and fixed-at-peak competitors, on CUB-200 and NAB-348 and for P1 and P3. For the SSR+ND run on NAB-348 P3, the encroachment decomposes into a peak-to-deployment cosine-similarity rise of $+0.00443$ (95\% class-bootstrap CI $[+0.00227,+0.00665]$) for the correct attribute-profile class and $+0.00799$ ($[+0.00629,+0.00973]$) for the same peak-selected wrong class, a paired difference of $+0.00356$ ($[+0.00194,+0.00518]$); both similarities rise, the competitor's more, so the correct-minus-wrong margin decreases by $0.00356$ (Figure~\ref{fig:mechanism}b). On NAB-348 P3 the fixed-competitor encroachment is resolved and positive for SSR+ND ($+3.6\times10^{-3}$, 59\% of classes positive) and consistent with zero for the independent text-frozen U control ($-1.3\times10^{-3}$, 47\%); the dynamic hardest-wrong statistic is positive for both, illustrating why it is not used as the primary metric. On NAB-348 P1 the fixed-competitor effect is positive for both models, and on the CUB target it is negative for both (the correct anchor gains most); P1 and P3 are distinct capabilities and are not pooled. Table~\ref{tab:G_enclink} reports the class-level relationship between encroachment and the change in class top-1 accuracy, which is negative in every tested model$\times$dataset$\times$probe cell but partly mechanical. A separate semantic-neighborhood analysis of the same saved embeddings (pretrained text-text similarity freezes neighbor membership; class bootstrap) found that acquisition and late specialization have opposite-signed relationships to pretrained semantic distance, and that held-out errors remain concentrated among pretrained semantic neighbors (rank-1 neighbor: 36.2\% of late-phase flips for U with text frozen on NAB-348 P1 against a 0.29\% chance baseline); P1 and P3 do not show identical held-out late-phase neighbor dynamics, and no single neighbor mechanism is claimed.
\input{supplement/tables/audit/G_enc.tex}
\input{supplement/tables/audit/G_enclink.tex}

\subsection{Text-frozen control}
\label{app:G_tf}
In the text-frozen U run the entire text encoder (including layer norms) is locked while the image encoder and logit scale are trained. The text anchors were verified to be bit-identical across PT, peak, and deployment, so every change in Table~\ref{tab:G_anchor} for this model is image-side by construction. This run has peak epoch 20 and deployment epoch 75; it is a separate run from the main U run (deployment epoch 25).

\subsection{Image-only oracle prototype diagnostic}
\label{app:G_loo}
Table~\ref{tab:G_loo} reports leave-one-out image-prototype accuracy (an oracle using held-out-image class prototypes, not a zero-shot capability): about 88--91\% across PT, peak, and deployment. Class structure is largely preserved, though NAB-348 declines by 2.7--3.1\,pp from peak to deployment.
\input{supplement/tables/audit/G_loo.tex}

\subsection{Geometry hypotheses and null results}
\label{app:G_geom}
Table~\ref{tab:G_geom} reports all coarse-geometry correlations, weak/non-null and null alike. Tangent-residual alignment relates weakly but measurably to $\Delta$P3 (all four intervals exclude 0, $\rho\in[0.15,0.33]$). Residual-direction coherence, manifold radius, and effective rank show no reliable relationship with $\Delta$P3; some show non-null relationships with $\Delta$P1, so the null statement applies to $\Delta$P3 only. No coarse geometry metric provides a complete explanation.
\input{supplement/tables/audit/G_geom.tex}

\FloatBarrier

\section{Additional Controls and Alternative Interventions}
\label{app:controls}

\subsection{WiSE interpolation}
\label{app:H_wise}

Weights are interpolated as $\theta(\alpha)=(1-\alpha)\theta_{\mathrm{PT}}+\alpha\theta_{\mathrm{FT}}$ at evaluation time for N (deployment epoch 40) and U (deployment epoch 25), DFN5B H/14-378 fine-tuned on CUB-200 (Table~\ref{tab:H_wise}). At $\alpha=0.5$, held-out NAB-348 P3 recovers toward pretrained while target CUB P1/P3 remain close to or above the deployed values (N: CUB P1 91.5 vs.\ 91.1 deployed, CUB P3 36.0 vs.\ 31.1, NAB P3 15.1 vs.\ 13.7, NAB P1 82.3 vs.\ 71.6; U: CUB P1 91.5 vs.\ 91.2, CUB P3 41.7 vs.\ 35.8, NAB P3 17.0 vs.\ 15.1, NAB P1 82.0 vs.\ 76.0). Moving weights toward the pretrained model can therefore recover substantial held-out capability while preserving strong target performance in this representative setting, so specialization need not be irreversible. For context, WiSE-interpolated U at $\alpha=0.5$ is comparable to or better than the deployed SSR run on these two P3 metrics (SSR: CUB 40.4, NAB 16.9), whereas SSR+ND has higher target P3 (46.3) and lower NAB P3 (14.5). WiSE is a probe of reversibility, not a competing method. SSR and SSR+ND are not evaluated because their special tokens and gates have no direct pretrained counterparts. Single seed, coarse grid, one backbone/domain.
\input{supplement/tables/audit/H_wise.tex}

\subsection{Hard-negative objective}
\label{app:H_hn}

\paragraph{Motivation.} SSR improves name-free semantic capability without explicitly training the model to separate a correct attribute profile from nearby semantic alternatives. This control asks whether directly strengthening such target-side discrimination is enough to improve transferable semantic accessibility, or whether it can prevent the later loss of transferable P3. Starting from the canonical SSR recipe, we add a controlled hard-negative ranking term in which the negative text differs from the correct profile by a single attribute value. This is a diagnostic intervention, not a proposed additional method.

\paragraph{Construction and objective.} For a CUB training image with its correct structured profile (name, concept and original attributes; the \emph{correct} or \emph{positive} profile), the controlled semantic hard negative keeps the same name, concept and remaining attributes and changes exactly one attribute phrase by a minimal edit (a single colour, shape or size change). The negative pool contains 618 valid candidates covering all 200 CUB classes. During training one negative is sampled per class at each effective optimizer step, shared by all images of that class in the step and resampled at the next step. Let $s_{\mathrm{pos}}$ and $s_{\mathrm{neg}}$ be the (logit-scaled) image--text similarities of the image to the final fused text representation of its positive profile and of its hard negative. The auxiliary loss is
\begin{equation}
L_{\mathrm{HN}}=\mathrm{mean}\,\mathrm{softplus}(s_{\mathrm{neg}}-s_{\mathrm{pos}}),\qquad L=L_{\mathrm{CLIP}}+\lambda L_{\mathrm{HN}},\quad \lambda=0.5,
\end{equation}
where $L_{\mathrm{CLIP}}$ is the standard multi-positive contrastive loss of the SSR recipe. The penalty is small when $s_{\mathrm{pos}}>s_{\mathrm{neg}}$ and grows as the negative approaches or exceeds the positive, so the term directly encourages positive-versus-negative separation. The run is DFN5B H/14-378, fine-tuned on CUB-200 only, one seed, one $\lambda$; the matched baseline is the canonical SSR recipe run from the identical starting checkpoint with the same epoch schedule. The target is CUB-200 and the strict held-out population is NAB-348.

\begin{figure}[!tb]
\centering
\includegraphics[width=\linewidth]{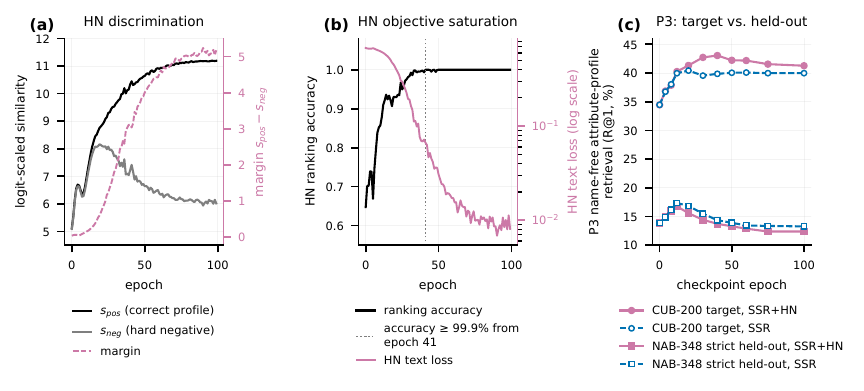}
\caption{\textbf{Hard-negative semantic-discrimination control,} DFN5B H/14-378 fine-tuned on CUB-200 (single seed, $\lambda=0.5$). Starting from SSR, a target-side ranking loss separates the correct attribute profile from a controlled hard negative differing in one attribute value. \textbf{(a)} Logit-scaled similarities of the image to the positive ($s_{\mathrm{pos}}$) and hard-negative ($s_{\mathrm{neg}}$) profiles and their margin, per training epoch (training-log quantities on the hard-negative batches, not an evaluation probe): $s_{\mathrm{neg}}$ first rises (until about epoch 19) and then declines while $s_{\mathrm{pos}}$ keeps increasing, so the relative margin grows. \textbf{(b)} Hard-negative ranking accuracy (fraction with $s_{\mathrm{pos}}>s_{\mathrm{neg}}$) first reaches 99.9\% at epoch 41 (dotted line) and stays there, while the hard-negative text loss keeps decreasing. \textbf{(c)} P3 (name-free attribute-profile retrieval) on CUB-200 (target) and strict held-out NAB-348 at the 11 evaluated checkpoints (no interpolation): the hard-negative run exceeds the matched SSR baseline on target CUB P3 at later checkpoints but not on held-out NAB-348 P3 at matched epochs. This representative control shows that stronger target-side semantic discrimination need not translate into stronger transferable P3.}
\label{fig:S_hardneg}
\end{figure}

\paragraph{Matched-epoch result.} Target CUB P3 is improved at later checkpoints and reaches a higher peak (43.04 at epoch 40 versus the SSR peak of 40.42 at epoch 20), and target image--P3 alignment is higher (0.355 versus 0.315 at epoch 100). Strict held-out NAB-348 P3 does not show the same improvement at matched evaluated epochs: from epoch 8 onward it is below the SSR baseline, and by epoch 100 it is 12.38 versus 13.30 despite the stronger target-side discrimination (held-out alignment is very similar, 0.231 versus 0.228). Held-out P2 is consistently lower for the hard-negative run, and P1 is slightly lower at matched epochs ($-0.1$ to $-1.6$\,pp). Exact values at five of the evaluated epochs are in Table~\ref{tab:H_hardneg}. (The table's SSR baseline comes from the alignment evaluation and differs from the trajectory evaluation plotted in Figure~\ref{fig:S_hardneg}c by at most about 0.05\,pp, e.g.\ NAB-348 P3 at epoch 100 is 13.30 versus 13.26; signs are unchanged.)

\paragraph{Checkpoint caveat.} The conclusion depends on the comparison rule. At matched training epochs, stronger target discrimination does not produce a corresponding held-out P3 gain. At each run's own P1-selected deployment checkpoint, however, the hard-negative run is higher on NAB-348 P3 (15.57 versus 13.34) because its deployment occurs much earlier (epoch 20 versus epoch 75 for the matched SSR baseline in this evaluation series). We therefore use this experiment only to show that stronger target-side discrimination is not by itself sufficient to guarantee improved transfer at a matched stage of training.
\input{supplement/tables/audit/H_hardneg.tex}

\paragraph{Interpretation and scope.} The control reinforces the capability-trajectory view: optimization can continue to strengthen semantic discrimination on the training ontology while held-out name-free retrieval has already passed its optimum (target CUB P3 stays strong or improves later, whereas NAB-348 P3 peaks early and declines). It is separate from the fixed-competitor analysis in \S\ref{sec:mechanism} and is not a mechanistic explanation of it: the hard-negative objective acts on target CUB semantic alternatives, whereas Figure~\ref{fig:mechanism}b analyzes the geometry of strict held-out NAB-348 P3 after its peak. The evidence is one backbone, one training domain, one seed and one $\lambda$ in a diagnostic control, not a method sweep; it does not show that hard negatives fail, hurt transfer, or cause late specialization, and the safe conclusion is only that stronger target-side discrimination need not improve transferable P3 at a matched training stage.

\subsection{Direct name-access ablation: interpretation}
\label{app:H_trackA}

See \S\ref{app:ssr_name_access} and \S\ref{app:trackA}. The name-access effect of the direct name-access ablation and the late specialization analyzed in \S\ref{app:mech} are mechanistically distinct and should not be conflated.

\subsection{Other verified controls}
\label{app:H_other}

No further controls beyond those in this appendix are reported.

\FloatBarrier

\section{Qualitative Probe Examples}
\label{app:qualitative}

The examples in Table~\ref{tab:I_examples} illustrate the probes; they are not performance claims and were not selected by model success or failure. Dataset photographs are not reproduced because the underlying images are third-party works. Reviewed CAPP profiles are available for CUB and NABirds, so no CAPP example is given for FineView or ImageNet.

\begin{table}[t]
\centering
\scriptsize
\setlength{\tabcolsep}{3pt}\begin{tabular}{p{1.8cm}p{5.0cm}p{5.0cm}}
\toprule
Item & CUB: Kentucky Warbler & NAB-348: adult male Lesser Goldfinch \\
\midrule
P1 & a photo of a Kentucky Warbler & a photo of a adult male Lesser Goldfinch \\
P3 template & a bird with \{attributes\} & a bird with \{attributes\} \\
CAPP correct profile & a bird with black crown, bright yellow underparts, olive-green upperparts & a bird with bright yellow belly, black cap, glossy cap, pointed wing, white wing patch \\
CAPP one-slot counterfactual & a bird with black crown, \textbf{gray underparts}, olive-green upperparts (slot: underparts color) & a bird with bright yellow belly, black cap, glossy cap, \textbf{rounded wing}, white wing patch (slot: wing shape) \\
\midrule
Item & FineView: Monarch & ImageNet: Goldfish \\
\midrule
P3 template & a butterfly with \{attributes\} & a object with \{attributes\} \\
Attributes & orange upperside; wide black borders; black veins; blurred black veins; white border spots; white wing-tip spots & orange-red body; bright yellow body; torpedo-shaped body; large reflective scales; bifurcated caudal fin; stiff dorsal fin; smooth operculum \\
CAPP & not available & not available \\
\bottomrule
\end{tabular}
\caption{Verified probe examples. The CAPP pairs are drawn from the reviewed full-class CAPP profiles (donor values for the swapped slot come from other classes). Only the changed slot is highlighted. The Figure~\ref{fig:method}c CAPP panel is an illustrative profile, not one of these verified pairs.}
\label{tab:I_examples}
\end{table}

\FloatBarrier

\section{Reproducibility and Result Provenance}
\label{app:provenance}

\subsection{Full statement of limitations}
\label{app:limitations}

The main-paper Limitations paragraph is a summary of the following full statement.

\paragraph{Scope and limitations.} The fixed-competitor and image-only analyses in \S\ref{sec:mechanism} are single-run diagnostics, and the broader movement result is a descriptive association over four backbones and N/U/SSR (SSR+ND not included), not a causal mechanism. Movement also tracks held-out P1 decrease more strongly over the same P3-peak$\to$deployment intervals (Appendix~\ref{app:G_p1move}), so it should be interpreted as accompanying broader held-out specialization rather than as a P3-specific mechanism; heterogeneous encoder-state swaps (Appendix~\ref{app:G_swapext}) likewise support no encoder-specific attribution. The six backbones are diverse rather than statistically representative or stochastic replicates, so six-backbone summaries are descriptive and carry no inferential intervals. Deployment is selected by target-domain P1 only; a different rule would select different trajectory points. P2 is not attribute-isolating, and neither P3 nor CAPP establishes spatial grounding, localization, or part binding. For CUB/NABirds and FineView, source descriptions are LLM-processed from public species resources, whereas ImageNet descriptions and attributes are LLM-generated (Appendix~\ref{app:attribute_sources}); residual source, template, or generation effects cannot be fully excluded.

\paragraph{Expanded interpretive controls.} Three controls constrain this interpretation (full numbers in Appendix~\ref{app:controls}). In the representative DFN5B/CUB setting, WiSE interpolation ($\alpha{=}0.5$) partially recovers held-out P3 for N and U while retaining most target P1/P3, so specialization need not be irreversible; this is not a general WiSE result, and SSR/SSR+ND are not tested because their structured components lack direct pretrained counterparts. The direct name-access ablation (\S\ref{sec:capability_differences}) shows that restricting direct name access changes P3/CAPP and attribute-route geometry while ordinary P2 remains name-dominated, a routing effect distinct from the late-specialization analyses in \S\ref{sec:mechanism}. A target-side hard-negative objective improves target P3 without a corresponding held-out advantage at matched later checkpoints, showing that stronger in-domain semantic discrimination need not imply stronger transferable accessibility at the same stage of training (Appendix~\ref{app:H_hn}).

\subsection{Main-item to supplement map}
\label{app:J_map}

\begin{table}[t]
\centering
\scriptsize
\setlength{\tabcolsep}{3pt}\begin{tabular}{p{2.0cm}p{3.3cm}p{3.0cm}p{2.2cm}p{1.8cm}}
\toprule
Main item & Supplement table(s) & Population & Checkpoint rule & Evidence type \\
\midrule
Fig.~\ref{fig:acquisition}b & Tables~\ref{tab:S1_full}--\ref{tab:S4_summary} & ImageNet target, same-ontology shifts, cross-ontology; six backbones & trajectory peak & six-backbone descriptive \\
Fig.~\ref{fig:acquisition}a & Table~\ref{tab:S3_dense} & CUB target, NAB-348 held-out; DFN5B & peak, own-P1 deploy & single run per method \\
Table~\ref{tab:deploy_profile} & Table~\ref{tab:S8_deployment} & ImageNet val; six backbones & own-P1 deploy & six-backbone descriptive \\
Fig.~\ref{fig:supervision}a & Table~\ref{tab:S9_name_dependency} & CUB target, NAB-348; DFN5B & own-P1 deploy & single run, 1000-replicate class bootstrap \\
Fig.~\ref{fig:supervision}b & Table~\ref{tab:S10_s_vs_snd_full} & 3 training domains $\times$ six backbones (18 pairs) & trajectory peak & six-backbone descriptive \\
Table~\ref{tab:capp_target_summary} & Tables~\ref{tab:S11_target_pairwise}--\ref{tab:S12_heldout_capp} & CUB/NAB target; six backbones & own-P1 deploy & six-backbone descriptive \\
Fig.~\ref{fig:specialization}a & Table~\ref{tab:S3_dense} & DFN5B CUB-trained, CUB-200 and NAB-348 & peak, deploy & single run per method \\
Fig.~\ref{fig:specialization}b & Tables~\ref{tab:F_prev}, \ref{tab:F_counts}, \ref{tab:F_openai}, \ref{tab:S1_full}--\ref{tab:S2_full} & six backbones; target and cross-ontology, by method & categorical peak vs.\ deploy & six-backbone descriptive \\
Fig.~\ref{fig:specialization}c & Table~\ref{tab:F_mag}, \ref{tab:S2_full} & six backbones; cross-ontology cells with $t_{P3}<t_{\mathrm{deploy}}$ & peak $-$ deploy P3 (pp) & six-backbone descriptive \\
Fig.~\ref{fig:S_capp_timing} & Table~\ref{tab:S_capp_timing} & DFN5B CUB-trained, CUB-200 and NAB-348 & P3 / CAPP peak, deploy & single run per method \\
Fig.~\ref{fig:mechanism}a & Tables~\ref{tab:G_move_order}--\ref{tab:G_move_corr} & N/U/SSR grid; joint image+text backbone movement (encoder-separated audit in Tables~\ref{tab:G_move_tower}--\ref{tab:G_move_joint}) & own P3 peak / P1 deploy & descriptive association \\
\S\ref{app:G_tower} (no main panel) & Tables~\ref{tab:G_tower}, \ref{tab:G_towerboot} & DFN5B SSR+ND, NAB-348 & epoch 20 / 60 & single run \\
\S\ref{app:G_swapext} (no panel) & Table~\ref{tab:G_swapext} & three additional SSR+ND runs (MetaCLIP H/14, DFN2B L/14, MetaCLIP L/14) & own P3 peak / deploy & independent runs, class bootstrap \\
Fig.~\ref{fig:S_P1_movement} & \S\ref{app:G_p1move} & same 78 held-out N/U/SSR cells as Fig.~\ref{fig:mechanism}a, held-out P1 at P3-peak vs deploy & same intervals & descriptive association \\
\S\ref{app:G_hardness} (no panel) & Table~\ref{tab:G_hardness} & SSR+ND DFN5B CUB$\to$NAB-348, P3 and P1, peak-ranked wrong classes & peak / deploy & single run, class bootstrap \\
Fig.~\ref{fig:mechanism}b & Tables~\ref{tab:G_enc}--\ref{tab:G_enclink} & SSR+ND (per-class values); text-frozen U control in the table, NAB-348 P3 & peak / deploy & single runs, class bootstrap \\
Fig.~\ref{fig:S_mech_old}a & Table~\ref{tab:G_anchor} & SSR+ND; text-frozen U & PT / peak / deploy & single runs \\
Fig.~\ref{fig:S_mech_old}b & Table~\ref{tab:G_crossstate} & SSR+ND, NAB-348 & four encoder states & single run \\
Discussion: WiSE & Table~\ref{tab:H_wise} & DFN5B CUB-trained, N and U & interpolation $\alpha$ & single seed \\
Discussion: hard negative & Fig.~\ref{fig:S_hardneg}, Table~\ref{tab:H_hardneg} (\S\ref{app:H_hn}) & DFN5B CUB-trained; CUB target, NAB-348 strict & matched epochs (own-deployment comparison in \S\ref{app:H_hn}) & single seed, single $\lambda$ \\
Discussion: name access & Tables~\ref{tab:E_trackA_p3}--\ref{tab:E_trackA_rsa}, Fig.~\ref{fig:S_trackA_traj} & DFN5B CUB-trained; CUB target, NAB-348 strict & own-P1 deploy (P3 and CAPP trajectories) & single run per variant \\
\S\ref{app:F_atlas} (P1/P3 atlas) & Figs.~\ref{fig:S_atlas_in_core}--\ref{fig:S_atlas_fg_openai} & ImageNet-trained (DFN/MetaCLIP, OpenAI); CUB-/NABirds-trained (DFN/MetaCLIP, OpenAI); target, full and strict held-out kept separate & own-P1 deploy marker; no separate P1 optimum & single run per method and backbone \\
P3 vs.\ CAPP change & Fig.~\ref{fig:S_p3_capp_scatter}, Tables~\ref{tab:S11_target_pairwise}--\ref{tab:S12_heldout_capp} & DFN/MetaCLIP and OpenAI; target and strict held-out & own-P1 deploy & descriptive, no coefficient \\
\bottomrule
\end{tabular}
\caption{Main-paper items mapped to supplement tables, evaluation populations, checkpoint rules, and evidence type. Evidence types are distinct: single training runs, six heterogeneous backbones (descriptive), class-level bootstrap resampling of the evaluation set, and archived historical multi-seed checks (Table~\ref{tab:C_multiseed}); they must not be conflated.}
\label{tab:J_map}
\end{table}

\FloatBarrier
\subsection{Experiment provenance and run naming}
\label{app:J_naming}

SSR is the structured NAME/ATTRIBUTE representation with routed causal attention and Structured Text Fusion defined above (CUB/NAB include a shared CONCEPT segment; ImageNet does not), and SSR+ND is SSR with NAME-summary dropout ($p=0.5$) at training-time fusion. For DFN5B/CUB the canonical SSR+ND run differs from the SSR run only by the added name-drop and is distinct from an earlier historical name-drop calibration run. The historical run is reported only in the supplemental probability-calibration analysis (\S\ref{app:C_pcal}) and is not used in the canonical main trajectory results. Historical multi-seed runs used earlier SSR routing variants that are not used for the main results and are not the canonical SSR.

\subsection{Limitations of replication and seed coverage}
\label{app:J_limits}

All main trajectory results come from one training run per backbone, domain, and method (seed 0); variability across the six backbones is descriptive heterogeneity, not a confidence interval. The mechanism, WiSE, hard-negative, name-access, and text-frozen analyses are single representative runs. Multi-seed evidence exists only for historical DFN5B configurations and only for in-domain retrieval. Bootstrap intervals quantify evaluation-set sampling uncertainty only. Unresolved provenance items: the CAPP evaluator source that averages cells into the class-slot-macro score; the NAB-348 derivation script; OpenAI ImageNet parameter files for steps 1--5; GPU type for the untyped allocations and exact per-run software versions.

%% file: supplement/tables/audit/C_recipe_imagenet.tex
\begin{table}[!htbp]
\centering
\small\setlength{\tabcolsep}{3.5pt}
\begin{tabular}{lrrrrl}
\toprule
Method & LR & LR$_{\mathrm{sl}}$ & WD & Eff.\ batch & Routing \\
\midrule
N & 5e-06 & 5e-06 & 0.1 & 2000 & -- \\
U & 5e-06 & 5e-06 & 0.1 & 2000 & -- \\
SSR & 5e-06 & 5e-06 & 0.1 & 2000 & no-concept \\
SSR+ND & 5e-06 & 5e-06 & 0.1 & 2000 & no-concept, $p{=}0.5$ \\
\bottomrule
\end{tabular}
\caption{ImageNet-1k training settings used across the six backbones; available run parameter records agree with these settings, with early OpenAI records unavailable as noted below. Training is a chain of 11 resumed steps, each one epoch over the incremental data slice up to cumulative fractions 0.25, 0.5, 1, 2, 5, 10, 20, 40, 60, 80, 100\% (4k to 1.3M images; 2 to 650 cumulative effective batches), with a constant learning rate (\texttt{--skip-scheduler}; the declared warmup of 10000 is therefore inert). Milestone indices are not epochs and not proportional to gradient updates. For OpenAI L/14 and B/32 N/U/SSR, run parameter files were located only for steps 6--11 and 7--11 respectively; these are consistent with the other backbones, and steps 1--5 are unresolved.}
\label{tab:C_recipe_imagenet}
\end{table}

%% file: supplement/tables/audit/C_recipe_cubnab.tex
\begin{center}
\footnotesize\setlength{\tabcolsep}{3.5pt}
\begin{longtable}{llrrrrrrl}
\caption{Training settings read from each canonical run's saved parameter record (CUB-200 and NAB-555 training, 100 epochs, seed 0, AdamW-style open\_clip trainer with multi-positive contrastive loss, bf16, no extra augmentation, cosine schedule). LR is the base learning rate; LR$_{\mathrm{sl}}$ is the learning rate of the structured-routing layers (for N/U it is the unused default). Effective batch is 20 $\times$ gradient-accumulation steps (100 for CUB, 111 for NAB). Settings are parent-specific and not normalized: the OpenAI L/14 and B/32 CUB runs use weight decay 0.2 and warmup 10000, whereas the DFN/MetaCLIP runs use 0.1 and 1000; DFN5B CUB SSR and SSR+ND use LR$_{\mathrm{sl}}=5\times10^{-6}$ while all other structured runs use $10^{-5}$. SSR+ND uses name-branch dropout $p=0.5$. Deploy ep.\ is the run's own P1-selected checkpoint.}\label{tab:C_recipe_cubnab}\\
\toprule
Backbone & Method & LR & LR$_{\mathrm{sl}}$ & WD & Eff.\ batch & Warmup & Deploy ep. & Routing \\
\midrule
\endfirsthead
\toprule
Backbone & Method & LR & LR$_{\mathrm{sl}}$ & WD & Eff.\ batch & Warmup & Deploy ep. & Routing \\
\midrule
\endhead
\midrule \multicolumn{9}{r}{\emph{continued on next page}} \\ \midrule
\endfoot
\bottomrule
\endlastfoot
\multicolumn{9}{l}{\textbf{CUB-200 training}} \\
DFN5B H/14-378 & N & 5e-06 & 1e-05 & 0.1 & 2000 & 1000 & 40 & -- \\
 & U & 5e-06 & 1e-05 & 0.1 & 2000 & 1000 & 25 & -- \\
 & SSR & 5e-06 & 5e-06 & 0.1 & 2000 & 1000 & 20 & shared-concept \\
 & SSR+ND & 5e-06 & 5e-06 & 0.1 & 2000 & 1000 & 60 & shared-concept \\
\midrule
DFN2B L/14 & N & 5e-06 & 1e-05 & 0.1 & 2000 & 1000 & 20 & -- \\
 & U & 5e-06 & 1e-05 & 0.1 & 2000 & 1000 & 20 & -- \\
 & SSR & 5e-06 & 1e-05 & 0.1 & 2000 & 1000 & 100 & shared-concept \\
 & SSR+ND & 5e-06 & 1e-05 & 0.1 & 2000 & 1000 & 40 & shared-concept \\
\midrule
MetaCLIP H/14 & N & 5e-06 & 1e-05 & 0.1 & 2000 & 1000 & 75 & -- \\
 & U & 5e-06 & 1e-05 & 0.1 & 2000 & 1000 & 75 & -- \\
 & SSR & 5e-06 & 1e-05 & 0.1 & 2000 & 1000 & 100 & shared-concept \\
 & SSR+ND & 5e-06 & 1e-05 & 0.1 & 2000 & 1000 & 75 & shared-concept \\
\midrule
MetaCLIP L/14 & N & 5e-06 & 1e-05 & 0.1 & 2000 & 1000 & 60 & -- \\
 & U & 5e-06 & 1e-05 & 0.1 & 2000 & 1000 & 60 & -- \\
 & SSR & 5e-06 & 1e-05 & 0.1 & 2000 & 1000 & 100 & shared-concept \\
 & SSR+ND & 5e-06 & 1e-05 & 0.1 & 2000 & 1000 & 100 & shared-concept \\
\midrule
OpenAI L/14 & N & 5e-06 & 1e-05 & 0.2 & 2000 & 10000 & 75 & -- \\
 & U & 5e-06 & 1e-05 & 0.2 & 2000 & 10000 & 100 & -- \\
 & SSR & 5e-06 & 1e-05 & 0.2 & 2000 & 10000 & 90 & shared-concept \\
 & SSR+ND & 5e-06 & 1e-05 & 0.2 & 2000 & 10000 & 90 & shared-concept \\
\midrule
OpenAI B/32 & N & 5e-06 & 1e-05 & 0.2 & 2000 & 10000 & 100 & -- \\
 & U & 5e-06 & 1e-05 & 0.2 & 2000 & 10000 & 60 & -- \\
 & SSR & 5e-06 & 1e-05 & 0.2 & 2000 & 10000 & 60 & shared-concept \\
 & SSR+ND & 5e-06 & 1e-05 & 0.2 & 2000 & 10000 & 100 & shared-concept \\
\midrule
\multicolumn{9}{l}{\textbf{NAB-555 training}} \\
DFN5B H/14-378 & N & 5e-06 & 1e-05 & 0.1 & 2220 & 1000 & 12 & -- \\
 & U & 5e-06 & 1e-05 & 0.1 & 2220 & 1000 & 15 & -- \\
 & SSR & 5e-06 & 1e-05 & 0.1 & 2220 & 1000 & 40 & shared-concept \\
 & SSR+ND & 5e-06 & 1e-05 & 0.1 & 2220 & 1000 & 75 & shared-concept \\
\midrule
DFN2B L/14 & N & 5e-06 & 1e-05 & 0.1 & 2220 & 1000 & 50 & -- \\
 & U & 5e-06 & 1e-05 & 0.1 & 2220 & 1000 & 35 & -- \\
 & SSR & 5e-06 & 1e-05 & 0.1 & 2220 & 1000 & 45 & shared-concept \\
 & SSR+ND & 5e-06 & 1e-05 & 0.1 & 2220 & 1000 & 45 & shared-concept \\
\midrule
MetaCLIP H/14 & N & 5e-06 & 1e-05 & 0.1 & 2220 & 1000 & 75 & -- \\
 & U & 5e-06 & 1e-05 & 0.1 & 2220 & 1000 & 20 & -- \\
 & SSR & 5e-06 & 1e-05 & 0.1 & 2220 & 1000 & 35 & shared-concept \\
 & SSR+ND & 5e-06 & 1e-05 & 0.1 & 2220 & 1000 & 35 & shared-concept \\
\midrule
MetaCLIP L/14 & N & 5e-06 & 1e-05 & 0.1 & 2220 & 1000 & 75 & -- \\
 & U & 5e-06 & 1e-05 & 0.1 & 2220 & 1000 & 30 & -- \\
 & SSR & 5e-06 & 1e-05 & 0.1 & 2220 & 1000 & 30 & shared-concept \\
 & SSR+ND & 5e-06 & 1e-05 & 0.1 & 2220 & 1000 & 40 & shared-concept \\
\midrule
OpenAI L/14 & N & 5e-06 & 1e-05 & 0.1 & 2220 & 1000 & 100 & -- \\
 & U & 5e-06 & 1e-05 & 0.1 & 2220 & 1000 & 30 & -- \\
 & SSR & 5e-06 & 1e-05 & 0.1 & 2220 & 1000 & 100 & shared-concept \\
 & SSR+ND & 5e-06 & 1e-05 & 0.1 & 2220 & 1000 & 45 & shared-concept \\
\midrule
OpenAI B/32 & N & 5e-06 & 1e-05 & 0.1 & 2220 & 1000 & 100 & -- \\
 & U & 5e-06 & 1e-05 & 0.1 & 2220 & 1000 & 60 & -- \\
 & SSR & 5e-06 & 1e-05 & 0.1 & 2220 & 1000 & 100 & shared-concept \\
 & SSR+ND & 5e-06 & 1e-05 & 0.1 & 2220 & 1000 & 75 & shared-concept \\
\midrule
\end{longtable}
\end{center}

%% file: supplement/tables/audit/C_multiseed.tex
\begin{table}[!htbp]
\centering
\small\setlength{\tabcolsep}{3.5pt}
\begin{tabular}{llrrrl}
\toprule
Dataset & Method & Seeds & i2t R@1 & t2i R@1 & Status \\
\midrule
CUB-200 & N & 5 & 91.98 $\pm$ 0.05 & 94.20 $\pm$ 0.06 & converged \\
 & U & 5 & 92.02 $\pm$ 0.10 & 94.03 $\pm$ 0.05 & converged \\
 & earlier SSR routing variant & 5 & 92.01 $\pm$ 0.10 & 94.30 $\pm$ 0.11 & converged \\
\midrule
NABirds & N & 5 & 89.00 $\pm$ 0.06 & 92.78 $\pm$ 0.04 & not conv. \\
 & U & 5 & 88.91 $\pm$ 0.07 & 92.86 $\pm$ 0.24 & not conv. \\
 & earlier SSR routing variant & 5 & 88.92 $\pm$ 0.09 & 93.14 $\pm$ 0.10 & not conv. \\
\midrule
FineView & N & 5 & 83.63 $\pm$ 0.17 & 89.42 $\pm$ 0.16 & converged \\
 & U & 5 & 83.70 $\pm$ 0.06 & 89.00 $\pm$ 0.12 & converged \\
 & earlier SSR routing variant & 5 & 83.28 $\pm$ 0.09 & 88.91 $\pm$ 0.20 & converged \\
\midrule
\bottomrule
\end{tabular}
\caption{Archived multi-seed runs (mean $\pm$ sample std; DFN5B H/14-378 only; seeds 10/20/30/40/50; in-domain validation R@1 at the best checkpoint; sample standard deviation). \textbf{None of these runs uses the current canonical SSR recipe} (current: shared-concept routing, LR $5\times10^{-6}$; archived structured runs: earlier routing variants, LR $10^{-5}$; the NABirds runs used a different earlier variant from the CUB-200 and FineView runs), so they are historical matched robustness checks, never pooled with the trajectory results, and no SSR+ND, ImageNet, or other-backbone multi-seed runs exist. No P1/P2/P3/CAPP was evaluated for them. NABirds runs were walltime-limited and are not converged (best checkpoints fall at epochs $\approx$19--23, with unequal numbers of logged epochs across seeds). On CUB, historical SSR used 30 accumulation steps versus 100 for N/U, so epoch-based comparisons are confounded.}
\label{tab:C_multiseed}
\end{table}

%% file: supplement/tables/audit/C_multiseed_perseed.tex
\begin{table}[!htbp]
\centering
\footnotesize\setlength{\tabcolsep}{3.5pt}
\begin{tabular}{llrrrrr}
\toprule
Dataset & Method & Seed 10 & Seed 20 & Seed 30 & Seed 40 & Seed 50 \\
\midrule
CUB-200 & N & 91.95/94.22 & 92.02/94.23 & 91.92/94.25 & 92.03/94.10 & 92.00/94.18 \\
 & U & 91.93/93.95 & 92.05/94.08 & 91.93/94.03 & 92.00/94.07 & 92.17/94.03 \\
 & hist.\ SSR & 92.08/94.30 & 91.98/94.18 & 91.93/94.20 & 92.15/94.38 & 91.92/94.43 \\
\midrule
NABirds & N & 89.07/92.76 & 88.94/92.74 & 89.02/92.79 & 89.02/92.84 & 88.93/92.79 \\
 & U & 88.92/92.66 & 89.02/92.74 & 88.88/92.66 & 88.86/93.18 & 88.87/93.04 \\
 & hist.\ SSR & 88.78/93.14 & 88.94/93.31 & 88.96/93.07 & 88.90/93.11 & 89.03/93.05 \\
\midrule
FineView & N & 83.81/89.58 & 83.65/89.39 & 83.39/89.17 & 83.56/89.42 & 83.77/89.52 \\
 & U & 83.69/89.15 & 83.60/88.92 & 83.75/88.86 & 83.71/88.98 & 83.75/89.10 \\
 & hist.\ SSR & 83.35/89.14 & 83.29/88.75 & 83.11/88.71 & 83.31/89.11 & 83.33/88.83 \\
\midrule
\bottomrule
\end{tabular}
\caption{Per-seed best-checkpoint i2t / t2i R@1 (\%) for the archived multi-seed runs summarized in Table~\ref{tab:C_multiseed}.}
\label{tab:C_multiseed_perseed}
\end{table}

%% file: supplement/tables/audit/C_pcal_peak.tex
\begin{table}[t]
\centering\footnotesize\setlength{\tabcolsep}{4pt}
\begin{tabular}{lrrrrrr}
\toprule
 & \multicolumn{3}{c}{CUB-trained} & \multicolumn{3}{c}{NABirds-trained} \\
\cmidrule(lr){2-4}\cmidrule(lr){5-7}
$p$ & CUB-200 (target) & NAB-348 & FineView & NAB-555 (target) & CUB-59 & FineView \\
\midrule
0 & 40.42 & 17.35 & 9.90 & 15.29 & 53.80 & 10.94 \\
0.25 & 44.60 & 17.34 & 9.48 & 16.11 & 54.17 & 11.13 \\
0.5 & 46.82 & 17.54 & 9.33 & 16.85 & 54.98 & 11.23 \\
0.75 & 48.95 & 17.87 & 9.48 & 17.41 & 55.79 & 11.15 \\
1 & 50.35 & 18.07 & 9.52 & 17.93 & 56.10 & 10.58 \\
\bottomrule
\end{tabular}
\caption{Historical NAME-drop probability calibration (DFN5B H/14-378; one run per $p$): maximum observed P3 (R@1, \%) over saved epochs $>0$ (the NABirds $p{=}0$ run was evaluated only through epoch 40, the others to epoch 100) on the training-domain target and on held-out sets. The $p{=}0.5$ CUB-trained run is the earlier weight-decay-0.2 run, not the canonical SSR+ND run. Held-out columns use each run's own P3 peak over epochs (a peak-anywhere convention, not the P1-selected checkpoint).}
\label{tab:C_pcal_peak}
\end{table}

%% file: supplement/tables/audit/C_pcal_ret.tex
\begin{table}[t]
\centering\footnotesize\setlength{\tabcolsep}{6pt}
\begin{tabular}{lrr}
\toprule
$p$ & $\to$ NAB-348 (\%) & $\to$ FineView (\%) \\
\midrule
0 & 23.1 & 12.4 \\
0.25 & 19.5 & 7.9 \\
0.5 & 18.8 & 9.5 \\
0.75 & 20.8 & 10.8 \\
1 & 20.9 & 15.4 \\
\bottomrule
\end{tabular}
\caption{Historical NAME-drop calibration, CUB-trained DFN5B H/14-378: relative held-out P3 decrease, $(\mathrm{P3}_{\mathrm{peak}}-\mathrm{P3}_{\mathrm{P1\text{-}selected}})/\mathrm{P3}_{\mathrm{peak}}$, where the P1-selected checkpoint is each run's own highest-CUB-P1 epoch. Values are those of the archived retention analysis (they agree with recomputation from the peak table's source to rounding). One run per $p$; the $p{=}0$ point is the baseline SSR run.}
\label{tab:C_pcal_ret}
\end{table}

%% file: supplement/tables/generated_supplement_tables.tex
% AUTO-GENERATED by scripts/build_supplement_tables.py -- do not hand-edit.

\subsection{ImageNet target trajectories}
\label{app:D_target}
Table~\ref{tab:S1_full} gives, for every backbone and method, the P3 pretrained value, peak, deployment, and post-peak change on the ImageNet-1k validation target. PT is the pretrained checkpoint evaluated on ImageNet val. Differences below roughly 0.5 pp should be read cautiously (\S\ref{app:B_bootstrap}).

\begin{center}
\scriptsize\setlength{\tabcolsep}{3.5pt}
\begin{longtable}{lrrrrrr}
\caption{All six backbones, fine-tuned on ImageNet-1k and evaluated on ImageNet val (target); PT is the pretrained checkpoint evaluated directly on ImageNet val. Deploy is each run's own P1-selected checkpoint and peak is the maximum P3 over the evaluated trajectory; best peak and deployment values are bolded independently within each backbone. Structured supervision produces substantially larger P3 acquisition than N/U, while U often shows an early gain that is reduced by deployment. Numbers in parentheses after P3 peak and P3 deploy give the ImageNet training milestone index (1--11), not an optimizer step count or epoch: milestones are 11 checkpoints at geometrically-spaced cumulative ImageNet-1k data fractions (0.25, 0.5, 1, 2, 5, 10, 20, 40, 60, 80, 100\%), with PT as milestone 0. Because each later milestone spans a much larger increment of data than each earlier one, the milestone index is not proportional to the number of gradient updates or to wall-clock training progress. Deploy is each run's own ImageNet-val P1-selected milestone, shared across all evaluation datasets for that run; peak is the milestone of the maximum P3 on that evaluation dataset, which can therefore differ from the deployment milestone and from peak timing on other datasets. When multiple milestones attain the same maximum P3, the earliest is reported.}\label{tab:S1_full}\\
\toprule
Method & P3 PT & P3 peak (milestone) & Gain$_{peak}$ & P3 deploy (milestone) & Gain$_{deploy}$ & Forget$_{post-peak}$ \\
\midrule
\endfirsthead
\toprule
Method & P3 PT & P3 peak (milestone) & Gain$_{peak}$ & P3 deploy (milestone) & Gain$_{deploy}$ & Forget$_{post-peak}$ \\
\midrule
\endhead
\midrule \multicolumn{7}{r}{\emph{continued on next page}} \\ \midrule
\endfoot
\bottomrule
\endlastfoot
\multicolumn{7}{l}{\textbf{ImageNet val (target)}} \\
\midrule
N & 33.12 & 33.78 (11) & 0.66 & 33.78 (11) & 0.66 & 0.00 \\
\noalign{\penalty10000}
U & 33.12 & 35.38 (2) & 2.26 & 24.80 (11) & -8.32 & 10.58 \\
\noalign{\penalty10000}
SSR & 33.12 & 45.34 (10) & 12.22 & 45.12 (11) & 12.00 & 0.22 \\
\noalign{\penalty10000}
SSR+ND & 33.12 & \textbf{48.12 (9)} & \textbf{15.00} & \textbf{48.00 (11)} & \textbf{14.88} & 0.12 \\
\noalign{\penalty10000}
\addlinespace[1pt]
\noalign{\penalty10000}
\multicolumn{7}{l}{\scriptsize DFN5B H/14-378} \\
\midrule
N & 28.60 & 30.06 (11) & 1.46 & 30.06 (11) & 1.46 & 0.00 \\
\noalign{\penalty10000}
U & 28.60 & 29.44 (2) & 0.84 & 28.44 (11) & -0.16 & 1.00 \\
\noalign{\penalty10000}
SSR & 28.60 & 33.90 (10) & 5.30 & 33.56 (11) & 4.96 & 0.34 \\
\noalign{\penalty10000}
SSR+ND & 28.60 & \textbf{36.70 (11)} & \textbf{8.10} & \textbf{36.70 (11)} & \textbf{8.10} & 0.00 \\
\noalign{\penalty10000}
\addlinespace[1pt]
\noalign{\penalty10000}
\multicolumn{7}{l}{\scriptsize DFN2B L/14} \\
\midrule
N & 35.94 & 44.08 (11) & 8.14 & 44.08 (11) & 8.14 & 0.00 \\
\noalign{\penalty10000}
U & 35.94 & 38.00 (2) & 2.06 & 32.44 (11) & -3.50 & 5.56 \\
\noalign{\penalty10000}
SSR & 35.94 & 43.04 (11) & 7.10 & 43.04 (11) & 7.10 & 0.00 \\
\noalign{\penalty10000}
SSR+ND & 35.94 & \textbf{50.60 (11)} & \textbf{14.66} & \textbf{50.60 (11)} & \textbf{14.66} & 0.00 \\
\noalign{\penalty10000}
\addlinespace[1pt]
\noalign{\penalty10000}
\multicolumn{7}{l}{\scriptsize MetaCLIP H/14} \\
\midrule
N & 31.12 & 35.86 (8) & 4.74 & 35.48 (11) & 4.36 & 0.38 \\
\noalign{\penalty10000}
U & 31.12 & 32.68 (1) & 1.56 & 29.36 (11) & -1.76 & 3.32 \\
\noalign{\penalty10000}
SSR & 31.12 & 37.70 (4) & 6.58 & 36.04 (11) & 4.92 & 1.66 \\
\noalign{\penalty10000}
SSR+ND & 31.12 & \textbf{43.86 (11)} & \textbf{12.74} & \textbf{43.86 (11)} & \textbf{12.74} & 0.00 \\
\noalign{\penalty10000}
\addlinespace[1pt]
\noalign{\penalty10000}
\multicolumn{7}{l}{\scriptsize MetaCLIP L/14} \\
\midrule
N & 25.48 & 30.72 (11) & 5.24 & 30.72 (11) & 5.24 & 0.00 \\
\noalign{\penalty10000}
U & 25.48 & 27.76 (11) & 2.28 & 27.76 (11) & 2.28 & 0.00 \\
\noalign{\penalty10000}
SSR & 25.48 & 35.14 (11) & 9.66 & 35.14 (11) & 9.66 & 0.00 \\
\noalign{\penalty10000}
SSR+ND & 25.48 & \textbf{45.36 (11)} & \textbf{19.88} & \textbf{45.36 (11)} & \textbf{19.88} & 0.00 \\
\noalign{\penalty10000}
\addlinespace[1pt]
\noalign{\penalty10000}
\multicolumn{7}{l}{\scriptsize OpenAI L/14} \\
\midrule
N & 20.44 & 21.98 (5) & 1.54 & 21.68 (11) & 1.24 & 0.30 \\
\noalign{\penalty10000}
U & 20.44 & 20.24 (2) & -0.20 & 19.92 (11) & -0.52 & 0.32 \\
\noalign{\penalty10000}
SSR & 20.44 & 24.62 (11) & 4.18 & 24.62 (11) & 4.18 & 0.00 \\
\noalign{\penalty10000}
SSR+ND & 20.44 & \textbf{29.42 (11)} & \textbf{8.98} & \textbf{29.42 (11)} & \textbf{8.98} & 0.00 \\
\noalign{\penalty10000}
\addlinespace[1pt]
\noalign{\penalty10000}
\multicolumn{7}{l}{\scriptsize OpenAI B/32} \\
\midrule
\end{longtable}
\end{center}

\FloatBarrier
\subsection{ImageNet same-ontology shifts}
\label{app:D_shift}
Table~\ref{tab:S1b_shift} reports the same quantities on ImageNet-V2, ImageNet-R, and ImageNet-Sketch, which share the ImageNet class ontology with the training data. PT is re-evaluated on each shift dataset.

\begin{center}
\scriptsize\setlength{\tabcolsep}{3.5pt}
\begin{longtable}{lrrrrrr}
\caption{All six backbones, fine-tuned on ImageNet-1k and evaluated on the same-ontology shifts ImageNet-V2/-R/-Sketch; PT is the pretrained checkpoint evaluated directly on each listed evaluation dataset. Deploy is each run's own (ImageNet-val) P1-selected checkpoint and peak is the maximum P3 over the evaluated trajectory on that dataset; best values are bolded independently within each backbone. Numbers in parentheses after P3 peak and P3 deploy give the ImageNet training milestone index (1--11), not an optimizer step count or epoch: milestones are 11 checkpoints at geometrically-spaced cumulative ImageNet-1k data fractions (0.25, 0.5, 1, 2, 5, 10, 20, 40, 60, 80, 100\%), with PT as milestone 0. Because each later milestone spans a much larger increment of data than each earlier one, the milestone index is not proportional to the number of gradient updates or to wall-clock training progress. Deploy is each run's own ImageNet-val P1-selected milestone, shared across all evaluation datasets for that run; peak is the milestone of the maximum P3 on that evaluation dataset, which can therefore differ from the deployment milestone and from peak timing on other datasets. When multiple milestones attain the same maximum P3, the earliest is reported.}\label{tab:S1b_shift}\\
\toprule
Method & P3 PT & P3 peak (milestone) & Gain$_{peak}$ & P3 deploy (milestone) & Gain$_{deploy}$ & Forget$_{post-peak}$ \\
\midrule
\endfirsthead
\toprule
Method & P3 PT & P3 peak (milestone) & Gain$_{peak}$ & P3 deploy (milestone) & Gain$_{deploy}$ & Forget$_{post-peak}$ \\
\midrule
\endhead
\midrule \multicolumn{7}{r}{\emph{continued on next page}} \\ \midrule
\endfoot
\bottomrule
\endlastfoot
\multicolumn{7}{l}{\textbf{ImageNet-V2 (same-ontology shift)}} \\
\midrule
N & 29.29 & 30.76 (11) & 1.47 & 30.76 (11) & 1.47 & 0.00 \\
\noalign{\penalty10000}
U & 29.29 & 30.90 (3) & 1.61 & 22.24 (11) & -7.04 & 8.66 \\
\noalign{\penalty10000}
SSR & 29.29 & 40.68 (10) & 11.39 & 40.48 (11) & 11.19 & 0.20 \\
\noalign{\penalty10000}
SSR+ND & 29.29 & \textbf{42.83 (10)} & \textbf{13.54} & \textbf{42.68 (11)} & \textbf{13.39} & 0.15 \\
\noalign{\penalty10000}
\addlinespace[1pt]
\noalign{\penalty10000}
\multicolumn{7}{l}{\scriptsize DFN5B H/14-378} \\
\midrule
N & 24.51 & 25.85 (11) & 1.34 & 25.85 (11) & 1.34 & 0.00 \\
\noalign{\penalty10000}
U & 24.51 & 25.43 (3) & 0.92 & 24.82 (11) & 0.31 & 0.61 \\
\noalign{\penalty10000}
SSR & 24.51 & 29.81 (10) & 5.30 & 29.78 (11) & 5.27 & 0.03 \\
\noalign{\penalty10000}
SSR+ND & 24.51 & \textbf{32.06 (11)} & \textbf{7.56} & \textbf{32.06 (11)} & \textbf{7.56} & 0.00 \\
\noalign{\penalty10000}
\addlinespace[1pt]
\noalign{\penalty10000}
\multicolumn{7}{l}{\scriptsize DFN2B L/14} \\
\midrule
N & 31.66 & 38.58 (11) & 6.91 & 38.58 (11) & 6.91 & 0.00 \\
\noalign{\penalty10000}
U & 31.66 & 33.53 (2) & 1.86 & 28.50 (11) & -3.17 & 5.03 \\
\noalign{\penalty10000}
SSR & 31.66 & 37.30 (11) & 5.64 & 37.30 (11) & 5.64 & 0.00 \\
\noalign{\penalty10000}
SSR+ND & 31.66 & \textbf{44.96 (11)} & \textbf{13.30} & \textbf{44.96 (11)} & \textbf{13.30} & 0.00 \\
\noalign{\penalty10000}
\addlinespace[1pt]
\noalign{\penalty10000}
\multicolumn{7}{l}{\scriptsize MetaCLIP H/14} \\
\midrule
N & 27.39 & 31.64 (11) & 4.25 & 31.64 (11) & 4.25 & 0.00 \\
\noalign{\penalty10000}
U & 27.39 & 30.05 (2) & 2.66 & 25.63 (11) & -1.76 & 4.42 \\
\noalign{\penalty10000}
SSR & 27.39 & 32.92 (4) & 5.52 & 31.74 (11) & 4.35 & 1.17 \\
\noalign{\penalty10000}
SSR+ND & 27.39 & \textbf{37.94 (11)} & \textbf{10.54} & \textbf{37.94 (11)} & \textbf{10.54} & 0.00 \\
\noalign{\penalty10000}
\addlinespace[1pt]
\noalign{\penalty10000}
\multicolumn{7}{l}{\scriptsize MetaCLIP L/14} \\
\midrule
N & 22.73 & 27.46 (11) & 4.74 & 27.46 (11) & 4.74 & 0.00 \\
\noalign{\penalty10000}
U & 22.73 & 24.62 (11) & 1.89 & 24.62 (11) & 1.89 & 0.00 \\
\noalign{\penalty10000}
SSR & 22.73 & 31.47 (11) & 8.75 & 31.47 (11) & 8.75 & 0.00 \\
\noalign{\penalty10000}
SSR+ND & 22.73 & \textbf{39.68 (11)} & \textbf{16.95} & \textbf{39.68 (11)} & \textbf{16.95} & 0.00 \\
\noalign{\penalty10000}
\addlinespace[1pt]
\noalign{\penalty10000}
\multicolumn{7}{l}{\scriptsize OpenAI L/14} \\
\midrule
N & 17.16 & 18.38 (5) & 1.21 & 18.10 (11) & 0.93 & 0.28 \\
\noalign{\penalty10000}
U & 17.16 & 17.54 (2) & 0.37 & 17.00 (11) & -0.16 & 0.53 \\
\noalign{\penalty10000}
SSR & 17.16 & 20.93 (9) & 3.77 & 20.63 (11) & 3.47 & 0.30 \\
\noalign{\penalty10000}
SSR+ND & 17.16 & \textbf{24.42 (10)} & \textbf{7.25} & \textbf{24.20 (11)} & \textbf{7.03} & 0.22 \\
\noalign{\penalty10000}
\addlinespace[1pt]
\noalign{\penalty10000}
\multicolumn{7}{l}{\scriptsize OpenAI B/32} \\
\midrule
\multicolumn{7}{l}{\textbf{ImageNet-R (same-ontology shift)}} \\
\midrule
N & 35.40 & 43.52 (6) & 8.12 & 42.77 (11) & 7.37 & 0.75 \\
\noalign{\penalty10000}
U & 35.40 & 38.52 (3) & 3.12 & 32.33 (11) & -3.07 & 6.18 \\
\noalign{\penalty10000}
SSR & 35.40 & 51.33 (5) & 15.93 & 50.07 (11) & 14.67 & 1.27 \\
\noalign{\penalty10000}
SSR+ND & 35.40 & \textbf{53.60 (6)} & \textbf{18.20} & \textbf{53.13 (11)} & \textbf{17.73} & 0.47 \\
\noalign{\penalty10000}
\addlinespace[1pt]
\noalign{\penalty10000}
\multicolumn{7}{l}{\scriptsize DFN5B H/14-378} \\
\midrule
N & 30.62 & 31.22 (11) & 0.60 & 31.22 (11) & 0.60 & 0.00 \\
\noalign{\penalty10000}
U & 30.62 & 32.90 (4) & 2.28 & 31.08 (11) & 0.47 & 1.82 \\
\noalign{\penalty10000}
SSR & 30.62 & 35.45 (6) & 4.83 & 34.83 (11) & 4.22 & 0.62 \\
\noalign{\penalty10000}
SSR+ND & 30.62 & \textbf{38.50 (11)} & \textbf{7.88} & \textbf{38.50 (11)} & \textbf{7.88} & 0.00 \\
\noalign{\penalty10000}
\addlinespace[1pt]
\noalign{\penalty10000}
\multicolumn{7}{l}{\scriptsize DFN2B L/14} \\
\midrule
N & 42.87 & 49.25 (10) & 6.38 & 49.20 (11) & 6.33 & 0.05 \\
\noalign{\penalty10000}
U & 42.87 & 47.12 (1) & 4.25 & 44.33 (11) & 1.47 & 2.78 \\
\noalign{\penalty10000}
SSR & 42.87 & 52.45 (4) & 9.58 & 46.80 (11) & 3.93 & 5.65 \\
\noalign{\penalty10000}
SSR+ND & 42.87 & \textbf{58.20 (11)} & \textbf{15.33} & \textbf{58.20 (11)} & \textbf{15.33} & 0.00 \\
\noalign{\penalty10000}
\addlinespace[1pt]
\noalign{\penalty10000}
\multicolumn{7}{l}{\scriptsize MetaCLIP H/14} \\
\midrule
N & 36.35 & 41.70 (11) & 5.35 & 41.70 (11) & 5.35 & 0.00 \\
\noalign{\penalty10000}
U & 36.35 & 41.92 (2) & 5.57 & 38.63 (11) & 2.28 & 3.28 \\
\noalign{\penalty10000}
SSR & 36.35 & 46.28 (4) & 9.93 & 40.25 (11) & 3.90 & 6.03 \\
\noalign{\penalty10000}
SSR+ND & 36.35 & \textbf{48.65 (5)} & \textbf{12.30} & \textbf{47.63 (11)} & \textbf{11.28} & 1.02 \\
\noalign{\penalty10000}
\addlinespace[1pt]
\noalign{\penalty10000}
\multicolumn{7}{l}{\scriptsize MetaCLIP L/14} \\
\midrule
N & 31.45 & 37.92 (11) & 6.47 & 37.92 (11) & 6.47 & 0.00 \\
\noalign{\penalty10000}
U & 31.45 & 37.28 (11) & 5.83 & 37.28 (11) & 5.83 & 0.00 \\
\noalign{\penalty10000}
SSR & 31.45 & 42.33 (11) & 10.88 & 42.33 (11) & 10.88 & 0.00 \\
\noalign{\penalty10000}
SSR+ND & 31.45 & \textbf{53.08 (11)} & \textbf{21.63} & \textbf{53.08 (11)} & \textbf{21.63} & 0.00 \\
\noalign{\penalty10000}
\addlinespace[1pt]
\noalign{\penalty10000}
\multicolumn{7}{l}{\scriptsize OpenAI L/14} \\
\midrule
N & 22.35 & 26.08 (5) & 3.73 & 24.13 (11) & 1.78 & 1.95 \\
\noalign{\penalty10000}
U & 22.35 & 25.10 (2) & 2.75 & 24.82 (11) & 2.47 & 0.28 \\
\noalign{\penalty10000}
SSR & 22.35 & 27.25 (6) & 4.90 & 24.83 (11) & 2.48 & 2.42 \\
\noalign{\penalty10000}
SSR+ND & 22.35 & \textbf{29.45 (6)} & \textbf{7.10} & \textbf{29.08 (11)} & \textbf{6.73} & 0.37 \\
\noalign{\penalty10000}
\addlinespace[1pt]
\noalign{\penalty10000}
\multicolumn{7}{l}{\scriptsize OpenAI B/32} \\
\midrule
\multicolumn{7}{l}{\textbf{ImageNet-Sketch (same-ontology shift)}} \\
\midrule
N & 28.35 & 30.67 (11) & 2.32 & 30.67 (11) & 2.32 & 0.00 \\
\noalign{\penalty10000}
U & 28.35 & 30.55 (3) & 2.20 & 21.37 (11) & -6.98 & 9.19 \\
\noalign{\penalty10000}
SSR & 28.35 & 37.43 (10) & 9.08 & 37.39 (11) & 9.04 & 0.04 \\
\noalign{\penalty10000}
SSR+ND & 28.35 & \textbf{39.43 (10)} & \textbf{11.08} & \textbf{39.40 (11)} & \textbf{11.05} & 0.03 \\
\noalign{\penalty10000}
\addlinespace[1pt]
\noalign{\penalty10000}
\multicolumn{7}{l}{\scriptsize DFN5B H/14-378} \\
\midrule
N & 23.85 & 23.79 (1) & -0.06 & 23.79 (11) & -0.06 & 0.00 \\
\noalign{\penalty10000}
U & 23.85 & 24.56 (3) & 0.71 & 23.40 (11) & -0.45 & 1.16 \\
\noalign{\penalty10000}
SSR & 23.85 & 27.70 (11) & 3.85 & 27.70 (11) & 3.85 & 0.00 \\
\noalign{\penalty10000}
SSR+ND & 23.85 & \textbf{29.25 (11)} & \textbf{5.40} & \textbf{29.25 (11)} & \textbf{5.40} & 0.00 \\
\noalign{\penalty10000}
\addlinespace[1pt]
\noalign{\penalty10000}
\multicolumn{7}{l}{\scriptsize DFN2B L/14} \\
\midrule
N & 29.31 & 36.24 (11) & 6.93 & 36.24 (11) & 6.93 & 0.00 \\
\noalign{\penalty10000}
U & 29.31 & 32.42 (2) & 3.11 & 28.44 (11) & -0.88 & 3.98 \\
\noalign{\penalty10000}
SSR & 29.31 & 34.52 (5) & 5.21 & 33.15 (11) & 3.83 & 1.38 \\
\noalign{\penalty10000}
SSR+ND & 29.31 & \textbf{38.80 (11)} & \textbf{9.49} & \textbf{38.80 (11)} & \textbf{9.49} & 0.00 \\
\noalign{\penalty10000}
\addlinespace[1pt]
\noalign{\penalty10000}
\multicolumn{7}{l}{\scriptsize MetaCLIP H/14} \\
\midrule
N & 25.44 & 28.92 (11) & 3.47 & 28.92 (11) & 3.47 & 0.00 \\
\noalign{\penalty10000}
U & 25.44 & 29.06 (2) & 3.61 & 24.72 (11) & -0.72 & 4.34 \\
\noalign{\penalty10000}
SSR & 25.44 & 29.92 (4) & 4.48 & 28.32 (11) & 2.87 & 1.60 \\
\noalign{\penalty10000}
SSR+ND & 25.44 & \textbf{31.74 (11)} & \textbf{6.30} & \textbf{31.74 (11)} & \textbf{6.30} & 0.00 \\
\noalign{\penalty10000}
\addlinespace[1pt]
\noalign{\penalty10000}
\multicolumn{7}{l}{\scriptsize MetaCLIP L/14} \\
\midrule
N & 19.41 & 24.96 (11) & 5.55 & 24.96 (11) & 5.55 & 0.00 \\
\noalign{\penalty10000}
U & 19.41 & 21.86 (11) & 2.46 & 21.86 (11) & 2.46 & 0.00 \\
\noalign{\penalty10000}
SSR & 19.41 & 26.72 (11) & 7.31 & 26.72 (11) & 7.31 & 0.00 \\
\noalign{\penalty10000}
SSR+ND & 19.41 & \textbf{31.88 (11)} & \textbf{12.47} & \textbf{31.88 (11)} & \textbf{12.47} & 0.00 \\
\noalign{\penalty10000}
\addlinespace[1pt]
\noalign{\penalty10000}
\multicolumn{7}{l}{\scriptsize OpenAI L/14} \\
\midrule
N & 12.70 & 15.08 (5) & 2.38 & 14.30 (11) & 1.60 & 0.78 \\
\noalign{\penalty10000}
U & 12.70 & 14.07 (2) & 1.37 & 13.10 (11) & 0.40 & 0.97 \\
\noalign{\penalty10000}
SSR & 12.70 & 15.07 (6) & 2.37 & 14.72 (11) & 2.02 & 0.35 \\
\noalign{\penalty10000}
SSR+ND & 12.70 & \textbf{17.25 (11)} & \textbf{4.56} & \textbf{17.25 (11)} & \textbf{4.56} & 0.00 \\
\noalign{\penalty10000}
\addlinespace[1pt]
\noalign{\penalty10000}
\multicolumn{7}{l}{\scriptsize OpenAI B/32} \\
\midrule
\end{longtable}
\end{center}

\FloatBarrier
\subsection{ImageNet-trained cross-ontology trajectories}
\label{app:D_cross}
Table~\ref{tab:S2_full} evaluates the ImageNet-trained models on CUB-200, NAB-348 (strict dataset-class-disjoint held-out), and FineView Butterfly. PT is re-evaluated on each held-out dataset; it is not carried over from ImageNet.

\begin{center}
\scriptsize\setlength{\tabcolsep}{3.5pt}
\begin{longtable}{lrrrrrr}
\caption{All six backbones, fine-tuned on ImageNet-1k and evaluated cross-ontology on CUB-200, NAB-348 (strict dataset-class-disjoint held-out), and FineView Butterfly (FVB); PT is the pretrained checkpoint evaluated directly on each held-out dataset. Deploy is each run's own P1-selected checkpoint (shared with Table~\ref{tab:S1_full}) and peak is the maximum P3 over the trajectory; best peak and deployment values are bolded independently within each backbone (trajectory regimes in Table~\ref{tab:S5_regime}). Cross-ontology P3 acquisition is substantially smaller and more backbone-dependent than acquisition on ImageNet and its same-ontology shifts, and its peak often occurs earlier than the shared deployment milestone. Numbers in parentheses after P3 peak and P3 deploy give the ImageNet training milestone index (1--11), not an optimizer step count or epoch: milestones are 11 checkpoints at geometrically-spaced cumulative ImageNet-1k data fractions (0.25, 0.5, 1, 2, 5, 10, 20, 40, 60, 80, 100\%), with PT as milestone 0. Because each later milestone spans a much larger increment of data than each earlier one, the milestone index is not proportional to the number of gradient updates or to wall-clock training progress. Deploy is each run's own ImageNet-val P1-selected milestone, shared across all evaluation datasets for that run; peak is the milestone of the maximum P3 on that evaluation dataset, which can therefore differ from the deployment milestone and from peak timing on other datasets. When multiple milestones attain the same maximum P3, the earliest is reported.}\label{tab:S2_full}\\
\toprule
Method & P3 PT & P3 peak (milestone) & Gain$_{peak}$ & P3 deploy (milestone) & Gain$_{deploy}$ & Forget$_{post-peak}$ \\
\midrule
\endfirsthead
\toprule
Method & P3 PT & P3 peak (milestone) & Gain$_{peak}$ & P3 deploy (milestone) & Gain$_{deploy}$ & Forget$_{post-peak}$ \\
\midrule
\endhead
\midrule \multicolumn{7}{r}{\emph{continued on next page}} \\ \midrule
\endfoot
\bottomrule
\endlastfoot
\multicolumn{7}{l}{\textbf{CUB-200 (cross-ontology held-out)}} \\
\midrule
N & 34.36 & 33.64 (1) & -0.72 & 20.73 (11) & -13.63 & 12.91 \\
\noalign{\penalty10000}
U & 34.36 & \textbf{35.48 (1)} & \textbf{1.12} & 15.26 (11) & -19.11 & 20.23 \\
\noalign{\penalty10000}
SSR & 34.36 & 34.76 (2) & 0.40 & 23.35 (11) & -11.01 & 11.41 \\
\noalign{\penalty10000}
SSR+ND & 34.36 & 34.88 (3) & 0.52 & \textbf{24.99 (11)} & \textbf{-9.37} & 9.89 \\
\noalign{\penalty10000}
\addlinespace[1pt]
\noalign{\penalty10000}
\multicolumn{7}{l}{\scriptsize DFN5B H/14-378} \\
\midrule
N & 27.96 & 27.96 (1) & 0.00 & 25.92 (11) & -2.04 & 2.04 \\
\noalign{\penalty10000}
U & 27.96 & 27.65 (1) & -0.31 & 20.14 (11) & -7.82 & 7.51 \\
\noalign{\penalty10000}
SSR & 27.96 & 28.15 (5) & 0.19 & 26.01 (11) & -1.95 & 2.14 \\
\noalign{\penalty10000}
SSR+ND & 27.96 & \textbf{28.27 (1)} & \textbf{0.31} & \textbf{26.92 (11)} & \textbf{-1.04} & 1.35 \\
\noalign{\penalty10000}
\addlinespace[1pt]
\noalign{\penalty10000}
\multicolumn{7}{l}{\scriptsize DFN2B L/14} \\
\midrule
N & 28.22 & 30.57 (6) & 2.35 & 27.74 (11) & -0.48 & 2.83 \\
\noalign{\penalty10000}
U & 28.22 & 23.16 (1) & -5.06 & 17.09 (11) & -11.13 & 6.08 \\
\noalign{\penalty10000}
SSR & 28.22 & 30.70 (3) & 2.49 & 26.79 (11) & -1.43 & 3.92 \\
\noalign{\penalty10000}
SSR+ND & 28.22 & \textbf{31.27 (3)} & \textbf{3.05} & \textbf{29.50 (11)} & \textbf{1.28} & 1.78 \\
\noalign{\penalty10000}
\addlinespace[1pt]
\noalign{\penalty10000}
\multicolumn{7}{l}{\scriptsize MetaCLIP H/14} \\
\midrule
N & 23.90 & 24.84 (7) & 0.93 & 24.70 (11) & 0.79 & 0.14 \\
\noalign{\penalty10000}
U & 23.90 & 19.78 (1) & -4.12 & 18.57 (11) & -5.33 & 1.21 \\
\noalign{\penalty10000}
SSR & 23.90 & 25.51 (3) & 1.61 & 22.35 (11) & -1.55 & 3.16 \\
\noalign{\penalty10000}
SSR+ND & 23.90 & \textbf{28.29 (10)} & \textbf{4.38} & \textbf{27.17 (11)} & \textbf{3.26} & 1.12 \\
\noalign{\penalty10000}
\addlinespace[1pt]
\noalign{\penalty10000}
\multicolumn{7}{l}{\scriptsize MetaCLIP L/14} \\
\midrule
N & 16.71 & 17.41 (1) & 0.71 & 16.19 (11) & -0.52 & 1.23 \\
\noalign{\penalty10000}
U & 16.71 & 16.64 (11) & -0.07 & 16.64 (11) & -0.07 & 0.00 \\
\noalign{\penalty10000}
SSR & 16.71 & 18.97 (3) & 2.26 & 17.26 (11) & 0.55 & 1.71 \\
\noalign{\penalty10000}
SSR+ND & 16.71 & \textbf{20.54 (8)} & \textbf{3.83} & \textbf{19.35 (11)} & \textbf{2.64} & 1.19 \\
\noalign{\penalty10000}
\addlinespace[1pt]
\noalign{\penalty10000}
\multicolumn{7}{l}{\scriptsize OpenAI L/14} \\
\midrule
N & 13.76 & 13.53 (1) & -0.22 & 12.46 (11) & -1.29 & 1.07 \\
\noalign{\penalty10000}
U & 13.76 & 12.58 (1) & -1.17 & 11.81 (11) & -1.95 & 0.78 \\
\noalign{\penalty10000}
SSR & 13.76 & \textbf{15.53 (7)} & \textbf{1.78} & \textbf{14.70 (11)} & \textbf{0.95} & 0.83 \\
\noalign{\penalty10000}
SSR+ND & 13.76 & 15.22 (8) & 1.47 & 14.27 (11) & 0.52 & 0.95 \\
\noalign{\penalty10000}
\addlinespace[1pt]
\noalign{\penalty10000}
\multicolumn{7}{l}{\scriptsize OpenAI B/32} \\
\midrule
\multicolumn{7}{l}{\textbf{NAB-348 (strict cross-ontology held-out)}} \\
\midrule
N & 13.81 & 13.41 (1) & -0.41 & 9.25 (11) & -4.56 & 4.15 \\
\noalign{\penalty10000}
U & 13.81 & 14.28 (1) & 0.47 & 4.61 (11) & -9.21 & 9.67 \\
\noalign{\penalty10000}
SSR & 13.81 & 14.68 (6) & 0.87 & 11.25 (11) & -2.56 & 3.43 \\
\noalign{\penalty10000}
SSR+ND & 13.81 & \textbf{15.04 (6)} & \textbf{1.23} & \textbf{12.22 (11)} & \textbf{-1.59} & 2.82 \\
\noalign{\penalty10000}
\addlinespace[1pt]
\noalign{\penalty10000}
\multicolumn{7}{l}{\scriptsize DFN5B H/14-378} \\
\midrule
N & 8.85 & 8.84 (1) & -0.01 & 7.07 (11) & -1.78 & 1.76 \\
\noalign{\penalty10000}
U & 8.85 & 8.65 (1) & -0.20 & 4.09 (11) & -4.76 & 4.56 \\
\noalign{\penalty10000}
SSR & 8.85 & 8.97 (2) & 0.11 & \textbf{8.17 (11)} & \textbf{-0.68} & 0.80 \\
\noalign{\penalty10000}
SSR+ND & 8.85 & \textbf{9.26 (4)} & \textbf{0.41} & 7.44 (11) & -1.41 & 1.82 \\
\noalign{\penalty10000}
\addlinespace[1pt]
\noalign{\penalty10000}
\multicolumn{7}{l}{\scriptsize DFN2B L/14} \\
\midrule
N & 11.59 & \textbf{12.32 (5)} & \textbf{0.72} & \textbf{10.75 (11)} & \textbf{-0.84} & 1.57 \\
\noalign{\penalty10000}
U & 11.59 & 8.49 (1) & -3.11 & 3.45 (11) & -8.15 & 5.04 \\
\noalign{\penalty10000}
SSR & 11.59 & 12.06 (2) & 0.47 & 10.08 (11) & -1.52 & 1.99 \\
\noalign{\penalty10000}
SSR+ND & 11.59 & 12.17 (1) & 0.58 & 9.79 (11) & -1.81 & 2.38 \\
\noalign{\penalty10000}
\addlinespace[1pt]
\noalign{\penalty10000}
\multicolumn{7}{l}{\scriptsize MetaCLIP H/14} \\
\midrule
N & 7.02 & 6.76 (1) & -0.25 & \textbf{6.25 (11)} & \textbf{-0.77} & 0.52 \\
\noalign{\penalty10000}
U & 7.02 & 4.20 (1) & -2.82 & 3.70 (11) & -3.32 & 0.50 \\
\noalign{\penalty10000}
SSR & 7.02 & 7.04 (2) & 0.02 & 4.17 (11) & -2.85 & 2.87 \\
\noalign{\penalty10000}
SSR+ND & 7.02 & \textbf{7.38 (1)} & \textbf{0.36} & 5.78 (11) & -1.24 & 1.60 \\
\noalign{\penalty10000}
\addlinespace[1pt]
\noalign{\penalty10000}
\multicolumn{7}{l}{\scriptsize MetaCLIP L/14} \\
\midrule
N & 7.41 & 7.57 (2) & 0.16 & 5.80 (11) & -1.60 & 1.77 \\
\noalign{\penalty10000}
U & 7.41 & 5.45 (1) & -1.96 & 4.14 (11) & -3.27 & 1.30 \\
\noalign{\penalty10000}
SSR & 7.41 & 8.05 (3) & 0.64 & 6.70 (11) & -0.71 & 1.35 \\
\noalign{\penalty10000}
SSR+ND & 7.41 & \textbf{8.40 (8)} & \textbf{0.99} & \textbf{7.56 (11)} & \textbf{0.15} & 0.84 \\
\noalign{\penalty10000}
\addlinespace[1pt]
\noalign{\penalty10000}
\multicolumn{7}{l}{\scriptsize OpenAI L/14} \\
\midrule
N & 4.23 & 4.12 (1) & -0.11 & 3.48 (11) & -0.75 & 0.64 \\
\noalign{\penalty10000}
U & 4.23 & 3.37 (10) & -0.87 & 3.18 (11) & -1.05 & 0.18 \\
\noalign{\penalty10000}
SSR & 4.23 & 4.89 (7) & 0.66 & 4.62 (11) & 0.39 & 0.27 \\
\noalign{\penalty10000}
SSR+ND & 4.23 & \textbf{4.96 (11)} & \textbf{0.73} & \textbf{4.96 (11)} & \textbf{0.73} & 0.00 \\
\noalign{\penalty10000}
\addlinespace[1pt]
\noalign{\penalty10000}
\multicolumn{7}{l}{\scriptsize OpenAI B/32} \\
\midrule
\multicolumn{7}{l}{\textbf{FineView Butterfly (FVB) (cross-ontology held-out)}} \\
\midrule
N & 8.52 & 8.42 (1) & -0.10 & 6.42 (11) & -2.10 & 2.00 \\
\noalign{\penalty10000}
U & 8.52 & 8.75 (3) & 0.23 & 4.06 (11) & -4.46 & 4.69 \\
\noalign{\penalty10000}
SSR & 8.52 & \textbf{8.85 (6)} & \textbf{0.33} & \textbf{7.13 (11)} & \textbf{-1.38} & 1.71 \\
\noalign{\penalty10000}
SSR+ND & 8.52 & 8.38 (1) & -0.13 & 6.77 (11) & -1.75 & 1.62 \\
\noalign{\penalty10000}
\addlinespace[1pt]
\noalign{\penalty10000}
\multicolumn{7}{l}{\scriptsize DFN5B H/14-378} \\
\midrule
N & 6.67 & 6.85 (11) & 0.17 & 6.85 (11) & 0.17 & 0.00 \\
\noalign{\penalty10000}
U & 6.67 & 6.44 (1) & -0.23 & 5.87 (11) & -0.81 & 0.58 \\
\noalign{\penalty10000}
SSR & 6.67 & \textbf{7.71 (11)} & \textbf{1.04} & \textbf{7.71 (11)} & \textbf{1.04} & 0.00 \\
\noalign{\penalty10000}
SSR+ND & 6.67 & 7.44 (11) & 0.77 & 7.44 (11) & 0.77 & 0.00 \\
\noalign{\penalty10000}
\addlinespace[1pt]
\noalign{\penalty10000}
\multicolumn{7}{l}{\scriptsize DFN2B L/14} \\
\midrule
N & 6.52 & 6.77 (5) & 0.25 & 6.19 (11) & -0.33 & 0.58 \\
\noalign{\penalty10000}
U & 6.52 & 5.48 (1) & -1.04 & 5.00 (11) & -1.52 & 0.48 \\
\noalign{\penalty10000}
SSR & 6.52 & \textbf{7.13 (6)} & \textbf{0.62} & 5.98 (11) & -0.54 & 1.15 \\
\noalign{\penalty10000}
SSR+ND & 6.52 & 7.04 (11) & 0.52 & \textbf{7.04 (11)} & \textbf{0.52} & 0.00 \\
\noalign{\penalty10000}
\addlinespace[1pt]
\noalign{\penalty10000}
\multicolumn{7}{l}{\scriptsize MetaCLIP H/14} \\
\midrule
N & 4.73 & 5.10 (11) & 0.37 & 5.10 (11) & 0.37 & 0.00 \\
\noalign{\penalty10000}
U & 4.73 & 4.71 (7) & -0.02 & 4.44 (11) & -0.29 & 0.27 \\
\noalign{\penalty10000}
SSR & 4.73 & 5.13 (8) & 0.40 & 4.73 (11) & 0.00 & 0.40 \\
\noalign{\penalty10000}
SSR+ND & 4.73 & \textbf{5.58 (10)} & \textbf{0.85} & \textbf{5.31 (11)} & \textbf{0.58} & 0.27 \\
\noalign{\penalty10000}
\addlinespace[1pt]
\noalign{\penalty10000}
\multicolumn{7}{l}{\scriptsize MetaCLIP L/14} \\
\midrule
N & 3.75 & 4.13 (5) & 0.38 & 3.25 (11) & -0.50 & 0.88 \\
\noalign{\penalty10000}
U & 3.75 & 4.12 (7) & 0.37 & 3.42 (11) & -0.33 & 0.69 \\
\noalign{\penalty10000}
SSR & 3.75 & \textbf{5.25 (5)} & \textbf{1.50} & 3.62 (11) & -0.13 & 1.63 \\
\noalign{\penalty10000}
SSR+ND & 3.75 & 4.23 (7) & 0.48 & \textbf{3.77 (11)} & \textbf{0.02} & 0.46 \\
\noalign{\penalty10000}
\addlinespace[1pt]
\noalign{\penalty10000}
\multicolumn{7}{l}{\scriptsize OpenAI L/14} \\
\midrule
N & 2.73 & 3.63 (4) & 0.90 & 3.13 (11) & 0.40 & 0.50 \\
\noalign{\penalty10000}
U & 2.73 & 3.23 (2) & 0.50 & 2.62 (11) & -0.12 & 0.62 \\
\noalign{\penalty10000}
SSR & 2.73 & 4.02 (9) & 1.29 & 3.44 (11) & 0.71 & 0.58 \\
\noalign{\penalty10000}
SSR+ND & 2.73 & \textbf{4.04 (9)} & \textbf{1.31} & \textbf{3.87 (11)} & \textbf{1.13} & 0.17 \\
\noalign{\penalty10000}
\addlinespace[1pt]
\noalign{\penalty10000}
\multicolumn{7}{l}{\scriptsize OpenAI B/32} \\
\midrule
\end{longtable}
\end{center}

\FloatBarrier
\subsection{Dense representative CUB$\to$NAB trajectory}
\label{app:D_dense}
Table~\ref{tab:S3_dense} gives the epoch-level summary underlying Figure~\ref{fig:acquisition}a for the representative DFN5B H/14-378 CUB-trained runs. This is a single representative backbone, not a six-backbone result.

\begin{center}
\small
\begin{longtable}{lrrrrrrrr}
\caption{Representative DFN5B H/14-378 fine-tuned on CUB-200, with dense trajectory evaluation on CUB-200 (target) and NAB-348 (strict cross-ontology held-out); PT is the pretrained checkpoint evaluated directly on each dataset. Peak is the maximum P3 over the evaluated trajectory and deploy is each method's own P1-selected checkpoint; best peak and deployment values are bolded independently within each dataset. All four fine-tuning conditions acquire held-out P3 above the pretrained state at some point, but their peak timing and post-peak retention differ substantially.}\label{tab:S3_dense}\\
\toprule
Method & PT & Peak & Peak ep. & Deploy & Deploy ep. & Gain$_{peak}$ & Gain$_{deploy}$ & Forget$_{post-peak}$ \\
\midrule
\endfirsthead
\toprule
Method & PT & Peak & Peak ep. & Deploy & Deploy ep. & Gain$_{peak}$ & Gain$_{deploy}$ & Forget$_{post-peak}$ \\
\midrule
\endhead
\midrule \multicolumn{9}{r}{\emph{continued on next page}} \\ \midrule
\endfoot
\bottomrule
\endlastfoot
\multicolumn{9}{l}{\textbf{CUB-200 (target)}} \\
\midrule
\noalign{\penalty10000}
N & 34.35 & 35.23 & 4 & 31.12 & 40 & 0.88 & -3.23 & 4.11 \\
\noalign{\penalty10000}
U & 34.45 & 38.87 & 12 & 35.78 & 25 & 4.42 & 1.33 & 3.09 \\
\noalign{\penalty10000}
SSR & 34.35 & 40.42 & 20 & 40.42 & 20 & 6.08 & 6.08 & 0.00 \\
\noalign{\penalty10000}
SSR+ND & 34.35 & \textbf{46.55} & 75 & \textbf{46.31} & 60 & \textbf{12.20} & \textbf{11.96} & 0.24 \\
\midrule
\multicolumn{9}{l}{\textbf{NAB-348 (strict cross-ontology held-out)}} \\
\midrule
\noalign{\penalty10000}
N & 13.85 & 15.59 & 12 & 13.74 & 40 & 1.74 & -0.11 & 1.85 \\
\noalign{\penalty10000}
U & 13.87 & 16.38 & 10 & 15.14 & 25 & 2.51 & 1.27 & 1.24 \\
\noalign{\penalty10000}
SSR & 13.85 & 17.32 & 12 & \textbf{16.90} & 20 & 3.48 & \textbf{3.06} & 0.42 \\
\noalign{\penalty10000}
SSR+ND & 13.85 & \textbf{17.66} & 20 & 14.50 & 60 & \textbf{3.81} & 0.65 & 3.16 \\
\midrule
\end{longtable}
\end{center}

\FloatBarrier
\subsection{Six-backbone acquisition summary}
\label{app:D_summary}
Table~\ref{tab:S4_summary} summarizes peak P3 gain across the six pretrained backbones. The backbones are heterogeneous and are not stochastic replicates; means and medians are descriptive.

\begin{center}
\small
\begin{longtable}{lrrrr}
\caption{Six-backbone descriptive summary of peak P3 gain over pretrained, fine-tuned on ImageNet-1k and evaluated on ImageNet val (target), its same-ontology shifts, and cross-ontology held-out data (CUB-200, NAB-348, FineView Butterfly); PT for each backbone is evaluated directly on the corresponding dataset. Means and medians are descriptive across the six deliberately diverse pretrained backbones, not stochastic-replicate estimates; best values within each distribution are bolded, with ties retained. Peak P3 gains are largest under structured supervision on ImageNet-family evaluations, whereas cross-ontology gains are markedly smaller.}\label{tab:S4_summary}\\
\toprule
Method & Mean Gain$_{peak}$ (pp) & Median Gain$_{peak}$ (pp) & N positive & N $>$0.5pp \\
\midrule
\endfirsthead
\toprule
Method & Mean Gain$_{peak}$ (pp) & Median Gain$_{peak}$ (pp) & N positive & N $>$0.5pp \\
\midrule
\endhead
\midrule \multicolumn{5}{r}{\emph{continued on next page}} \\ \midrule
\endfoot
\bottomrule
\endlastfoot
\multicolumn{5}{l}{\textbf{ImageNet val (target)}} \\
\midrule
\noalign{\penalty10000}
N & 3.63 & 3.14 & \textbf{6/6} & \textbf{6/6} \\
\noalign{\penalty10000}
U & 1.47 & 1.81 & 5/6 & 5/6 \\
\noalign{\penalty10000}
SSR & 7.51 & 6.84 & \textbf{6/6} & \textbf{6/6} \\
\noalign{\penalty10000}
SSR+ND & \textbf{13.23} & \textbf{13.70} & \textbf{6/6} & \textbf{6/6} \\
\midrule
\multicolumn{5}{l}{\textbf{ImageNet-V2 (same-ontology shift)}} \\
\midrule
\noalign{\penalty10000}
N & 3.32 & 2.86 & \textbf{6/6} & \textbf{6/6} \\
\noalign{\penalty10000}
U & 1.55 & 1.74 & \textbf{6/6} & 5/6 \\
\noalign{\penalty10000}
SSR & 6.73 & 5.58 & \textbf{6/6} & \textbf{6/6} \\
\noalign{\penalty10000}
SSR+ND & \textbf{11.52} & \textbf{11.92} & \textbf{6/6} & \textbf{6/6} \\
\midrule
\multicolumn{5}{l}{\textbf{ImageNet-R (same-ontology shift)}} \\
\midrule
\noalign{\penalty10000}
N & 5.11 & 5.87 & \textbf{6/6} & \textbf{6/6} \\
\noalign{\penalty10000}
U & 3.97 & 3.68 & \textbf{6/6} & \textbf{6/6} \\
\noalign{\penalty10000}
SSR & 9.34 & 9.76 & \textbf{6/6} & \textbf{6/6} \\
\noalign{\penalty10000}
SSR+ND & \textbf{13.74} & \textbf{13.82} & \textbf{6/6} & \textbf{6/6} \\
\midrule
\multicolumn{5}{l}{\textbf{ImageNet-Sketch (same-ontology shift)}} \\
\midrule
\noalign{\penalty10000}
N & 3.43 & 2.93 & 5/6 & 5/6 \\
\noalign{\penalty10000}
U & 2.24 & 2.33 & \textbf{6/6} & \textbf{6/6} \\
\noalign{\penalty10000}
SSR & 5.38 & 4.84 & \textbf{6/6} & \textbf{6/6} \\
\noalign{\penalty10000}
SSR+ND & \textbf{8.21} & \textbf{7.89} & \textbf{6/6} & \textbf{6/6} \\
\midrule
\multicolumn{5}{l}{\textbf{CUB-200 (cross-ontology)}} \\
\midrule
\noalign{\penalty10000}
N & 0.51 & 0.35 & 3/6 & 3/6 \\
\noalign{\penalty10000}
U & -1.60 & -0.74 & 1/6 & 1/6 \\
\noalign{\penalty10000}
SSR & 1.45 & 1.69 & \textbf{6/6} & 4/6 \\
\noalign{\penalty10000}
SSR+ND & \textbf{2.26} & \textbf{2.26} & \textbf{6/6} & \textbf{5/6} \\
\midrule
\multicolumn{5}{l}{\textbf{NAB-348 (cross-ontology, strict held-out)}} \\
\midrule
\noalign{\penalty10000}
N & 0.02 & -0.06 & 2/6 & 1/6 \\
\noalign{\penalty10000}
U & -1.41 & -1.42 & 1/6 & 0/6 \\
\noalign{\penalty10000}
SSR & 0.46 & 0.55 & \textbf{6/6} & 3/6 \\
\noalign{\penalty10000}
SSR+ND & \textbf{0.72} & \textbf{0.65} & \textbf{6/6} & \textbf{4/6} \\
\midrule
\multicolumn{5}{l}{\textbf{FineView Butterfly (FVB) (cross-ontology)}} \\
\midrule
\noalign{\penalty10000}
N & 0.33 & 0.31 & 5/6 & 1/6 \\
\noalign{\penalty10000}
U & -0.03 & 0.11 & 3/6 & 0/6 \\
\noalign{\penalty10000}
SSR & \textbf{0.86} & \textbf{0.83} & \textbf{6/6} & \textbf{4/6} \\
\noalign{\penalty10000}
SSR+ND & 0.63 & 0.64 & 5/6 & \textbf{4/6} \\
\midrule
\end{longtable}
\end{center}

\FloatBarrier
\subsection{Trajectory-regime summary}
\label{app:D_regime}
Table~\ref{tab:S5_regime} classifies each cross-ontology cell by whether peak gain was retained, partially reduced, lost, or never clearly acquired; Table~\ref{tab:S6_panel} lists the pretrained backbones.

\begin{center}
\scriptsize\setlength{\tabcolsep}{3.5pt}
\begin{longtable}{llccc}
\caption{All six ImageNet-1k-trained backbones, evaluated cross-ontology on CUB-200, NAB-348 (strict held-out), and FineView Butterfly (FVB); PT is the pretrained checkpoint evaluated directly on the relevant held-out dataset. Each cell reports peak P3 gain over pretrained (pp) and a trajectory regime describing whether that gain was retained, partially reduced, or fully lost by deployment, or never clearly acquired. Poor deployed transfer can arise either because a capability was never clearly acquired or because it was acquired earlier and subsequently reduced.}\label{tab:S5_regime}\\
\toprule
Backbone & Method & CUB-200 & NAB-348 & FVB \\
\midrule
\endfirsthead
\toprule
Backbone & Method & CUB-200 & NAB-348 & FVB \\
\midrule
\endhead
\midrule \multicolumn{5}{r}{\emph{continued on next page}} \\ \midrule
\endfoot
\bottomrule
\endlastfoot
DFN5B H/14-378 & N & -0.72, no acquisition / degraded & -0.41, no acquisition / degraded & -0.10, no acquisition / degraded \\
\noalign{\penalty10000}
 & U & 1.12, acquire$\to$lose & 0.47, acquire$\to$lose & 0.23, acquire$\to$lose \\
\noalign{\penalty10000}
 & SSR & 0.40, acquire$\to$lose & 0.87, acquire$\to$lose & 0.33, acquire$\to$lose \\
\noalign{\penalty10000}
 & SSR+ND & 0.52, acquire$\to$lose & 1.23, acquire$\to$lose & -0.14, no acquisition / degraded \\
\midrule
DFN2B L/14 & N & 0.00, no acquisition / degraded & -0.01, no acquisition / degraded & 0.17, acquire$\to$retain \\
\noalign{\penalty10000}
 & U & -0.31, no acquisition / degraded & -0.20, no acquisition / degraded & -0.23, no acquisition / degraded \\
\noalign{\penalty10000}
 & SSR & 0.19, acquire$\to$lose & 0.12, acquire$\to$lose & 1.04, acquire$\to$retain \\
\noalign{\penalty10000}
 & SSR+ND & 0.31, acquire$\to$lose & 0.41, acquire$\to$lose & 0.77, acquire$\to$retain \\
\midrule
MetaCLIP H/14 & N & 2.35, acquire$\to$lose & 0.72, acquire$\to$lose & 0.25, acquire$\to$lose \\
\noalign{\penalty10000}
 & U & -5.06, no acquisition / degraded & -3.11, no acquisition / degraded & -1.04, no acquisition / degraded \\
\noalign{\penalty10000}
 & SSR & 2.48, acquire$\to$lose & 0.47, acquire$\to$lose & 0.62, acquire$\to$lose \\
\noalign{\penalty10000}
 & SSR+ND & 3.06, acquire$\to$partial decline & 0.58, acquire$\to$lose & 0.52, acquire$\to$retain \\
\midrule
MetaCLIP L/14 & N & 0.93, acquire$\to$partial decline & -0.25, no acquisition / degraded & 0.36, acquire$\to$retain \\
\noalign{\penalty10000}
 & U & -4.12, no acquisition / degraded & -2.82, no acquisition / degraded & -0.02, no acquisition / degraded \\
\noalign{\penalty10000}
 & SSR & 1.60, acquire$\to$lose & 0.02, no acquisition / degraded & 0.40, acquire$\to$lose \\
\noalign{\penalty10000}
 & SSR+ND & 4.38, acquire$\to$partial decline & 0.36, acquire$\to$lose & 0.85, acquire$\to$partial decline \\
\midrule
OpenAI L/14 & N & 0.71, acquire$\to$lose & 0.16, acquire$\to$lose & 0.38, acquire$\to$lose \\
\noalign{\penalty10000}
 & U & -0.07, no clear acquisition & -1.96, no acquisition / degraded & 0.36, acquire$\to$lose \\
\noalign{\penalty10000}
 & SSR & 2.26, acquire$\to$partial decline & 0.64, acquire$\to$lose & 1.50, acquire$\to$lose \\
\noalign{\penalty10000}
 & SSR+ND & 3.83, acquire$\to$partial decline & 0.99, acquire$\to$partial decline & 0.48, acquire$\to$lose \\
\midrule
OpenAI B/32 & N & -0.22, no acquisition / degraded & -0.12, no acquisition / degraded & 0.90, acquire$\to$partial decline \\
\noalign{\penalty10000}
 & U & -1.17, no acquisition / degraded & -0.87, no acquisition / degraded & 0.50, acquire$\to$lose \\
\noalign{\penalty10000}
 & SSR & 1.78, acquire$\to$partial decline & 0.66, acquire$\to$lose & 1.29, acquire$\to$partial decline \\
\noalign{\penalty10000}
 & SSR+ND & 1.47, acquire$\to$partial decline & 0.73, acquire$\to$retain & 1.31, acquire$\to$partial decline \\
\midrule
\end{longtable}
\end{center}

\begin{center}
\small
\begin{longtable}{llrlr}
\caption{The six pretrained backbones used throughout the ImageNet-1k experiments (Figure~\ref{fig:acquisition}b), listed before any downstream fine-tuning. The final column is pretrained P1 (class-name retrieval R@1) evaluated on ImageNet val; no downstream trajectory result is represented in this table.}\label{tab:S6_panel}\\
\toprule
Backbone & Family & Params & Pretraining source & ImageNet-val P1 (\%) \\
\midrule
\endfirsthead
\toprule
Backbone & Family & Params & Pretraining source & ImageNet-val P1 (\%) \\
\midrule
\endhead
\midrule \multicolumn{5}{r}{\emph{continued on next page}} \\ \midrule
\endfoot
\bottomrule
\endlastfoot
DFN5B H/14-378 & DFN & 986M & DFN5B (5B pool) & 83.9 \\
DFN2B L/14 & DFN & 427M & DFN-2B (2B pool) & 81.8 \\
MetaCLIP H/14 & MetaCLIP & 986M & MetaCLIP (fullcc) & 80.3 \\
MetaCLIP L/14 & MetaCLIP & 427M & MetaCLIP (fullcc) & 79.0 \\
OpenAI L/14 & OpenAI & 427M & OpenAI (WIT-400M) & 73.3 \\
OpenAI B/32 & OpenAI & 151M & OpenAI (WIT-400M) & 63.1 \\
\end{longtable}
\end{center}

\FloatBarrier
\subsection{P1 context}
\label{app:D_p1}
Table~\ref{tab:S7_P1_context} reports deployed class-name retrieval (P1) next to P3 peak, to show that large differences in P3 acquisition occur while deployed P1 is comparatively similar.

\begin{center}
\footnotesize
\begin{longtable}{lrrrrr}
\caption{All six backbones, fine-tuned on ImageNet-1k and evaluated on ImageNet val (target) and its same-ontology shifts; PT is the pretrained checkpoint evaluated directly on each listed dataset. P1 deploy is class-name retrieval at each run's own P1-selected deployment checkpoint, and P3 peak is the maximum P3 over the trajectory; within each backbone, the highest deployed P1 and peak P3 are bolded independently. Large differences in P3 acquisition can occur while deployed P1 remains comparatively similar.}\label{tab:S7_P1_context}\\
\toprule
Method & P1 PT & P1 deploy & P3 PT & P3 peak & Gain$_{peak}$(P3) \\
\midrule
\endfirsthead
\toprule
Method & P1 PT & P1 deploy & P3 PT & P3 peak & Gain$_{peak}$(P3) \\
\midrule
\endhead
\midrule \multicolumn{6}{r}{\emph{continued on next page}} \\ \midrule
\endfoot
\bottomrule
\endlastfoot
\multicolumn{6}{l}{\textbf{ImageNet val (target)}} \\
\midrule
N & 83.90 & \textbf{87.82} & 33.12 & 33.78 & 0.66 \\
\noalign{\penalty10000}
U & 83.90 & 87.66 & 33.12 & 35.38 & 2.26 \\
\noalign{\penalty10000}
SSR & 83.90 & 87.00 & 33.12 & 45.34 & 12.22 \\
\noalign{\penalty10000}
SSR+ND & 83.90 & 86.96 & 33.12 & \textbf{48.12} & 15.00 \\
\noalign{\penalty10000}
\addlinespace[1pt]
\noalign{\penalty10000}
\multicolumn{6}{l}{\scriptsize DFN5B H/14-378} \\
\midrule
N & 81.84 & \textbf{85.98} & 28.60 & 30.06 & 1.46 \\
\noalign{\penalty10000}
U & 81.84 & 85.66 & 28.60 & 29.44 & 0.84 \\
\noalign{\penalty10000}
SSR & 81.84 & 85.00 & 28.60 & 33.90 & 5.30 \\
\noalign{\penalty10000}
SSR+ND & 81.84 & 84.86 & 28.60 & \textbf{36.70} & 8.10 \\
\noalign{\penalty10000}
\addlinespace[1pt]
\noalign{\penalty10000}
\multicolumn{6}{l}{\scriptsize DFN2B L/14} \\
\midrule
N & 80.26 & 86.32 & 35.94 & 44.08 & 8.14 \\
\noalign{\penalty10000}
U & 80.26 & \textbf{86.74} & 35.94 & 38.00 & 2.06 \\
\noalign{\penalty10000}
SSR & 80.26 & 85.84 & 35.94 & 43.04 & 7.10 \\
\noalign{\penalty10000}
SSR+ND & 80.26 & 85.42 & 35.94 & \textbf{50.60} & 14.66 \\
\noalign{\penalty10000}
\addlinespace[1pt]
\noalign{\penalty10000}
\multicolumn{6}{l}{\scriptsize MetaCLIP H/14} \\
\midrule
N & 79.04 & 85.66 & 31.12 & 35.86 & 4.74 \\
\noalign{\penalty10000}
U & 79.04 & \textbf{85.84} & 31.12 & 32.68 & 1.56 \\
\noalign{\penalty10000}
SSR & 79.04 & 84.86 & 31.12 & 37.70 & 6.58 \\
\noalign{\penalty10000}
SSR+ND & 79.04 & 84.54 & 31.12 & \textbf{43.86} & 12.74 \\
\noalign{\penalty10000}
\addlinespace[1pt]
\noalign{\penalty10000}
\multicolumn{6}{l}{\scriptsize MetaCLIP L/14} \\
\midrule
N & 73.32 & \textbf{84.98} & 25.48 & 30.72 & 5.24 \\
\noalign{\penalty10000}
U & 73.32 & 84.66 & 25.48 & 27.76 & 2.28 \\
\noalign{\penalty10000}
SSR & 73.32 & 83.62 & 25.48 & 35.14 & 9.66 \\
\noalign{\penalty10000}
SSR+ND & 73.32 & 83.30 & 25.48 & \textbf{45.36} & 19.88 \\
\noalign{\penalty10000}
\addlinespace[1pt]
\noalign{\penalty10000}
\multicolumn{6}{l}{\scriptsize OpenAI L/14} \\
\midrule
N & 63.08 & \textbf{75.42} & 20.44 & 21.98 & 1.54 \\
\noalign{\penalty10000}
U & 63.08 & 75.16 & 20.44 & 20.24 & -0.20 \\
\noalign{\penalty10000}
SSR & 63.08 & 71.26 & 20.44 & 24.62 & 4.18 \\
\noalign{\penalty10000}
SSR+ND & 63.08 & 72.00 & 20.44 & \textbf{29.42} & 8.98 \\
\noalign{\penalty10000}
\addlinespace[1pt]
\noalign{\penalty10000}
\multicolumn{6}{l}{\scriptsize OpenAI B/32} \\
\midrule
\multicolumn{6}{l}{\textbf{ImageNet-V2 (same-ontology shift)}} \\
\midrule
N & 77.98 & 81.40 & 29.29 & 30.76 & 1.47 \\
\noalign{\penalty10000}
U & 77.98 & \textbf{81.44} & 29.29 & 30.90 & 1.61 \\
\noalign{\penalty10000}
SSR & 77.98 & 80.24 & 29.29 & 40.68 & 11.39 \\
\noalign{\penalty10000}
SSR+ND & 77.98 & 80.51 & 29.29 & \textbf{42.83} & 13.54 \\
\noalign{\penalty10000}
\addlinespace[1pt]
\noalign{\penalty10000}
\multicolumn{6}{l}{\scriptsize DFN5B H/14-378} \\
\midrule
N & 75.10 & 78.76 & 24.51 & 25.85 & 1.34 \\
\noalign{\penalty10000}
U & 75.10 & \textbf{79.00} & 24.51 & 25.43 & 0.92 \\
\noalign{\penalty10000}
SSR & 75.10 & 78.50 & 24.51 & 29.81 & 5.30 \\
\noalign{\penalty10000}
SSR+ND & 75.10 & 78.07 & 24.51 & \textbf{32.06} & 7.55 \\
\noalign{\penalty10000}
\addlinespace[1pt]
\noalign{\penalty10000}
\multicolumn{6}{l}{\scriptsize DFN2B L/14} \\
\midrule
N & 73.06 & 79.10 & 31.66 & 38.58 & 6.91 \\
\noalign{\penalty10000}
U & 73.06 & \textbf{79.11} & 31.66 & 33.53 & 1.86 \\
\noalign{\penalty10000}
SSR & 73.06 & 78.44 & 31.66 & 37.30 & 5.64 \\
\noalign{\penalty10000}
SSR+ND & 73.06 & 78.65 & 31.66 & \textbf{44.96} & 13.30 \\
\noalign{\penalty10000}
\addlinespace[1pt]
\noalign{\penalty10000}
\multicolumn{6}{l}{\scriptsize MetaCLIP H/14} \\
\midrule
N & 72.20 & 78.04 & 27.39 & 31.64 & 4.25 \\
\noalign{\penalty10000}
U & 72.20 & \textbf{78.67} & 27.39 & 30.05 & 2.65 \\
\noalign{\penalty10000}
SSR & 72.20 & 77.31 & 27.39 & 32.92 & 5.52 \\
\noalign{\penalty10000}
SSR+ND & 72.20 & 77.69 & 27.39 & \textbf{37.94} & 10.54 \\
\noalign{\penalty10000}
\addlinespace[1pt]
\noalign{\penalty10000}
\multicolumn{6}{l}{\scriptsize MetaCLIP L/14} \\
\midrule
N & 67.31 & \textbf{77.55} & 22.73 & 27.46 & 4.74 \\
\noalign{\penalty10000}
U & 67.31 & 77.41 & 22.73 & 24.62 & 1.89 \\
\noalign{\penalty10000}
SSR & 67.31 & 76.32 & 22.73 & 31.47 & 8.75 \\
\noalign{\penalty10000}
SSR+ND & 67.31 & 76.22 & 22.73 & \textbf{39.68} & 16.95 \\
\noalign{\penalty10000}
\addlinespace[1pt]
\noalign{\penalty10000}
\multicolumn{6}{l}{\scriptsize OpenAI L/14} \\
\midrule
N & 54.73 & \textbf{64.70} & 17.16 & 18.38 & 1.21 \\
\noalign{\penalty10000}
U & 54.73 & 64.43 & 17.16 & 17.54 & 0.37 \\
\noalign{\penalty10000}
SSR & 54.73 & 61.14 & 17.16 & 20.93 & 3.77 \\
\noalign{\penalty10000}
SSR+ND & 54.73 & 61.16 & 17.16 & \textbf{24.42} & 7.25 \\
\noalign{\penalty10000}
\addlinespace[1pt]
\noalign{\penalty10000}
\multicolumn{6}{l}{\scriptsize OpenAI B/32} \\
\midrule
\multicolumn{6}{l}{\textbf{ImageNet-R (same-ontology shift)}} \\
\midrule
N & 91.50 & 89.72 & 35.40 & 43.52 & 8.12 \\
\noalign{\penalty10000}
U & 91.50 & 89.82 & 35.40 & 38.52 & 3.12 \\
\noalign{\penalty10000}
SSR & 91.50 & 89.73 & 35.40 & 51.33 & 15.93 \\
\noalign{\penalty10000}
SSR+ND & 91.50 & \textbf{90.20} & 35.40 & \textbf{53.60} & 18.20 \\
\noalign{\penalty10000}
\addlinespace[1pt]
\noalign{\penalty10000}
\multicolumn{6}{l}{\scriptsize DFN5B H/14-378} \\
\midrule
N & 88.77 & 88.42 & 30.62 & 31.22 & 0.60 \\
\noalign{\penalty10000}
U & 88.77 & \textbf{88.95} & 30.62 & 32.90 & 2.28 \\
\noalign{\penalty10000}
SSR & 88.77 & 88.12 & 30.62 & 35.45 & 4.83 \\
\noalign{\penalty10000}
SSR+ND & 88.77 & 88.02 & 30.62 & \textbf{38.50} & 7.88 \\
\noalign{\penalty10000}
\addlinespace[1pt]
\noalign{\penalty10000}
\multicolumn{6}{l}{\scriptsize DFN2B L/14} \\
\midrule
N & 90.33 & 90.58 & 42.87 & 49.25 & 6.38 \\
\noalign{\penalty10000}
U & 90.33 & \textbf{91.30} & 42.87 & 47.12 & 4.25 \\
\noalign{\penalty10000}
SSR & 90.33 & 90.13 & 42.87 & 52.45 & 9.58 \\
\noalign{\penalty10000}
SSR+ND & 90.33 & 90.87 & 42.87 & \textbf{58.20} & 15.33 \\
\noalign{\penalty10000}
\addlinespace[1pt]
\noalign{\penalty10000}
\multicolumn{6}{l}{\scriptsize MetaCLIP H/14} \\
\midrule
N & 88.93 & 89.03 & 36.35 & 41.70 & 5.35 \\
\noalign{\penalty10000}
U & 88.93 & \textbf{89.60} & 36.35 & 41.92 & 5.57 \\
\noalign{\penalty10000}
SSR & 88.93 & 87.88 & 36.35 & 46.28 & 9.93 \\
\noalign{\penalty10000}
SSR+ND & 88.93 & 88.88 & 36.35 & \textbf{48.65} & 12.30 \\
\noalign{\penalty10000}
\addlinespace[1pt]
\noalign{\penalty10000}
\multicolumn{6}{l}{\scriptsize MetaCLIP L/14} \\
\midrule
N & 83.47 & 86.00 & 31.45 & 37.92 & 6.47 \\
\noalign{\penalty10000}
U & 83.47 & \textbf{86.77} & 31.45 & 37.28 & 5.83 \\
\noalign{\penalty10000}
SSR & 83.47 & 85.45 & 31.45 & 42.33 & 10.88 \\
\noalign{\penalty10000}
SSR+ND & 83.47 & 86.17 & 31.45 & \textbf{53.08} & 21.63 \\
\noalign{\penalty10000}
\addlinespace[1pt]
\noalign{\penalty10000}
\multicolumn{6}{l}{\scriptsize OpenAI L/14} \\
\midrule
N & 65.03 & 64.77 & 22.35 & 26.08 & 3.73 \\
\noalign{\penalty10000}
U & 65.03 & \textbf{66.07} & 22.35 & 25.10 & 2.75 \\
\noalign{\penalty10000}
SSR & 65.03 & 62.60 & 22.35 & 27.25 & 4.90 \\
\noalign{\penalty10000}
SSR+ND & 65.03 & 63.53 & 22.35 & \textbf{29.45} & 7.10 \\
\noalign{\penalty10000}
\addlinespace[1pt]
\noalign{\penalty10000}
\multicolumn{6}{l}{\scriptsize OpenAI B/32} \\
\midrule
\multicolumn{6}{l}{\textbf{ImageNet-Sketch (same-ontology shift)}} \\
\midrule
N & 72.45 & 71.68 & 28.35 & 30.67 & 2.32 \\
\noalign{\penalty10000}
U & 72.45 & 70.54 & 28.35 & 30.55 & 2.20 \\
\noalign{\penalty10000}
SSR & 72.45 & 70.84 & 28.35 & 37.43 & 9.08 \\
\noalign{\penalty10000}
SSR+ND & 72.45 & \textbf{71.74} & 28.35 & \textbf{39.43} & 11.08 \\
\noalign{\penalty10000}
\addlinespace[1pt]
\noalign{\penalty10000}
\multicolumn{6}{l}{\scriptsize DFN5B H/14-378} \\
\midrule
N & 68.64 & 68.43 & 23.85 & 23.79 & -0.06 \\
\noalign{\penalty10000}
U & 68.64 & \textbf{68.78} & 23.85 & 24.56 & 0.71 \\
\noalign{\penalty10000}
SSR & 68.64 & 68.64 & 23.85 & 27.70 & 3.85 \\
\noalign{\penalty10000}
SSR+ND & 68.64 & 68.68 & 23.85 & \textbf{29.25} & 5.40 \\
\noalign{\penalty10000}
\addlinespace[1pt]
\noalign{\penalty10000}
\multicolumn{6}{l}{\scriptsize DFN2B L/14} \\
\midrule
N & 69.53 & 69.52 & 29.31 & 36.24 & 6.93 \\
\noalign{\penalty10000}
U & 69.53 & \textbf{69.61} & 29.31 & 32.42 & 3.11 \\
\noalign{\penalty10000}
SSR & 69.53 & 69.11 & 29.31 & 34.52 & 5.21 \\
\noalign{\penalty10000}
SSR+ND & 69.53 & 69.44 & 29.31 & \textbf{38.80} & 9.49 \\
\noalign{\penalty10000}
\addlinespace[1pt]
\noalign{\penalty10000}
\multicolumn{6}{l}{\scriptsize MetaCLIP H/14} \\
\midrule
N & 67.51 & 67.50 & 25.44 & 28.92 & 3.47 \\
\noalign{\penalty10000}
U & 67.51 & \textbf{67.56} & 25.44 & 29.06 & 3.61 \\
\noalign{\penalty10000}
SSR & 67.51 & 66.47 & 25.44 & 29.92 & 4.48 \\
\noalign{\penalty10000}
SSR+ND & 67.51 & 66.99 & 25.44 & \textbf{31.74} & 6.30 \\
\noalign{\penalty10000}
\addlinespace[1pt]
\noalign{\penalty10000}
\multicolumn{6}{l}{\scriptsize MetaCLIP L/14} \\
\midrule
N & 57.34 & \textbf{61.36} & 19.41 & 24.96 & 5.55 \\
\noalign{\penalty10000}
U & 57.34 & 61.16 & 19.41 & 21.86 & 2.46 \\
\noalign{\penalty10000}
SSR & 57.34 & 60.51 & 19.41 & 26.72 & 7.32 \\
\noalign{\penalty10000}
SSR+ND & 57.34 & 60.47 & 19.41 & \textbf{31.88} & 12.47 \\
\noalign{\penalty10000}
\addlinespace[1pt]
\noalign{\penalty10000}
\multicolumn{6}{l}{\scriptsize OpenAI L/14} \\
\midrule
N & 41.18 & 42.57 & 12.70 & 15.08 & 2.38 \\
\noalign{\penalty10000}
U & 41.18 & \textbf{42.65} & 12.70 & 14.07 & 1.37 \\
\noalign{\penalty10000}
SSR & 41.18 & 39.81 & 12.70 & 15.07 & 2.37 \\
\noalign{\penalty10000}
SSR+ND & 41.18 & 40.63 & 12.70 & \textbf{17.25} & 4.56 \\
\noalign{\penalty10000}
\addlinespace[1pt]
\noalign{\penalty10000}
\multicolumn{6}{l}{\scriptsize OpenAI B/32} \\
\midrule
\end{longtable}
\end{center}

%% file: supplement/tables_sec5/generated_section5_supplement.tex
% AUTO-GENERATED by scripts/build_section5_supplement.py -- do not hand-edit.

\subsection{Deployment P1/P2/P3}
\label{app:E_deploy}
Table~\ref{tab:S8_deployment} gives P1, P2, and P3 at each run's own P1-selected deployment checkpoint on ImageNet val for all six backbones.

\begin{table}[t]
\centering
\caption{All six backbones, fine-tuned on ImageNet-1k and evaluated on ImageNet val (target); PT is the pretrained checkpoint evaluated on ImageNet val. P1/P2/P3 (R@1 \%) are all measured at each run's own P1-selected \emph{deployment} checkpoint, not peak P3 (underlies Table~\ref{tab:deploy_profile}). U produces the strongest name-containing P2 behavior, whereas structured SSR and especially SSR+ND yield substantially stronger name-free P3 despite broadly similar (not identical) P1.}
\label{tab:S8_deployment}
\begin{minipage}[t]{0.315\textwidth}\centering\footnotesize
\textbf{DFN5B H/14-378}\\[2pt]
\setlength{\tabcolsep}{3.2pt}\begin{tabular}{lrrr}\toprule
Method & P1 & P2 & P3 \\\midrule
PT & 83.90 & 79.48 & 33.12 \\
N & 87.82 & 86.54 & 33.78 \\
U & 87.66 & 87.90 & 24.80 \\
SSR & 87.00 & 86.10 & 45.12 \\
SSR+ND & 86.96 & 86.12 & 48.00 \\
\bottomrule\end{tabular}\end{minipage} \hfill
\begin{minipage}[t]{0.315\textwidth}\centering\footnotesize
\textbf{DFN2B L/14}\\[2pt]
\setlength{\tabcolsep}{3.2pt}\begin{tabular}{lrrr}\toprule
Method & P1 & P2 & P3 \\\midrule
PT & 81.84 & 76.32 & 28.60 \\
N & 85.98 & 79.66 & 30.06 \\
U & 85.66 & 85.98 & 28.44 \\
SSR & 85.00 & 81.86 & 33.56 \\
SSR+ND & 84.86 & 81.98 & 36.70 \\
\bottomrule\end{tabular}\end{minipage} \hfill
\begin{minipage}[t]{0.315\textwidth}\centering\footnotesize
\textbf{MetaCLIP H/14}\\[2pt]
\setlength{\tabcolsep}{3.2pt}\begin{tabular}{lrrr}\toprule
Method & P1 & P2 & P3 \\\midrule
PT & 80.26 & 75.56 & 35.94 \\
N & 86.32 & 83.24 & 44.08 \\
U & 86.74 & 86.64 & 32.44 \\
SSR & 85.84 & 83.96 & 43.04 \\
SSR+ND & 85.42 & 84.38 & 50.60 \\
\bottomrule\end{tabular}\end{minipage}
\\[8pt]
\begin{minipage}[t]{0.315\textwidth}\centering\footnotesize
\textbf{MetaCLIP L/14}\\[2pt]
\setlength{\tabcolsep}{3.2pt}\begin{tabular}{lrrr}\toprule
Method & P1 & P2 & P3 \\\midrule
PT & 79.04 & 72.70 & 31.12 \\
N & 85.66 & 80.84 & 35.48 \\
U & 85.84 & 86.04 & 29.36 \\
SSR & 84.86 & 81.88 & 36.04 \\
SSR+ND & 84.54 & 82.26 & 43.86 \\
\bottomrule\end{tabular}\end{minipage} \hfill
\begin{minipage}[t]{0.315\textwidth}\centering\footnotesize
\textbf{OpenAI L/14}\\[2pt]
\setlength{\tabcolsep}{3.2pt}\begin{tabular}{lrrr}\toprule
Method & P1 & P2 & P3 \\\midrule
PT & 73.32 & 69.70 & 25.48 \\
N & 84.98 & 81.36 & 30.72 \\
U & 84.66 & 85.18 & 27.76 \\
SSR & 83.62 & 82.14 & 35.14 \\
SSR+ND & 83.30 & 82.48 & 45.36 \\
\bottomrule\end{tabular}\end{minipage} \hfill
\begin{minipage}[t]{0.315\textwidth}\centering\footnotesize
\textbf{OpenAI B/32}\\[2pt]
\setlength{\tabcolsep}{3.2pt}\begin{tabular}{lrrr}\toprule
Method & P1 & P2 & P3 \\\midrule
PT & 63.08 & 58.32 & 20.44 \\
N & 75.42 & 68.80 & 21.68 \\
U & 75.16 & 75.82 & 19.92 \\
SSR & 71.26 & 69.24 & 24.62 \\
SSR+ND & 72.00 & 70.48 & 29.42 \\
\bottomrule\end{tabular}\end{minipage}
\end{table}
\FloatBarrier

\FloatBarrier
\subsection{P2 name-swap and attribute-swap diagnostic}
\label{app:E_p2}
Table~\ref{tab:S9_name_dependency} reports ordinary, name-swapped, and attribute-swapped P2 with class-level bootstrap intervals (\S\ref{app:B_bootstrap}). Ordinary P2 remains overwhelmingly class-name anchored under every condition, including PT; SSR does not remove this name reliance.

\begin{center}
\scriptsize
\begin{longtable}{lrrrrrr}
\caption{DFN5B H/14-378 fine-tuned on CUB-200, evaluated on CUB-200 (target) and NAB-348 (strict cross-ontology held-out) under five conditions (PT, N, U, SSR, SSR+ND); PT is the pretrained checkpoint evaluated directly on each listed dataset, and all fine-tuned methods are at their own P1-selected deployment checkpoint. Ordinary, name-swapped, and attribute-swapped P2 (R@1 \%) are shown with $\Delta_{name}$/$\Delta_{attr}$, the pp change from ordinary, with class-level bootstrap 95\% CIs (1000 replicates). Name-swap nearly eliminates P2 retrieval, whereas attribute-swap preserves most performance, showing that ordinary name+attribute retrieval remains strongly class-name anchored.}\label{tab:S9_name_dependency}\\
\toprule
Method & P2 ord. & P2 name-sw. & P2 attr-sw. & $\Delta_{name}$ [95\% CI] & $\Delta_{attr}$ [95\% CI] & P3 \\
\midrule
\endfirsthead
\toprule
Method & P2 ord. & P2 name-sw. & P2 attr-sw. & $\Delta_{name}$ [95\% CI] & $\Delta_{attr}$ [95\% CI] & P3 \\
\midrule
\endhead
\midrule \multicolumn{7}{r}{\emph{continued on next page}} \\ \midrule
\endfoot
\bottomrule
\endlastfoot
\multicolumn{7}{l}{\textbf{CUB-200 (target)}} \\
\midrule
PT & 86.31 & 0.10 & 76.54 & -86.2 [-88.6, -83.4] & -9.8 [-13.2, -6.6] & 34.36 \\
\noalign{\penalty10000}
N & 90.68 & 0.07 & 89.68 & -90.6 [-92.1, -88.8] & -1.0 [-1.9, -0.2] & 31.10 \\
\noalign{\penalty10000}
U & 91.35 & 0.05 & 90.63 & -91.3 [-92.8, -89.7] & -0.7 [-1.5, -0.1] & 35.78 \\
\noalign{\penalty10000}
SSR & 89.44 & 0.14 & 85.33 & -89.3 [-91.3, -86.9] & -4.1 [-6.6, -2.0] & 40.40 \\
\noalign{\penalty10000}
SSR+ND & 90.04 & 0.22 & 86.92 & -89.8 [-91.5, -87.8] & -3.1 [-4.7, -1.7] & 46.31 \\
\midrule
\multicolumn{7}{l}{\textbf{NAB-348 (strict held-out)}} \\
\midrule
PT & 82.78 & 0.20 & 74.67 & -82.6 [-84.6, -80.3] & -8.1 [-10.4, -5.9] & 13.86 \\
\noalign{\penalty10000}
N & 73.14 & 0.40 & 64.41 & -72.7 [-75.8, -69.8] & -8.7 [-11.6, -6.0] & 13.76 \\
\noalign{\penalty10000}
U & 78.09 & 0.11 & 73.28 & -78.0 [-80.6, -75.2] & -4.8 [-6.6, -3.1] & 15.14 \\
\noalign{\penalty10000}
SSR & 78.98 & 0.33 & 68.39 & -78.6 [-81.0, -76.0] & -10.6 [-13.5, -7.8] & 16.82 \\
\noalign{\penalty10000}
SSR+ND & 73.00 & 0.40 & 63.31 & -72.6 [-75.3, -69.6] & -9.7 [-12.6, -6.7] & 14.54 \\
\midrule
\end{longtable}
\end{center}

\FloatBarrier
\subsection{SSR versus SSR+ND target P3 acquisition}
\label{app:E_snd}
Table~\ref{tab:S10_s_vs_snd_full} lists all 18 backbone$\times$training-domain pairs.

\begin{center}
\scriptsize
\begin{longtable}{llrrrrrr}
\caption{All six backbones, fine-tuned separately on ImageNet-1k, CUB-200, and NABirds (18 backbone$\times$training-domain pairs), each evaluated on its own target training domain (ImageNet-trained $\to$ ImageNet val, CUB-trained $\to$ CUB-200, NAB-trained $\to$ NAB-555) -- this is a target-domain comparison, not a held-out-transfer table. Gain$_{peak}$ is target P3 peak minus target PT P3 (PT evaluated directly on the target dataset), Gain$_{deploy}$ uses each method's own P1-selected deployment checkpoint, and Forget is their difference (underlies Figure~\ref{fig:supervision}b; all values pp). SSR+ND exceeds SSR in peak target P3 acquisition in all 18 tested backbone $\times$ training-domain pairs.}\label{tab:S10_s_vs_snd_full}\\
\toprule
 & & \multicolumn{3}{c}{SSR} & \multicolumn{3}{c}{SSR+ND} \\
\cmidrule(lr){3-5}\cmidrule(lr){6-8}
Backbone & Train & Gain$_{peak}$ & Gain$_{deploy}$ & Forget & Gain$_{peak}$ & Gain$_{deploy}$ & Forget \\
\midrule
\endfirsthead
\toprule
 & & \multicolumn{3}{c}{SSR} & \multicolumn{3}{c}{SSR+ND} \\
\cmidrule(lr){3-5}\cmidrule(lr){6-8}
Backbone & Train & Gain$_{peak}$ & Gain$_{deploy}$ & Forget & Gain$_{peak}$ & Gain$_{deploy}$ & Forget \\
\midrule
\endhead
\midrule \multicolumn{8}{r}{\emph{continued on next page}} \\ \midrule
\endfoot
\bottomrule
\endlastfoot
DFN5B H/14-378 & ImageNet-trained & 12.22 & 12.00 & 0.22 & \textbf{15.00} & 14.88 & 0.12 \\
\noalign{\penalty10000}
DFN5B H/14-378 & CUB-trained & 6.08 & 6.08 & 0.00 & \textbf{12.20} & 11.96 & 0.24 \\
\noalign{\penalty10000}
DFN5B H/14-378 & NAB-trained & 4.09 & -0.23 & 4.32 & \textbf{5.65} & 1.02 & 4.62 \\
\midrule
DFN2B L/14 & ImageNet-trained & 5.30 & 4.96 & 0.34 & \textbf{8.10} & 8.10 & 0.00 \\
\noalign{\penalty10000}
DFN2B L/14 & CUB-trained & 3.78 & 3.69 & 0.09 & \textbf{8.99} & 8.99 & 0.00 \\
\noalign{\penalty10000}
DFN2B L/14 & NAB-trained & 1.01 & 0.58 & 0.42 & \textbf{1.56} & 1.54 & 0.03 \\
\midrule
MetaCLIP H/14 & ImageNet-trained & 7.10 & 7.10 & 0.00 & \textbf{14.66} & 14.38 & 0.28 \\
\noalign{\penalty10000}
MetaCLIP H/14 & CUB-trained & 5.70 & 5.63 & 0.07 & \textbf{13.77} & 13.77 & 0.00 \\
\noalign{\penalty10000}
MetaCLIP H/14 & NAB-trained & 3.09 & 2.78 & 0.31 & \textbf{6.19} & 6.13 & 0.06 \\
\midrule
MetaCLIP L/14 & ImageNet-trained & 6.58 & 4.92 & 1.66 & \textbf{12.74} & 12.74 & 0.00 \\
\noalign{\penalty10000}
MetaCLIP L/14 & CUB-trained & 6.39 & 6.26 & 0.12 & \textbf{18.43} & 18.23 & 0.21 \\
\noalign{\penalty10000}
MetaCLIP L/14 & NAB-trained & 3.51 & 3.36 & 0.15 & \textbf{6.92} & 6.92 & 0.00 \\
\midrule
OpenAI L/14 & ImageNet-trained & 9.66 & 9.66 & 0.00 & \textbf{19.88} & 19.88 & 0.00 \\
\noalign{\penalty10000}
OpenAI L/14 & CUB-trained & 9.91 & 9.68 & 0.23 & \textbf{29.65} & 29.48 & 0.17 \\
\noalign{\penalty10000}
OpenAI L/14 & NAB-trained & 2.57 & 1.36 & 1.22 & \textbf{14.18} & 14.05 & 0.13 \\
\midrule
OpenAI B/32 & ImageNet-trained & 4.18 & 3.98 & 0.20 & \textbf{8.98} & 8.98 & 0.00 \\
\noalign{\penalty10000}
OpenAI B/32 & CUB-trained & 6.80 & 6.73 & 0.07 & \textbf{11.86} & 11.82 & 0.04 \\
\noalign{\penalty10000}
OpenAI B/32 & NAB-trained & 2.38 & 2.23 & 0.15 & \textbf{3.80} & 3.80 & 0.00 \\
\midrule
\end{longtable}
\end{center}

\FloatBarrier
\subsection{Pairwise CAPP target results}
\label{app:E_capp_pair}
CAPP is defined in \S\ref{app:B_capp}.

\begin{center}
\small
\begin{longtable}{llrrrrr}
\caption{All six backbones (DFN/MetaCLIP backbones plus the additional OpenAI L/14 and B/32 backbones, marked below) fine-tuned separately on CUB-200 and on NABirds, each evaluated on its own target training ontology (CUB-trained $\to$ CUB-200, NAB-trained $\to$ NAB-555); PT is the pretrained checkpoint evaluated directly on the target dataset. Pairwise class-slot-macro accuracy at each method's own P1-selected deployment checkpoint; multi-way top-1 in Table~\ref{tab:S11_multi}. Structured supervision consistently improves controlled local attribute discrimination on the CUB target ontology, whereas NAB-target effects are more backbone-dependent.}\label{tab:S11_target_pairwise}\\
\toprule
Backbone & Train domain & PT & N & U & SSR & SSR+ND \\
\midrule
\endfirsthead
\toprule
Backbone & Train domain & PT & N & U & SSR & SSR+ND \\
\midrule
\endhead
\midrule \multicolumn{7}{r}{\emph{continued on next page}} \\ \midrule
\endfoot
\bottomrule
\endlastfoot
DFN5B H/14-378 & CUB-200 & 72.99 & 73.90 & 74.85 & 76.30 & 76.76 \\
\noalign{\penalty10000}
DFN5B H/14-378 & NAB-555 & 65.92 & 67.56 & 67.52 & 65.65 & 65.79 \\
\midrule
DFN2B L/14 & CUB-200 & 72.70 & 72.46 & 70.36 & 73.38 & 74.42 \\
\noalign{\penalty10000}
DFN2B L/14 & NAB-555 & 64.23 & 63.35 & 64.08 & 63.67 & 64.67 \\
\midrule
MetaCLIP H/14 & CUB-200 & 70.57 & 70.47 & 69.43 & 71.68 & 72.93 \\
\noalign{\penalty10000}
MetaCLIP H/14 & NAB-555 & 63.14 & 60.88 & 63.48 & 65.63 & 66.76 \\
\midrule
MetaCLIP L/14 & CUB-200 & 71.02 & 71.25 & 71.35 & 72.92 & 74.92 \\
\noalign{\penalty10000}
MetaCLIP L/14 & NAB-555 & 63.57 & 59.89 & 63.06 & 65.74 & 66.76 \\
\midrule
\multicolumn{7}{l}{\textit{Additional OpenAI backbones}} \\
\midrule
OpenAI L/14 & CUB-200 & 66.63 & 67.51 & 69.11 & 70.86 & 75.05 \\
\noalign{\penalty10000}
OpenAI L/14 & NAB-555 & 60.39 & 57.81 & 60.79 & 60.42 & 65.90 \\
\midrule
OpenAI B/32 & CUB-200 & 64.22 & 65.32 & 67.19 & 68.06 & 69.84 \\
\noalign{\penalty10000}
OpenAI B/32 & NAB-555 & 58.88 & 60.35 & 60.97 & 61.80 & 62.43 \\
\midrule
\end{longtable}
\end{center}

\FloatBarrier
\subsection{CAPP-Multi target results}
\label{app:E_capp_multi}

\begin{center}
\small
\begin{longtable}{llrrrrr}
\caption{All six backbones (DFN/MetaCLIP backbones plus the additional OpenAI L/14 and B/32 backbones, marked below) fine-tuned separately on CUB-200 and on NABirds, each evaluated on its own target training ontology (CUB-trained $\to$ CUB-200, NAB-trained $\to$ NAB-555); PT is the pretrained checkpoint evaluated directly on the target dataset. Multi-way class-slot-macro top-1 accuracy at each method's own P1-selected deployment checkpoint; pairwise version in Table~\ref{tab:S11_target_pairwise}. Structured supervision consistently improves controlled local attribute discrimination on the CUB target ontology, whereas NAB-target effects are more backbone-dependent.}\label{tab:S11_multi}\\
\toprule
Backbone & Train domain & PT & N & U & SSR & SSR+ND \\
\midrule
\endfirsthead
\toprule
Backbone & Train domain & PT & N & U & SSR & SSR+ND \\
\midrule
\endhead
\midrule \multicolumn{7}{r}{\emph{continued on next page}} \\ \midrule
\endfoot
\bottomrule
\endlastfoot
DFN5B H/14-378 & CUB-200 & 49.13 & 49.44 & 50.15 & 52.97 & 53.53 \\
\noalign{\penalty10000}
DFN5B H/14-378 & NAB-555 & 35.93 & 37.34 & 37.77 & 35.33 & 35.29 \\
\midrule
DFN2B L/14 & CUB-200 & 46.32 & 46.24 & 43.19 & 47.37 & 49.70 \\
\noalign{\penalty10000}
DFN2B L/14 & NAB-555 & 32.81 & 31.31 & 32.58 & 32.29 & 32.97 \\
\midrule
MetaCLIP H/14 & CUB-200 & 45.26 & 44.85 & 41.64 & 46.19 & 47.89 \\
\noalign{\penalty10000}
MetaCLIP H/14 & NAB-555 & 31.80 & 27.98 & 30.64 & 33.80 & 35.10 \\
\midrule
MetaCLIP L/14 & CUB-200 & 44.69 & 44.35 & 44.22 & 47.85 & 50.44 \\
\noalign{\penalty10000}
MetaCLIP L/14 & NAB-555 & 31.30 & 26.09 & 29.69 & 32.69 & 33.65 \\
\midrule
\multicolumn{7}{l}{\textit{Additional OpenAI backbones}} \\
\midrule
OpenAI L/14 & CUB-200 & 38.67 & 39.33 & 40.92 & 44.52 & 50.18 \\
\noalign{\penalty10000}
OpenAI L/14 & NAB-555 & 28.45 & 26.52 & 29.16 & 29.05 & 34.12 \\
\midrule
OpenAI B/32 & CUB-200 & 36.19 & 37.27 & 39.47 & 39.89 & 42.04 \\
\noalign{\penalty10000}
OpenAI B/32 & NAB-555 & 28.89 & 28.84 & 29.62 & 29.68 & 29.83 \\
\midrule
\end{longtable}
\end{center}

\FloatBarrier
\subsection{Cross-ontology CAPP}
\label{app:E_capp_held}
Full and strict held-out populations are reported separately and never pooled (\S\ref{app:B_pop}). Held-out CAPP effects are backbone- and transfer-direction-dependent.

\begin{center}
\scriptsize
\begin{longtable}{lllrrrrr}
\caption{All six backbones (DFN/MetaCLIP backbones plus the additional OpenAI L/14 and B/32 backbones, marked below), evaluated cross-ontology in both directions: CUB-trained $\to$ NAB (full held-out NAB-555, strict dataset-class-disjoint NAB-348) and NAB-trained $\to$ CUB (full held-out CUB-200, strict dataset-class-disjoint CUB-59); ``full'' and ``strict'' are evaluation populations, not different training datasets, and are never pooled. PT is the pretrained checkpoint evaluated directly on each held-out population, and all methods are at their own P1-selected deployment checkpoint (pairwise class-slot-macro accuracy). Held-out CAPP effects are backbone- and transfer-direction-dependent; structured target-domain gains (Tables~\ref{tab:S11_target_pairwise}--\ref{tab:S11_multi}) do not imply universal cross-ontology gains.}\label{tab:S12_heldout_capp}\\
\toprule
Backbone & Train$\to$eval & Population & PT & N & U & SSR & SSR+ND \\
\midrule
\endfirsthead
\toprule
Backbone & Train$\to$eval & Population & PT & N & U & SSR & SSR+ND \\
\midrule
\endhead
\midrule \multicolumn{8}{r}{\emph{continued on next page}} \\ \midrule
\endfoot
\bottomrule
\endlastfoot
DFN5B H/14-378 & CUB$\to$NAB & NAB-555 (full) & 65.95 & 67.23 & 68.05 & 67.66 & 66.56 \\
\noalign{\penalty10000}
DFN5B H/14-378 & CUB$\to$NAB & NAB-348 (strict) & 65.63 & 67.05 & 68.22 & 66.85 & 65.92 \\
\noalign{\penalty10000}
DFN5B H/14-378 & NAB$\to$CUB & CUB-200 (full) & 72.99 & 73.99 & 72.88 & 71.10 & 70.50 \\
\noalign{\penalty10000}
DFN5B H/14-378 & NAB$\to$CUB & CUB-59 (strict) & 74.73 & 74.72 & 73.57 & 70.00 & 69.06 \\
\midrule
DFN2B L/14 & CUB$\to$NAB & NAB-555 (full) & 64.24 & 64.07 & 63.56 & 64.01 & 63.97 \\
\noalign{\penalty10000}
DFN2B L/14 & CUB$\to$NAB & NAB-348 (strict) & 63.70 & 63.23 & 62.86 & 63.44 & 63.09 \\
\noalign{\penalty10000}
DFN2B L/14 & NAB$\to$CUB & CUB-200 (full) & 72.71 & 71.65 & 70.93 & 72.81 & 72.41 \\
\noalign{\penalty10000}
DFN2B L/14 & NAB$\to$CUB & CUB-59 (strict) & 71.75 & 71.68 & 70.91 & 72.31 & 72.29 \\
\midrule
MetaCLIP H/14 & CUB$\to$NAB & NAB-555 (full) & 63.13 & 63.27 & 63.29 & 64.70 & 64.72 \\
\noalign{\penalty10000}
MetaCLIP H/14 & CUB$\to$NAB & NAB-348 (strict) & 62.60 & 61.98 & 62.62 & 64.29 & 64.12 \\
\noalign{\penalty10000}
MetaCLIP H/14 & NAB$\to$CUB & CUB-200 (full) & 70.57 & 69.81 & 70.00 & 70.62 & 72.32 \\
\noalign{\penalty10000}
MetaCLIP H/14 & NAB$\to$CUB & CUB-59 (strict) & 68.48 & 66.64 & 68.34 & 68.77 & 70.51 \\
\midrule
MetaCLIP L/14 & CUB$\to$NAB & NAB-555 (full) & 63.61 & 63.26 & 63.36 & 64.95 & 64.69 \\
\noalign{\penalty10000}
MetaCLIP L/14 & CUB$\to$NAB & NAB-348 (strict) & 63.04 & 62.47 & 62.45 & 64.25 & 63.92 \\
\noalign{\penalty10000}
MetaCLIP L/14 & NAB$\to$CUB & CUB-200 (full) & 71.02 & 65.69 & 70.47 & 72.74 & 73.10 \\
\noalign{\penalty10000}
MetaCLIP L/14 & NAB$\to$CUB & CUB-59 (strict) & 69.55 & 64.56 & 70.06 & 72.25 & 71.49 \\
\midrule
\multicolumn{8}{l}{\textit{Additional OpenAI backbones}} \\
\midrule
OpenAI L/14 & CUB$\to$NAB & NAB-555 (full) & 60.39 & 60.17 & 60.15 & 61.66 & 63.07 \\
\noalign{\penalty10000}
OpenAI L/14 & CUB$\to$NAB & NAB-348 (strict) & 59.71 & 59.14 & 58.80 & 61.08 & 62.25 \\
\noalign{\penalty10000}
OpenAI L/14 & NAB$\to$CUB & CUB-200 (full) & 66.63 & 64.80 & 68.18 & 68.17 & 71.46 \\
\noalign{\penalty10000}
OpenAI L/14 & NAB$\to$CUB & CUB-59 (strict) & 66.67 & 63.46 & 68.44 & 68.74 & 72.35 \\
\midrule
OpenAI B/32 & CUB$\to$NAB & NAB-555 (full) & 58.91 & 60.14 & 59.93 & 60.55 & 61.29 \\
\noalign{\penalty10000}
OpenAI B/32 & CUB$\to$NAB & NAB-348 (strict) & 58.40 & 59.56 & 58.80 & 60.40 & 60.74 \\
\noalign{\penalty10000}
OpenAI B/32 & NAB$\to$CUB & CUB-200 (full) & 64.22 & 65.50 & 65.56 & 67.77 & 68.42 \\
\noalign{\penalty10000}
OpenAI B/32 & NAB$\to$CUB & CUB-59 (strict) & 64.71 & 66.01 & 62.72 & 65.34 & 67.35 \\
\midrule
\end{longtable}
\end{center}

\FloatBarrier
\subsection{Direct name-access ablation: quantitative results}
\label{app:E_trackA}
The FULL/PARTIAL/NO variants are defined in \S\ref{app:ssr_name_access}. They control ATTRIBUTE-route NAME access only; EOT remains globally name-accessible, so even NO is not globally name-free.

%% file: supplement/tables/audit/E_trackA_p3.tex
\begin{table}[!htbp]
\centering
\small\setlength{\tabcolsep}{3.5pt}
\begin{tabular}{llrrrrrr}
\toprule
Variant & Eval. & P3 PT & P3 peak (ep.) & P3 deploy (ep.) & Gain$_{peak}$ & Gain$_{deploy}$ & Forget \\
\midrule
FULL & CUB-200 & 34.36 & 36.16 (4) & 34.17 (20) & 1.79 & -0.19 & 1.99 \\
 & NAB-348 & 13.81 & 16.26 (12) & 15.91 (20) & 2.45 & 2.10 & 0.35 \\
\midrule
PARTIAL & CUB-200 & 34.45 & 37.80 (12) & 36.97 (75) & 3.35 & 2.52 & 0.83 \\
 & NAB-348 & 13.87 & 17.01 (12) & 13.38 (75) & 3.14 & -0.49 & 3.63 \\
\midrule
NO & CUB-200 & 34.35 & 40.42 (20) & 40.42 (20) & 6.08 & 6.08 & 0.00 \\
 & NAB-348 & 13.85 & 17.32 (12) & 16.90 (20) & 3.48 & 3.06 & 0.42 \\
\midrule
\bottomrule
\end{tabular}
\caption{Direct name-access variants (FULL / PARTIAL / NO), DFN5B H/14-378 fine-tuned on CUB-200, one run per variant (seed 0). Only the routing mask differs. FULL: the ATTRIBUTE route sees the full NAME segment; PARTIAL: only the NAME\_END boundary signal; NO: no NAME access (identical to the canonical SSR run). EOT remains globally name-accessible in all three. Deployment is the CUB-P1-selected epoch (FULL 20, PARTIAL 75, NO 20), so PARTIAL's deployment epoch differs from the others.}
\label{tab:E_trackA_p3}
\end{table}

%% file: supplement/tables/audit/E_trackA_capp.tex
\begin{table}[!htbp]
\centering
\small\setlength{\tabcolsep}{3.5pt}
\begin{tabular}{llrrrr}
\toprule
Variant & Eval. & Ep.\ 0 (PT) & Ep.\ 8 & Ep.\ 20 & Ep.\ 100 \\
\midrule
FULL & CUB-200 & 72.99 & 74.28 & 74.85 & 73.74 \\
 & NAB-348 & 65.62 & 65.65 & 66.72 & 65.81 \\
\midrule
PARTIAL & CUB-200 & 72.97 & 75.14 & 75.59 & 74.32 \\
 & NAB-348 & 65.62 & 65.94 & 67.05 & 66.07 \\
\midrule
NO & CUB-200 & 72.99 & 75.44 & 76.31 & 75.06 \\
 & NAB-348 & 65.62 & 65.94 & 66.85 & 65.60 \\
\midrule
\bottomrule
\end{tabular}
\caption{Pairwise CAPP (class-slot-macro accuracy, \%) for the name-access variants at matched epochs. On the CUB target, CAPP follows NO $>$ PARTIAL $>$ FULL at epochs 8, 20, and 100 (differences of 0.3--1.5\,pp); on NAB-348 the variants are nearly indistinguishable and the gain is transient.}
\label{tab:E_trackA_capp}
\end{table}

%% file: supplement/tables/audit/E_trackA_p2.tex
\begin{table}[!htbp]
\centering
\small\setlength{\tabcolsep}{3.5pt}
\begin{tabular}{llrrrrrr}
\toprule
Variant & Eval. & Dep.\ ep. & P2 ord. & P2 name-sw. & P2 attr-sw. & Name-sw.\ damage & Attr-sw.\ damage \\
\midrule
FULL & CUB-200 & 20 & 89.90 & 0.05 & 86.52 & 89.85 & 3.38 \\
 & NAB-348 & 20 & 79.66 & 0.30 & 69.82 & 79.35 & 9.83 \\
\midrule
PARTIAL & CUB-200 & 75 & 90.28 & 0.07 & 88.64 & 90.21 & 1.64 \\
 & NAB-348 & 75 & 73.42 & 0.23 & 65.60 & 73.19 & 7.83 \\
\midrule
NO & CUB-200 & 20 & 89.51 & 0.12 & 85.42 & 89.39 & 4.09 \\
 & NAB-348 & 20 & 78.97 & 0.33 & 68.39 & 78.64 & 10.59 \\
\midrule
\bottomrule
\end{tabular}
\caption{Ordinary, name-swapped, and attribute-swapped P2 (R@1 \%) at each variant's deployment epoch. Name-swap collapses P2 to $\le$0.33\% in every variant, so restricting the ATTRIBUTE route's name access does not remove the class-name dominance of ordinary P2.}
\label{tab:E_trackA_p2}
\end{table}

%% file: supplement/tables/audit/E_trackA_rsa.tex
\begin{table}[!htbp]
\centering
\small\setlength{\tabcolsep}{3.5pt}
\begin{tabular}{lrrrr}
\toprule
Variant & Attr.\ end-mean & Name end-mean & Attr.\ pool-proj & Name pool-proj \\
\midrule
FULL & 0.39 & 0.84 & 0.40 & 0.83 \\
PARTIAL & 0.40 & 0.76 & 0.40 & 0.76 \\
NO & 0.65 & 0.35 & 0.68 & 0.36 \\
\bottomrule
\end{tabular}
\caption{Representational similarity (RSA) of the ATTRIBUTE-route representation at the last layer, epoch 20, to the pretrained name-free attribute-profile geometry and to class-name geometry, under two readouts (att\_end\_mean and att\_pool\_proj). Restricting direct NAME access raises similarity to the attribute-profile geometry and lowers similarity to class-name geometry under both readouts (main-text summary values use att\_end\_mean for the attribute profile and att\_pool\_proj for class name; both are shown here).}
\label{tab:E_trackA_rsa}
\end{table}

%% file: supplement/tables/audit/F_prev.tex
\begin{table}[!htbp]
\centering\small\setlength{\tabcolsep}{5pt}
\begin{tabular}{llrrrr}
\toprule
Population & Method & Cells & Before & At & After \\
\midrule
target & N & 6 & 2 & 4 & 0 \\
 & U & 6 & 5 & 1 & 0 \\
 & SSR & 6 & 3 & 3 & 0 \\
 & SSR+ND & 6 & 1 & 5 & 0 \\
& \emph{all} & 24 & 11 & 13 & 0 \\
\midrule
same-ontology & N & 18 & 6 & 12 & 0 \\
 & U & 18 & 15 & 3 & 0 \\
 & SSR & 18 & 13 & 5 & 0 \\
 & SSR+ND & 18 & 6 & 12 & 0 \\
& \emph{all} & 72 & 40 & 32 & 0 \\
\midrule
cross-ontology & N & 18 & 16 & 2 & 0 \\
 & U & 18 & 17 & 1 & 0 \\
 & SSR & 18 & 17 & 1 & 0 \\
 & SSR+ND & 18 & 15 & 3 & 0 \\
& \emph{all} & 72 & 65 & 7 & 0 \\
\bottomrule
\end{tabular}
\caption{Timing of each evaluation distribution's own P3-peak milestone relative to the run's P1-selected deployment milestone, by method (six ImageNet-trained backbones). Target = ImageNet val (6 cells per method); same-ontology = ImageNet-V2/-R/-Sketch (18); cross-ontology = CUB-200, NAB-348, FineView Butterfly (18). Milestones are cumulative-data indices and are compared only categorically. Panel~b of Figure~\ref{fig:specialization} plots ``before'' for target and cross-ontology. Six heterogeneous backbones, not replicates.}
\label{tab:F_prev}
\end{table}

%% file: supplement/tables/audit/F_mag.tex
\begin{table}[!htbp]
\centering\small\setlength{\tabcolsep}{5pt}
\begin{tabular}{llrrrr}
\toprule
Population & Method & Valid $n$ & Median (pp) & IQR & $>0.5$\,pp \\
\midrule
target & N & 2 & 0.34 & [0.32, 0.36] & 0 \\
 & U & 5 & 3.32 & [1.00, 5.56] & 4 \\
 & SSR & 3 & 0.34 & [0.28, 1.00] & 1 \\
 & SSR+ND & 1 & 0.12 & [0.12, 0.12] & 0 \\
& \emph{all} & 11 & 0.38 & [0.31, 2.49] & 5 \\
\midrule
cross-ontology & N & 16 & 1.40 & [0.62, 2.01] & 14 \\
 & U & 17 & 1.21 & [0.58, 5.04] & 13 \\
 & SSR & 17 & 1.71 & [0.83, 2.87] & 15 \\
 & SSR+ND & 15 & 1.35 & [0.90, 1.80] & 12 \\
& \emph{all} & 65 & 1.35 & [0.64, 2.82] & 54 \\
\midrule
by dataset & CUB-200 & 23 & 1.78 & [1.16, 5.00] & 22 \\
 & NAB-348 & 23 & 1.76 & [0.82, 2.85] & 20 \\
 & FineView Butterfly & 19 & 0.58 & [0.47, 1.38] & 12 \\
\bottomrule
\end{tabular}
\caption{Peak-to-deployment P3 decrease, $\mathrm{P3}_{\mathrm{peak}}-\mathrm{P3}_{\mathrm{deploy}}$ (absolute pp), over cells whose own P3 peak precedes deployment ($t_{P3}<t_{\mathrm{deploy}}$); cells with peak at or after deployment have no evaluated post-peak interval and are excluded. Only 11 target cells qualify (n = 1--5 per method), too few for stable per-method target summaries, so Figure~\ref{fig:specialization}c shows cross-ontology cells only. The $>0.5$\,pp count is a descriptive robustness count, not a significance test. The by-dataset rows are cross-ontology cells pooled over methods. OpenAI L/14 target and same-ontology cells all have the peak at the final milestone (Table~\ref{tab:F_openai}), so absence of observed decline there is bounded by the evaluated schedule.}
\label{tab:F_mag}
\end{table}

%% file: supplement/tables/generated_timing_tables.tex
% AUTO-GENERATED by scripts/build_supplement_tables.py

\begin{center}
\small
\begin{longtable}{lrrrrrrrrr}
\caption{Per-backbone counts of evaluation cells (over the four methods N/U/SSR/SSR+ND and the datasets in each population) whose P3 peak milestone falls before, at, or after the run's P1-selected deployment milestone. T = ImageNet val target (4 cells per backbone), S = same-ontology shifts ImageNet-V2/-R/-Sketch (12 cells), X = cross-ontology CUB-200/NAB-348/FVB (12 cells). Classification compares peak and deployment milestone indices from Tables~\ref{tab:S1_full}--\ref{tab:S2_full}; milestones are cumulative-data indices, not epochs, and are never pooled as absolute values across schedules. Column sums over backbones give the population totals in Table~\ref{tab:F_prev}.}\label{tab:F_counts}\\
\toprule
Backbone & T-bef & T-at & T-aft & S-bef & S-at & S-aft & X-bef & X-at & X-aft \\
\midrule
\endfirsthead
\toprule
Backbone & T-bef & T-at & T-aft & S-bef & S-at & S-aft & X-bef & X-at & X-aft \\
\midrule
\endhead
\midrule \multicolumn{10}{r}{\emph{continued on next page}} \\ \midrule
\endfoot
\bottomrule
\endlastfoot
DFN5B H/14-378 & 3 & 1 & 0 & 10 & 2 & 0 & 12 & 0 & 0 \\
DFN2B L/14 & 2 & 2 & 0 & 6 & 6 & 0 & 9 & 3 & 0 \\
MetaCLIP H/14 & 1 & 3 & 0 & 6 & 6 & 0 & 11 & 1 & 0 \\
MetaCLIP L/14 & 3 & 1 & 0 & 7 & 5 & 0 & 11 & 1 & 0 \\
OpenAI L/14 & 0 & 4 & 0 & 0 & 12 & 0 & 11 & 1 & 0 \\
OpenAI B/32 & 2 & 2 & 0 & 11 & 1 & 0 & 11 & 1 & 0 \\
\midrule
All six & 11 & 13 & 0 & 40 & 32 & 0 & 65 & 7 & 0 \\
\end{longtable}
\end{center}

\FloatBarrier

\begin{center}
\small
\begin{longtable}{llrrrr}
\caption{OpenAI L/14 (ImageNet-trained): peak and deployment milestone indices for the ImageNet target and same-ontology shifts. In every one of these 16 cells the P1-selected deployment checkpoint is the final evaluated milestone (11) and coincides with the P3 maximum, so no post-peak interval exists in the evaluated trajectory; this is a statement about the audited schedule, not evidence that specialization cannot occur for this backbone. Cross-ontology cells (Table~\ref{tab:S2_full}) do show peaks before deployment.}\label{tab:F_openai}\\
\toprule
Dataset & Method & Peak ms. & Deploy ms. & Gain$_{peak}$ & Forget \\
\midrule
\endfirsthead
\toprule
Dataset & Method & Peak ms. & Deploy ms. & Gain$_{peak}$ & Forget \\
\midrule
\endhead
\midrule \multicolumn{6}{r}{\emph{continued on next page}} \\ \midrule
\endfoot
\bottomrule
\endlastfoot
ImageNet val & N & 11 & 11 & 5.24 & 0.00 \\
 & U & 11 & 11 & 2.28 & 0.00 \\
 & SSR & 11 & 11 & 9.66 & 0.00 \\
 & SSR+ND & 11 & 11 & 19.88 & 0.00 \\
\midrule
ImageNet-V2 & N & 11 & 11 & 4.74 & 0.00 \\
 & U & 11 & 11 & 1.89 & 0.00 \\
 & SSR & 11 & 11 & 8.75 & 0.00 \\
 & SSR+ND & 11 & 11 & 16.95 & 0.00 \\
\midrule
ImageNet-R & N & 11 & 11 & 6.47 & 0.00 \\
 & U & 11 & 11 & 5.83 & 0.00 \\
 & SSR & 11 & 11 & 10.88 & 0.00 \\
 & SSR+ND & 11 & 11 & 21.63 & 0.00 \\
\midrule
ImageNet-Sketch & N & 11 & 11 & 5.55 & 0.00 \\
 & U & 11 & 11 & 2.46 & 0.00 \\
 & SSR & 11 & 11 & 7.31 & 0.00 \\
 & SSR+ND & 11 & 11 & 12.47 & 0.00 \\
\midrule
\end{longtable}
\end{center}

%% file: supplement/tables/audit/G_move_order.tex
\begin{table}[!htbp]
\centering\small\setlength{\tabcolsep}{4pt}
\begin{tabular}{llrrr}
\toprule
Training domain & Cells & $t_{P3}<t_{\mathrm{deploy}}$ & $t_{P3}=t_{\mathrm{deploy}}$ & $t_{P3}>t_{\mathrm{deploy}}$ \\
\midrule
CUB-trained & held-out & 23 & 0 & 1 \\
CUB-trained & target & 11 & 1 & 0 \\
ImageNet-trained & held-out & 33 & 3 & 0 \\
ImageNet-trained & target & 10 & 2 & 0 \\
NABirds-trained & held-out & 22 & 1 & 1 \\
NABirds-trained & target & 8 & 3 & 1 \\
\midrule
All training domains & held-out & 78 & 4 & 2 \\
All training domains & target & 29 & 6 & 1 \\
\bottomrule
\end{tabular}
\caption{Temporal order of each evaluation distribution's own P3-peak checkpoint and the P1-selected deployment checkpoint over the 120 N/U/SSR cells (4 backbones $\times$ 3 training domains $\times$ 3 methods $\times$ evaluation sets). Only $t_{P3}<t_{\mathrm{deploy}}$ cells are temporally valid ``post-peak'' trajectories.}
\label{tab:G_move_order}
\end{table}

%% file: supplement/tables/audit/G_move_tower.tex
\begin{table}[!htbp]
\centering\scriptsize\setlength{\tabcolsep}{2.6pt}
\begin{tabular}{lrcccrccc}
\toprule
& \multicolumn{4}{c}{Pre-peak vs.\ $\mathrm{Gain}_{\mathrm{peak}}$} & \multicolumn{4}{c}{Post-peak vs.\ $\mathrm{Forget}_{\mathrm{postpeak}}$} \\
\cmidrule(lr){2-5}\cmidrule(lr){6-9}
Subset & $n$ & Image & Text & Joint & $n$ & Image & Text & Joint \\
\midrule
Held-out pooled (primary) & 84 & $-$0.17/+0.33 & $-$0.14/+0.35 & $-$0.16/+0.32 & 78 & +0.43/+0.51 & +0.54/+0.64 & +0.50/+0.61 \\
\midrule
CUB-trained & 24 & +0.49/+0.58 & +0.63/+0.69 & +0.58/+0.63 & 23 & +0.42/+0.49 & +0.68/+0.72 & +0.58/+0.59 \\
NABirds-trained & 24 & $-$0.72/$-$0.65 & $-$0.73/$-$0.70 & $-$0.73/$-$0.72 & 22 & +0.64/+0.64 & +0.56/+0.79 & +0.64/+0.75 \\
ImageNet-trained & 36 & +0.33/+0.57 & +0.35/+0.57 & +0.34/+0.53 & 33 & +0.10/$-$0.02 & +0.56/+0.54 & +0.39/+0.43 \\
\midrule
N & 28 & $-$0.25/+0.31 & $-$0.23/+0.35 & $-$0.24/+0.32 & 26 & +0.45/+0.54 & +0.40/+0.58 & +0.44/+0.60 \\
U & 28 & +0.01/+0.49 & +0.08/+0.52 & +0.04/+0.48 & 26 & +0.33/+0.16 & +0.61/+0.46 & +0.48/+0.35 \\
SSR & 28 & $-$0.31/$-$0.01 & $-$0.31/$-$0.06 & $-$0.31/$-$0.09 & 26 & +0.59/+0.80 & +0.71/+0.85 & +0.66/+0.83 \\
\midrule
DFN5B H/14-378 & 21 & $-$0.29/+0.32 & $-$0.29/+0.25 & $-$0.29/+0.25 & 21 & +0.65/+0.83 & +0.64/+0.83 & +0.65/+0.83 \\
DFN2B L/14 & 21 & $-$0.08/+0.68 & $-$0.04/+0.64 & $-$0.06/+0.66 & 17 & +0.51/+0.63 & +0.55/+0.69 & +0.55/+0.69 \\
MetaCLIP H/14 & 21 & $-$0.25/+0.09 & $-$0.22/+0.17 & $-$0.24/+0.08 & 21 & +0.41/+0.44 & +0.39/+0.43 & +0.40/+0.44 \\
MetaCLIP L/14 & 21 & $-$0.15/+0.12 & $-$0.12/+0.07 & $-$0.13/+0.09 & 19 & +0.29/+0.32 & +0.21/+0.32 & +0.26/+0.32 \\
\midrule
\multicolumn{9}{l}{\emph{Text-component robustness (pooled held-out; Pearson/Spearman)}} \\
Text transformer only & 84 & & $-$0.14/+0.35 & & 78 & & +0.54/+0.64 & \\
Text transformer + projection + final LayerNorm & 84 & & $-$0.12/+0.40 & & 78 & & +0.54/+0.61 & \\
Pretrained token-embedding rows only & 84 & & $-$0.15/+0.27 & & 78 & & +0.41/+0.53 & \\
\bottomrule
\end{tabular}
\caption{Encoder-separated and joint movement analysis (Pearson $r$ / Spearman $\rho$), held-out cells of the N/U/SSR grid; post-peak columns use $t_{P3}<t_{\mathrm{deploy}}$ cells. Image movement: mean over image-encoder blocks (\texttt{visual.transformer.resblocks.*}) of $\|\Delta W\|_F/\|W_{\mathrm{PT}}\|_F$. Text movement: the same quantity over the shared pretrained text transformer (\texttt{transformer.resblocks.*}; 24 blocks for DFN5B/MetaCLIP H/14, 12 for DFN2B/MetaCLIP L/14). Joint = (image + text)/2, the ``pretrained-backbone movement'' used in Figure~\ref{fig:mechanism}a. Excluded because they lack a pretrained analogue or are not comparable: AIP and StructuredTextFusion parameters, the newly appended SSR token-embedding rows (4--6), positional embeddings, and the logit scale. Descriptive associations across heterogeneous backbones; not replicates.}
\label{tab:G_move_tower}
\end{table}

%% file: supplement/tables/audit/G_move_joint.tex
\begin{table}[!htbp]
\centering\scriptsize\setlength{\tabcolsep}{2.4pt}
\begin{tabular}{lrrrrrrrrrr}
\toprule
& & \multicolumn{3}{c}{Pearson $r$} & \multicolumn{2}{c}{Partial $r$} & \multicolumn{2}{c}{Std.\ coef.} & & \\
\cmidrule(lr){3-5}\cmidrule(lr){6-7}\cmidrule(lr){8-9}
Subset / normalization & $n$ & Img--$\mathrm{F}$ & Txt--$\mathrm{F}$ & Img--Txt & Img$|$Txt & Txt$|$Img & Img & Txt & $R^2$ & VIF \\
\midrule
Pooled (raw) & 78 & +0.43 & +0.54 & +0.85 & $-$0.06 & +0.36 & $-$0.10 & +0.62 & 0.29 & 3.5 \\
CUB-trained & 23 & +0.42 & +0.68 & +0.72 & $-$0.14 & +0.60 & $-$0.15 & +0.79 & 0.48 & 2.1 \\
NABirds-trained & 22 & +0.64 & +0.56 & +0.76 & +0.40 & +0.14 & +0.51 & +0.17 & 0.42 & 2.4 \\
ImageNet-trained & 33 & +0.10 & +0.56 & +0.45 & $-$0.22 & +0.59 & $-$0.20 & +0.66 & 0.35 & 1.3 \\
Domain-standardized & 78 & +0.34 & +0.60 & +0.62 & $-$0.04 & +0.52 & $-$0.04 & +0.62 & 0.36 & 1.6 \\
Within-backbone standardized & 78 & +0.47 & +0.45 & +0.95 & +0.16 & +0.00 & +0.46 & +0.01 & 0.22 & 10.8 \\
Within backbone$\times$domain standardized & 78 & +0.54 & +0.56 & +0.90 & +0.10 & +0.21 & +0.19 & +0.39 & 0.33 & 5.1 \\
\bottomrule
\end{tabular}
\caption{Post-peak image versus text movement, held-out $t_{P3}<t_{\mathrm{deploy}}$ cells (n = 78 pooled). F = $\mathrm{Forget}_{\mathrm{postpeak}}$; Img--Txt is the image/text movement correlation; Partial $r$ controls the other encoder; Std.\ coef.\ are from the descriptive regression $\mathrm{F}\sim z(D_{\mathrm{image}})+z(D_{\mathrm{text}})$. Image and text movement are strongly coupled, and the apparent relative importance of the encoders depends on normalization (text dominates pooled and domain-standardized, image and text are indistinguishable within backbone; VIF up to about 11), so broad encoder attribution is not identifiable. Descriptive only; not a population-level regression or causal claim.}
\label{tab:G_move_joint}
\end{table}

%% file: supplement/tables/audit/G_move_corr.tex
\begin{table}[!htbp]
\centering\footnotesize\setlength{\tabcolsep}{3pt}
\begin{tabular}{lrrrrrr}
\toprule
& \multicolumn{3}{c}{Pre-peak: $D_{\mathrm{pre}}$ vs.\ $\mathrm{Gain}_{\mathrm{peak}}$} & \multicolumn{3}{c}{Post-peak: $D_{\mathrm{post}}$ vs.\ $\mathrm{Forget}_{\mathrm{postpeak}}$} \\
\cmidrule(lr){2-4}\cmidrule(lr){5-7}
Subset & $n$ & $r$ & $\rho$ & $n$ & $r$ & $\rho$ \\
\midrule
Held-out, all (primary) & 84 & $-$0.17 & +0.33 & 78 & +0.43 & +0.51 \\
All valid cells & 108 & $-$0.15 & +0.28 & 107 & +0.44 & +0.52 \\
\midrule
CUB-trained (held-out) & 24 & +0.49 & +0.58 & 23 & +0.42 & +0.49 \\
NABirds-trained (held-out) & 24 & $-$0.72 & $-$0.65 & 22 & +0.64 & +0.64 \\
ImageNet-trained (held-out) & 36 & +0.33 & +0.57 & 33 & +0.10 & $-$0.02 \\
CUB+NABirds-trained (held-out) & 48 & $-$0.37 & $-$0.08 & 45 & +0.65 & +0.69 \\
\midrule
N (held-out) & 28 & $-$0.25 & +0.31 & 26 & +0.45 & +0.54 \\
U (held-out) & 28 & +0.01 & +0.49 & 26 & +0.33 & +0.16 \\
SSR (held-out) & 28 & $-$0.31 & $-$0.01 & 26 & +0.59 & +0.80 \\
\midrule
DFN5B H/14-378 (held-out) & 21 & $-$0.29 & +0.32 & 21 & +0.65 & +0.83 \\
DFN2B L/14 (held-out) & 21 & $-$0.08 & +0.68 & 17 & +0.51 & +0.63 \\
MetaCLIP H/14 (held-out) & 21 & $-$0.25 & +0.09 & 21 & +0.41 & +0.44 \\
MetaCLIP L/14 (held-out) & 21 & $-$0.15 & +0.12 & 19 & +0.29 & +0.32 \\
\midrule
\multicolumn{7}{l}{\emph{Robustness (post-peak, held-out, $t_{P3}<t_{\mathrm{deploy}}$)}} \\
Peak-normalized $D_{\mathrm{post}}$ & & & & 78 & +0.43 & +0.51 \\
Distance-to-pretrained radial quantity (distance-to-PT difference) & & & & 78 & +0.56 & +0.63 \\
Distance-to-pretrained radial, layer-aggregated & & & & 78 & +0.50 & +0.57 \\
Full grid, symmetric distance (held-out, incl.\ $t_{P3}\ge t_{\mathrm{deploy}}$) & & & & 84 & +0.46 & +0.60 \\
Domain-demeaned pooled & & & & 78 & +0.31 & +0.26 \\
\bottomrule
\end{tabular}
\caption{Direct image-encoder movement versus P3 change across the N/U/SSR trajectory grid (Pearson $r$, Spearman $\rho$; absolute percentage points). $D_{\mathrm{pre}}$ and $D_{\mathrm{post}}$ are the layer-averaged Frobenius displacements $\|W_{\mathrm{peak}}-W_{\mathrm{PT}}\|/\|W_{\mathrm{PT}}\|$ and $\|W_{\mathrm{deploy}}-W_{\mathrm{peak}}\|/\|W_{\mathrm{PT}}\|$ over all image-encoder transformer-block parameters (attention, MLP, LayerNorm; projection, embeddings, text encoder and SSR modules excluded); net checkpoint displacement, not SGD path length. Post-peak rows use held-out cells with $t_{P3}<t_{\mathrm{deploy}}$ unless stated; pre-peak rows need a pretrained P3 baseline. Backbones are heterogeneous models, not replicates, and all values are descriptive associations.}
\label{tab:G_move_corr}
\end{table}

%% file: supplement/tables/audit/G_hardness.tex
\begin{table}[!htbp]
\centering\scriptsize\setlength{\tabcolsep}{2.2pt}
\begin{tabular}{lrrrrlrl}
\toprule
Probe, level & Rank & Wrong sim.\ peak$\to$deploy & $\Delta$corr. & $\Delta$wrong & Encroachment [95\% CI] & \% cls.\ $\downarrow$ margin & Split-half [95\% CI] \\
\midrule
P3 hardest & 1 & 0.203 $\to$ 0.211 & +0.0044 & +0.0080 & +0.0036 [+0.0019, +0.0052] & 59.5 & +0.0034 [+0.0014, +0.0053] \\
P3 hard & 35 & 0.090 $\to$ 0.098 & +0.0044 & +0.0084 & +0.0039 [+0.0023, +0.0055] & 62.6 & +0.0047 [+0.0024, +0.0070] \\
P3 medium & 174 & $-$0.007 $\to$ 0.005 & +0.0044 & +0.0120 & +0.0076 [+0.0057, +0.0094] & 67.0 & +0.0072 [+0.0049, +0.0095] \\
P3 easy & 312 & $-$0.096 $\to$ $-$0.079 & +0.0044 & +0.0168 & +0.0124 [+0.0103, +0.0144] & 74.1 & +0.0116 [+0.0088, +0.0141] \\
\midrule
P1 hardest & 1 & 0.356 $\to$ 0.399 & +0.0408 & +0.0430 & +0.0022 [+0.0006, +0.0038] & 56.0 & +0.0021 [+0.0004, +0.0039] \\
P1 hard & 35 & 0.089 $\to$ 0.127 & +0.0408 & +0.0375 & $-$0.0033 [$-$0.0054, $-$0.0010] & 42.0 & $-$0.0044 [$-$0.0074, $-$0.0013] \\
P1 medium & 174 & $-$0.039 $\to$ 0.005 & +0.0408 & +0.0436 & +0.0029 [+0.0007, +0.0051] & 54.9 & +0.0028 [$-$0.0002, +0.0058] \\
P1 easy & 312 & $-$0.152 $\to$ $-$0.103 & +0.0408 & +0.0491 & +0.0083 [+0.0060, +0.0107] & 64.1 & +0.0087 [+0.0055, +0.0120] \\
\bottomrule
\end{tabular}
\caption{Peak-ranked wrong-class levels for the SSR+ND DFN5B H/14-378 CUB$\to$NAB-348 run. For every image the 347 wrong classes are ranked by similarity at the peak (epoch 20) only (rank 1 = most similar); the class at rank 1, 35 (about the 90th similarity percentile), 174 (median) and 312 (about the 10th percentile) is held fixed and reevaluated at deployment (epoch 60). Encroachment $=\Delta s_{\mathrm{wrong}}-\Delta s_{\mathrm{correct}}=-\Delta$margin, with margin = correct $-$ wrong cosine similarity; class-macro means over the 348 NAB-348 classes, paired class bootstrap (10,000 resamples). The split-half column ranks wrong classes from half of each class's images and evaluates on the other half. P1 and P3 are separate capabilities and are not pooled.}
\label{tab:G_hardness}
\end{table}

%% file: supplement/tables/audit/G_tower.tex
\begin{table}[!htbp]
\centering
\small\setlength{\tabcolsep}{3.5pt}
\begin{tabular}{lllrrrr}
\toprule
State & Image encoder & Text/routing encoder & CUB P1 & CUB P3 & NAB-348 P1 & NAB-348 P3 \\
\midrule
peak & ep. 20 & ep. 20 & 90.02 & 45.56 & 76.10 & 17.63 \\
image late only & ep. 60 & ep. 20 & 89.97 & 45.20 & 71.23 & 15.15 \\
text late only & ep. 20 & ep. 60 & 89.97 & 47.07 & 76.07 & 17.09 \\
deploy & ep. 60 & ep. 60 & 90.23 & 46.29 & 70.81 & 14.50 \\
\bottomrule
\end{tabular}
\caption{Encoder-state swap, DFN5B H/14-378 SSR+ND fine-tuned on CUB-200 (single run). Image encoder and text/routing encoder are taken from epoch 20 (peak) or epoch 60 (deployment); ``image late only'' combines the epoch-60 image encoder with the epoch-20 text/routing encoder, ``text late only'' the reverse. Values are R@1 (\%).}
\label{tab:G_tower}
\end{table}

%% file: supplement/tables/audit/G_towerboot.tex
\begin{table}[!htbp]
\centering\small\setlength{\tabcolsep}{4pt}
\begin{tabular}{lllr}
\toprule
Eval. & Probe & Effect & Change (pp) [95\% CI] \\
\midrule
CUB-200 & P3 & image-late & $-$0.44 [$-$1.84, +0.89] \\
CUB-200 & P3 & text/SSR-late & +1.45 [+0.55, +2.37] \\
CUB-200 & P3 & total & +0.59 [$-$1.06, +2.18] \\
CUB-200 & P3 & interaction & $-$0.42 [$-$1.15, +0.29] \\
CUB-200 & P3 & image-late $-$ text/SSR-late & $-$1.90 [$-$3.36, $-$0.44] \\
\addlinespace
NAB-348 & P3 & image-late & $-$2.53 [$-$3.53, $-$1.56] \\
NAB-348 & P3 & text/SSR-late & $-$0.66 [$-$1.19, $-$0.15] \\
NAB-348 & P3 & total & $-$3.21 [$-$4.49, $-$2.03] \\
NAB-348 & P3 & interaction & $-$0.02 [$-$0.52, +0.46] \\
NAB-348 & P3 & image-late $-$ text/SSR-late & $-$1.88 [$-$2.84, $-$0.90] \\
\addlinespace
NAB-348 & P1 & image-late & $-$4.94 [$-$6.13, $-$3.79] \\
NAB-348 & P1 & text/SSR-late & $-$0.04 [$-$0.51, +0.45] \\
NAB-348 & P1 & total & $-$5.28 [$-$6.56, $-$4.01] \\
NAB-348 & P1 & interaction & $-$0.30 [$-$0.77, +0.15] \\
NAB-348 & P1 & image-late $-$ text/SSR-late & $-$4.90 [$-$6.12, $-$3.74] \\
\addlinespace
CUB-200 & P1 & image-late & $-$0.08 [$-$0.65, +0.49] \\
CUB-200 & P1 & text/SSR-late & $-$0.04 [$-$0.70, +0.64] \\
CUB-200 & P1 & total & +0.23 [$-$0.45, +0.95] \\
CUB-200 & P1 & interaction & +0.36 [$-$0.14, +0.85] \\
CUB-200 & P1 & image-late $-$ text/SSR-late & $-$0.03 [$-$0.78, +0.73] \\
\addlinespace
\bottomrule
\end{tabular}
\caption{Paired class-bootstrap decomposition of the encoder-state swap (Table~\ref{tab:G_tower}); representative DFN5B H/14-378 SSR+ND CUB-trained run. ``Image-late'' is the effect of substituting the deployment image encoder at fixed peak text/routing state, ``text/SSR-late'' the reverse, ``interaction'' the residual of the additive decomposition. Class bootstrap over evaluation classes (10,000 resamples).}
\label{tab:G_towerboot}
\end{table}

%% file: supplement/tables/audit/G_swapext.tex
\begin{table}[!htbp]
\centering\tiny\setlength{\tabcolsep}{2.2pt}
\resizebox{\linewidth}{!}{%
\begin{tabular}{lrrrrllll}
\toprule
& \multicolumn{4}{c}{Class-macro R@1 (\%)} & \multicolumn{4}{c}{Effect (pp) [95\% CI]} \\
\cmidrule(lr){2-5}\cmidrule(lr){6-9}
Evaluation & P$_I$+P$_T$ & D$_I$+P$_T$ & P$_I$+D$_T$ & D$_I$+D$_T$ & Image-late & Text/SSR-late & Image$-$text & Interaction \\
\midrule
\multicolumn{9}{l}{\emph{MetaCLIP H/14, CUB$\to$NAB-348 (ep.\ 12/75)}} \\
held-out, NAB-348 P3 & 12.39 & 11.66 & 12.46 & 11.64 & $-$0.73 [$-$1.54, +0.08] & +0.06 [$-$0.38, +0.50] & $-$0.80 [$-$1.68, +0.10] & $-$0.08 [$-$0.52, +0.34] \\
held-out, NAB-348 P1 & 72.12 & 69.67 & 72.20 & 70.37 & $-$2.44 [$-$3.62, $-$1.30] & +0.08 [$-$0.58, +0.74] & $-$2.52 [$-$3.82, $-$1.25] & +0.61 [+0.12, +1.11] \\
target, CUB-200 P3 & 38.35 & 40.35 & 39.33 & 42.01 & +2.00 [+0.51, +3.53] & +0.98 [$-$0.01, +1.95] & +1.02 [$-$0.32, +2.35] & +0.68 [$-$0.26, +1.69] \\
target, CUB-200 P1 & 83.30 & 85.87 & 85.16 & 87.95 & +2.56 [+1.40, +3.77] & +1.85 [+0.66, +3.13] & +0.71 [$-$0.90, +2.33] & +0.23 [$-$0.65, +1.14] \\
\midrule
\multicolumn{9}{l}{\emph{DFN2B L/14, NABirds$\to$CUB-59 (ep.\ 8/45)}} \\
held-out, CUB-59 P3 & 46.35 & 45.79 & 45.72 & 44.42 & $-$0.56 [$-$2.76, +1.59] & $-$0.63 [$-$2.38, +0.97] & +0.07 [$-$2.52, +2.69] & $-$0.74 [$-$2.27, +0.84] \\
held-out, CUB-59 P1 & 87.69 & 87.98 & 88.40 & 87.96 & +0.29 [$-$0.77, +1.40] & +0.71 [$-$0.53, +2.03] & $-$0.42 [$-$1.38, +0.52] & $-$0.74 [$-$1.52, +0.02] \\
target, NAB P3 & 8.11 & 8.63 & 8.00 & 8.49 & +0.52 [$-$0.04, +1.08] & $-$0.11 [$-$0.45, +0.25] & +0.63 [$-$0.04, +1.31] & $-$0.03 [$-$0.28, +0.23] \\
target, NAB P1 & 78.78 & 79.66 & 80.58 & 81.70 & +0.88 [+0.10, +1.65] & +1.80 [+0.90, +2.72] & $-$0.92 [$-$2.03, +0.15] & +0.25 [$-$0.34, +0.81] \\
\midrule
\multicolumn{9}{l}{\emph{MetaCLIP L/14, NABirds$\to$CUB-59 (ep.\ 4/40)}} \\
held-out, CUB-59 P3 & 47.43 & 46.74 & 46.70 & 45.49 & $-$0.69 [$-$4.30, +3.14] & $-$0.73 [$-$3.77, +2.47] & +0.04 [$-$4.48, +4.62] & $-$0.52 [$-$2.67, +1.54] \\
held-out, CUB-59 P1 & 83.62 & 82.17 & 82.98 & 80.51 & $-$1.45 [$-$4.27, +1.34] & $-$0.64 [$-$3.30, +1.78] & $-$0.80 [$-$3.92, +2.60] & $-$1.02 [$-$2.79, +0.73] \\
target, NAB P3 & 9.41 & 10.76 & 10.78 & 12.67 & +1.35 [+0.47, +2.23] & +1.37 [+0.71, +2.06] & $-$0.02 [$-$1.08, +1.06] & +0.54 [$-$0.02, +1.11] \\
target, NAB P1 & 70.79 & 76.15 & 76.36 & 81.97 & +5.36 [+4.20, +6.54] & +5.58 [+4.18, +7.03] & $-$0.21 [$-$1.82, +1.41] & +0.24 [$-$0.66, +1.15] \\
\bottomrule
\end{tabular}}
\caption{Additional encoder-state swaps on three additional SSR+ND runs, selected by a rule fixed before any swap was run (Appendix~\ref{app:G_swapext}). States: P = P3-peak checkpoint, D = deployment checkpoint, I = image encoder, T = text/SSR state (checkpoint epochs peak/deploy in the run label); effects use the same definitions as Table~\ref{tab:G_towerboot} (image-late = D$_I$+P$_T$ $-$ P$_I$+P$_T$; text/SSR-late = P$_I$+D$_T$ $-$ P$_I$+P$_T$). Paired class bootstrap over evaluation classes (10,000 resamples); NABirds-trained runs are evaluated on the dataset-class-disjoint CUB-59 subset (59 classes), target rows on the run's own training-domain test set. Independent existing runs, not stochastic replicates.}
\label{tab:G_swapext}
\end{table}

%% file: supplement/tables/audit/G_crossstate.tex
\begin{table}[!htbp]
\centering
\footnotesize\setlength{\tabcolsep}{3.5pt}
\begin{tabular}{lrrrrrr}
\toprule
State & P1 correct & P1 wrong & P1 margin & P3 correct & P3 wrong & P3 margin \\
\midrule
\multicolumn{7}{l}{\textbf{CUB-200 ($n{=}200$)}} \\
peak & 0.4533 & 0.3183 & 0.1349 & 0.2157 & 0.2280 & -0.0123 \\
image late only & 0.5089 & 0.3453 & 0.1636 & 0.2352 & 0.2485 & -0.0134 \\
text late only & 0.4567 & 0.3064 & 0.1504 & 0.2117 & 0.2201 & -0.0084 \\
deploy & 0.5232 & 0.3366 & 0.1866 & 0.2327 & 0.2416 & -0.0089 \\
\midrule
\multicolumn{7}{l}{\textbf{NAB-348 ($n{=}348$)}} \\
peak & 0.4077 & 0.3559 & 0.0518 & 0.1383 & 0.2028 & -0.0645 \\
image late only & 0.4492 & 0.4042 & 0.0450 & 0.1467 & 0.2228 & -0.0761 \\
text late only & 0.4069 & 0.3554 & 0.0515 & 0.1335 & 0.1985 & -0.0650 \\
deploy & 0.4484 & 0.4038 & 0.0446 & 0.1427 & 0.2197 & -0.0770 \\
\midrule
\bottomrule
\end{tabular}
\caption{Cross-state embedding recombination, SSR+ND (DFN5B H/14-378, CUB-trained; single run). For each class, ``correct'' is the image-to-text cosine similarity to the correct class text anchor, ``wrong'' is the similarity to the hardest competing class anchor, and margin is correct minus wrong; entries are class-macro averages over the stated number of classes, computed for P1 and P3 text anchors. States follow Table~\ref{tab:G_tower}. Figure~\ref{fig:mechanism}c plots the NAB-348 margins as change relative to the peak state.}
\label{tab:G_crossstate}
\end{table}

%% file: supplement/tables/audit/G_anchor.tex
\begin{table}[!htbp]
\centering
\footnotesize\setlength{\tabcolsep}{3.5pt}
\begin{tabular}{lllrrrrrr}
\toprule
Model & Eval. & Anchor & PT corr. & PT wrong & Peak corr. & Peak wrong & Dep. corr. & Dep. wrong \\
\midrule
SSR+ND & CUB-200 & P3 & 0.3100 & 0.3217 & 0.2157 & 0.2280 & 0.2327 & 0.2416 \\
 & CUB-200 & P1 & 0.3659 & 0.3290 & 0.4533 & 0.3183 & 0.5232 & 0.3366 \\
 & NAB-348 & P3 & 0.2796 & 0.3082 & 0.1383 & 0.2028 & 0.1427 & 0.2197 \\
 & NAB-348 & P1 & 0.3669 & 0.3425 & 0.4077 & 0.3559 & 0.4484 & 0.4038 \\
\midrule
U (text-frozen) & CUB-200 & P3 & 0.3100 & 0.3217 & 0.1993 & 0.2109 & 0.2648 & 0.2462 \\
 & CUB-200 & P1 & 0.3659 & 0.3290 & 0.4017 & 0.3157 & 0.4669 & 0.3386 \\
 & NAB-348 & P3 & 0.2796 & 0.3082 & 0.1325 & 0.1792 & 0.1519 & 0.2109 \\
 & NAB-348 & P1 & 0.3669 & 0.3425 & 0.3872 & 0.3457 & 0.4129 & 0.3835 \\
\midrule
\bottomrule
\end{tabular}
\caption{Correct-class versus hardest-competitor text-anchor similarity (class-macro means) at pretrained (PT), peak, and deployment checkpoints for SSR+ND (peak epoch 20, deployment epoch 60) and an independent text-frozen U run (peak epoch 20, deployment epoch 75; a separate run from the main U run, whose deployment epoch is 25). In the text-frozen run the entire text encoder including layer norms is locked, and the text anchors were verified to be bit-identical across PT, peak, and deployment, so all change is image-side.}
\label{tab:G_anchor}
\end{table}

%% file: supplement/tables/audit/G_enc.tex
\begin{table}[!htbp]
\centering\footnotesize\setlength{\tabcolsep}{3pt}
\begin{tabular}{lllllr}
\toprule
Model & Eval. & Probe & Competitor & Encroachment ($\times10^{-3}$) [95\% CI] & Classes $>0$ \\
\midrule
SSR+ND (joint) & CUB-200 & P1 & hardest wrong per checkpoint & $-51.66$ [-55.49, -47.88] & 2\% \\
SSR+ND (joint) & CUB-200 & P1 & fixed at peak (per image) & $-63.80$ [-67.61, -60.04] & 1\% \\
SSR+ND (joint) & CUB-200 & P1 & fixed at peak (per class) & $-71.66$ [-75.93, -67.35] & 2\% \\
\addlinespace
SSR+ND (joint) & CUB-200 & P3 & hardest wrong per checkpoint & $-3.41$ [-6.03, -0.85] & 40\% \\
SSR+ND (joint) & CUB-200 & P3 & fixed at peak (per image) & $-11.84$ [-14.41, -9.35] & 22\% \\
SSR+ND (joint) & CUB-200 & P3 & fixed at peak (per class) & $-11.34$ [-14.31, -8.40] & 28\% \\
\addlinespace
SSR+ND (joint) & NAB-348 & P1 & hardest wrong per checkpoint & $+7.21$ [+5.55, +8.88] & 70\% \\
SSR+ND (joint) & NAB-348 & P1 & fixed at peak (per image) & $+2.22$ [+0.60, +3.82] & 56\% \\
SSR+ND (joint) & NAB-348 & P1 & fixed at peak (per class) & $+2.12$ [+0.31, +3.88] & 57\% \\
\addlinespace
SSR+ND (joint) & NAB-348 & P3 & hardest wrong per checkpoint & $+12.49$ [+10.80, +14.12] & 76\% \\
SSR+ND (joint) & NAB-348 & P3 & fixed at peak (per image) & $+3.56$ [+1.94, +5.18] & 59\% \\
SSR+ND (joint) & NAB-348 & P3 & fixed at peak (per class) & $+3.42$ [+1.43, +5.42] & 55\% \\
\addlinespace
U (text frozen) & CUB-200 & P1 & hardest wrong per checkpoint & $-42.33$ [-45.97, -38.68] & 6\% \\
U (text frozen) & CUB-200 & P1 & fixed at peak (per image) & $-50.74$ [-54.61, -46.86] & 3\% \\
U (text frozen) & CUB-200 & P1 & fixed at peak (per class) & $-51.53$ [-55.22, -47.82] & 2\% \\
\addlinespace
U (text frozen) & CUB-200 & P3 & hardest wrong per checkpoint & $-30.21$ [-33.84, -26.57] & 12\% \\
U (text frozen) & CUB-200 & P3 & fixed at peak (per image) & $-42.14$ [-45.98, -38.34] & 6\% \\
U (text frozen) & CUB-200 & P3 & fixed at peak (per class) & $-41.83$ [-46.10, -37.63] & 8\% \\
\addlinespace
U (text frozen) & NAB-348 & P1 & hardest wrong per checkpoint & $+12.06$ [+10.19, +14.01] & 76\% \\
U (text frozen) & NAB-348 & P1 & fixed at peak (per image) & $+4.88$ [+3.09, +6.68] & 61\% \\
U (text frozen) & NAB-348 & P1 & fixed at peak (per class) & $+5.83$ [+3.89, +7.81] & 64\% \\
\addlinespace
U (text frozen) & NAB-348 & P3 & hardest wrong per checkpoint & $+12.26$ [+10.38, +14.20] & 74\% \\
U (text frozen) & NAB-348 & P3 & fixed at peak (per image) & $-1.31$ [-3.21, +0.67] & 47\% \\
U (text frozen) & NAB-348 & P3 & fixed at peak (per class) & $-1.09$ [-3.42, +1.33] & 45\% \\
\addlinespace
\bottomrule
\end{tabular}
\caption{Peak$\to$deployment competitor encroachment, $\Delta s_{\mathrm{wrong}}-\Delta s_{\mathrm{correct}}$ (class-macro mean of image--text-anchor cosine changes; positive = the competitor closes the gap relative to the correct anchor). ``Hardest wrong per checkpoint'' re-selects the maximizing wrong class independently at peak and at deployment and can rise merely because wrong-class similarities spread or the maximizer changes; the fixed-competitor rows freeze each image's (or class's) hardest wrong class at the peak checkpoint and re-evaluate it at deployment. Intervals: paired class bootstrap over evaluation classes (10,000 resamples), not training-seed replication. Peak epoch 20; deployment epoch 60 (SSR+ND) or 75 (U, text encoder frozen; text anchors bit-identical across checkpoints). Single runs, DFN5B H/14-378 fine-tuned on CUB-200.}
\label{tab:G_enc}
\end{table}

%% file: supplement/tables/audit/G_enclink.tex
\begin{table}[!htbp]
\centering\footnotesize\setlength{\tabcolsep}{3pt}
\begin{tabular}{lllllr}
\toprule
Model & Eval. & Probe & Competitor & Spearman $\rho$ [95\% CI] & Mean $\Delta$acc.\ (pp) \\
\midrule
SSR+ND (joint) & CUB-200 & P1 & hardest wrong per checkpoint & $-$0.09 [$-$0.24, +0.06] & +0.2 \\
SSR+ND (joint) & CUB-200 & P1 & fixed at peak (per image) & $-$0.09 [$-$0.24, +0.06] & +0.2 \\
SSR+ND (joint) & CUB-200 & P3 & hardest wrong per checkpoint & $-$0.60 [$-$0.69, $-$0.51] & +0.6 \\
SSR+ND (joint) & CUB-200 & P3 & fixed at peak (per image) & $-$0.57 [$-$0.65, $-$0.47] & +0.6 \\
SSR+ND (joint) & NAB-348 & P1 & hardest wrong per checkpoint & $-$0.57 [$-$0.63, $-$0.49] & $-$5.3 \\
SSR+ND (joint) & NAB-348 & P1 & fixed at peak (per image) & $-$0.58 [$-$0.64, $-$0.50] & $-$5.3 \\
SSR+ND (joint) & NAB-348 & P3 & hardest wrong per checkpoint & $-$0.39 [$-$0.48, $-$0.30] & $-$3.2 \\
SSR+ND (joint) & NAB-348 & P3 & fixed at peak (per image) & $-$0.35 [$-$0.44, $-$0.26] & $-$3.2 \\
U (text frozen) & CUB-200 & P1 & hardest wrong per checkpoint & $-$0.20 [$-$0.34, $-$0.05] & +1.4 \\
U (text frozen) & CUB-200 & P1 & fixed at peak (per image) & $-$0.17 [$-$0.32, $-$0.02] & +1.4 \\
U (text frozen) & CUB-200 & P3 & hardest wrong per checkpoint & $-$0.36 [$-$0.48, $-$0.22] & +16.6 \\
U (text frozen) & CUB-200 & P3 & fixed at peak (per image) & $-$0.39 [$-$0.51, $-$0.25] & +16.6 \\
U (text frozen) & NAB-348 & P1 & hardest wrong per checkpoint & $-$0.64 [$-$0.70, $-$0.56] & $-$9.9 \\
U (text frozen) & NAB-348 & P1 & fixed at peak (per image) & $-$0.60 [$-$0.66, $-$0.52] & $-$9.9 \\
U (text frozen) & NAB-348 & P3 & hardest wrong per checkpoint & $-$0.45 [$-$0.53, $-$0.36] & $-$2.6 \\
U (text frozen) & NAB-348 & P3 & fixed at peak (per image) & $-$0.41 [$-$0.49, $-$0.32] & $-$2.6 \\
\bottomrule
\end{tabular}
\caption{Class-level Spearman correlation between encroachment and the peak$\to$deployment change in class top-1 accuracy (class bootstrap over evaluation classes). The relationship is partly mechanical, because encroachment and top-1 retrieval derive from the same similarity scores; it is supportive consistency, not independent mechanistic evidence.}
\label{tab:G_enclink}
\end{table}

%% file: supplement/tables/audit/G_loo.tex
\begin{table}[!htbp]
\centering
\small\setlength{\tabcolsep}{3.5pt}
\begin{tabular}{llrrr}
\toprule
Model & Eval. & PT & Peak & Deploy \\
\midrule
SSR+ND & CUB-200 & 89.2 & 90.4 & 90.6 \\
 & NAB-348 & 90.9 & 90.6 & 88.4 \\
U (text-frozen) & CUB-200 & 89.2 & 90.3 & 90.6 \\
 & NAB-348 & 90.8 & 91.2 & 88.0 \\
\bottomrule
\end{tabular}
\caption{Image-only leave-one-out prototype accuracy (class-macro, \%). For each image, the class prototype is the mean image embedding of the other images of the same class; the image is assigned to the nearest prototype. This is an oracle diagnostic that uses held-out-image prototypes; it is not a zero-shot capability. Class structure remains largely preserved (88.0--91.2\%), although NAB-348 shows a 2.7--3.1\,pp decline from peak to deployment.}
\label{tab:G_loo}
\end{table}

%% file: supplement/tables/audit/G_geom.tex
\begin{table}[!htbp]
\centering
\footnotesize\setlength{\tabcolsep}{3.5pt}
\begin{tabular}{llrrrr}
\toprule
Geometry metric & Target & SSR+ND CUB & SSR+ND NAB & U-tf CUB & U-tf NAB \\
\midrule
Tangent-residual alignment (P3) & $\Delta$P3 & 0.33* & 0.25* & 0.15* & 0.21* \\
 & $\Delta$P1 & 0.05 & -0.09 & -0.04 & 0.00 \\
\midrule
Tangent-residual alignment (P2) & $\Delta$P3 & 0.05 & 0.11* & 0.14 & 0.19* \\
 & $\Delta$P1 & 0.01 & 0.05 & 0.03 & 0.10 \\
\midrule
Raw alignment (P3) & $\Delta$P3 & 0.23* & 0.19* & 0.12 & 0.19* \\
 & $\Delta$P1 & 0.02 & -0.15* & -0.06 & -0.08 \\
\midrule
Residual-direction coherence & $\Delta$P3 & -0.04 & -0.06 & -0.05 & -0.07 \\
 & $\Delta$P1 & 0.02 & 0.21* & 0.04 & 0.13* \\
\midrule
Manifold radius & $\Delta$P3 & -0.03 & 0.06 & 0.06 & 0.06 \\
 & $\Delta$P1 & -0.05 & -0.26* & -0.07 & -0.20* \\
\midrule
Effective rank & $\Delta$P3 & 0.08 & -0.01 & 0.04 & -0.02 \\
 & $\Delta$P1 & 0.10 & 0.11* & 0.17* & 0.02 \\
\midrule
Participation ratio & $\Delta$P3 & 0.07 & 0.00 & 0.03 & -0.02 \\
 & $\Delta$P1 & -- & -- & -- & -- \\
\midrule
PC1 variance fraction & $\Delta$P3 & -0.04 & 0.01 & -0.00 & 0.02 \\
 & $\Delta$P1 & -- & -- & -- & -- \\
\midrule
\bottomrule
\end{tabular}
\caption{Class-level Spearman correlations between late-minus-peak changes in coarse geometry metrics and changes in class accuracy (n = 200 CUB classes, 348 NAB classes; 10,000 class resamples for the 95\% intervals). Asterisks mark intervals excluding 0. Tangent-residual alignment vs.\ $\Delta$P3 is weak but non-null in all four cells ($\rho\in[0.15,0.33]$); residual-direction coherence, manifold radius, and effective rank show no reliable relationship with $\Delta$P3, although some show non-null relationships with $\Delta$P1. The two models are SSR+ND and an independent text-frozen U run (U-tf). Full intervals are provided in the supporting results.}
\label{tab:G_geom}
\end{table}

%% file: supplement/tables/audit/H_wise.tex
\begin{table}[!htbp]
\centering
\small\setlength{\tabcolsep}{3.5pt}
\begin{tabular}{lrrrrr}
\toprule
Method & $\alpha$ & CUB P1 & CUB P3 & NAB-348 P1 & NAB-348 P3 \\
\midrule
N & 0 & 88.83 & 34.35 & 82.93 & 13.85 \\
 & 0.25 & 90.87 & 36.61 & 83.60 & 14.60 \\
 & 0.5 & 91.46 & 35.95 & 82.25 & 15.06 \\
 & 0.75 & 91.39 & 33.62 & 78.06 & 15.00 \\
 & 1 & 91.06 & 31.12 & 71.55 & 13.74 \\
\midrule
U & 0 & 88.75 & 34.45 & 83.03 & 13.87 \\
 & 0.25 & 90.82 & 39.01 & 83.25 & 15.57 \\
 & 0.5 & 91.47 & 41.68 & 82.03 & 17.00 \\
 & 0.75 & 91.49 & 40.30 & 79.97 & 16.95 \\
 & 1 & 91.16 & 35.78 & 76.00 & 15.14 \\
\midrule
\bottomrule
\end{tabular}
\caption{WiSE-style weight interpolation $\theta(\alpha)=(1-\alpha)\theta_{\mathrm{PT}}+\alpha\theta_{\mathrm{FT}}$ for N (deployment epoch 40) and U (deployment epoch 25), DFN5B H/14-378 fine-tuned on CUB-200; $\alpha=0$ is the pretrained model and $\alpha=1$ the deployed checkpoint. Interpolation is applied at evaluation time to all trainable parameters; no training is involved. Values are R@1 (\%) on CUB-200 (target) and NAB-348 (strict held-out). Single seed, coarse $\alpha$ grid, one backbone and training domain.}
\label{tab:H_wise}
\end{table}

%% file: supplement/tables/audit/H_hardneg.tex
\begin{table}[!htbp]
\centering
\small\setlength{\tabcolsep}{3.5pt}
\begin{tabular}{rrrrrr}
\toprule
Epoch & CUB P3 & NAB-348 P3 & CUB P1 & CUB align. & NAB align. \\
\midrule
8 & 37.87 / 38.02 & 15.91 / 16.18 & 88.28 / 88.44 & 0.463 / 0.460 & 0.263 / 0.262 \\
12 & 40.25 / 39.87 & 16.76 / 17.32 & 88.99 / 89.02 & 0.469 / 0.457 & 0.287 / 0.283 \\
20 & 41.32 / 40.42 & 15.57 / 16.90 & 90.06 / 90.30 & 0.454 / 0.412 & 0.281 / 0.268 \\
40 & 43.04 / 39.82 & 13.71 / 14.36 & 89.75 / 89.94 & 0.410 / 0.352 & 0.254 / 0.244 \\
100 & 41.27 / 40.01 & 12.38 / 13.30 & 90.02 / 90.25 & 0.355 / 0.315 & 0.231 / 0.228 \\
\bottomrule
\end{tabular}
\caption{Hard-negative control versus the matched SSR baseline, DFN5B H/14-378, CUB-trained (single seed, single $\lambda$). Each cell is hard-negative / SSR. The objective adds $\lambda L_{\mathrm{HN}}$ with $\lambda=0.5$ to the multi-positive contrastive loss, where $L_{\mathrm{HN}}=\mathrm{mean}\,\mathrm{softplus}(s_{\mathrm{neg}}-s_{\mathrm{pos}})$ is computed on the final fused text representation, and each negative differs from the true profile by one controlled attribute edit (pool of 618 candidates covering all 200 classes). Alignment is representational similarity between image and P3 text geometry. At each run's own P1-selected checkpoint (hard-negative epoch 20, SSR epoch 75) CUB P3 is 41.32 vs.\ 39.97 and NAB-348 P3 is 15.57 vs.\ 13.34; at epoch 100, NAB-348 P3 is 12.38 vs.\ 13.30. Peak NAB-348 P3 is 16.76 (hard-negative) vs.\ 17.35 (SSR).}
\label{tab:H_hardneg}
\end{table}